%% file: main.tex
\documentclass[10pt,twocolumn,letterpaper]{article}
\usepackage{cvpr} 
\input{preamble}

\begin{document}
\twocolumn[{%
  \renewcommand\twocolumn[1][]{#1}%
  \maketitle
  \begin{minipage}{\textwidth}
    \centering
    \input{sections/teaser}
  \end{minipage}
}]

\begingroup
\renewcommand{\thefootnote}{}
\footnotetext{%
  \textsuperscript{*} Equal contribution.\par
  \textsuperscript{$\dagger$} Work done during internship at Kling AI.\par
  \textsuperscript{$\ddagger$} Corresponding author.}
\endgroup

\input{sections/abstract}
\input{sections/intro}
\input{sections/related-work}
\input{sections/method}

\input{sections/benchmark}
\input{sections/experiment}

\input{sections/conclusion}
\input{sections/llm_statement}

\begingroup
  \small
  \bibliographystyle{ieeenat_fullname}
  \bibliography{references}
\endgroup

\clearpage
\input{appendix}

\end{document}

%% file: preamble.tex
\usepackage{microtype}
\usepackage{multirow}
\usepackage{array}
\usepackage{colortbl}
\usepackage{supertabular,needspace}
\makeatletter
\patchcmd{\x@supertabular}
  {\let\org@startpbox=\@startpbox
   \let\org@endpbox=\@endpbox
   \let\@startpbox=\ST@astartpbox
   \let\@endpbox=\ST@aendpbox}
  {\ifx\undefined\@startpbox@action
     \let\org@startpbox=\@startpbox
     \let\@startpbox=\ST@astartpbox
   \else
     \let\org@startpbox@action=\@startpbox@action
     \let\@startpbox@action=\ST@astartpbox
   \fi
   \let\org@endpbox=\@endpbox
   \let\@endpbox=\ST@aendpbox}
  {\typeout{SEGUE compatibility: patched supertabular startup}}
  {\PackageError{SEGUE}{Failed to patch supertabular startup}{Check supertabular package version}}
\patchcmd{\ST@restore}
  {\let\@startpbox\org@startpbox
   \let\@endpbox\org@endpbox}
  {\ifx\undefined\@startpbox@action
     \let\@startpbox\org@startpbox
   \else
     \let\@startpbox@action\org@startpbox@action
   \fi
   \let\@endpbox\org@endpbox}
  {\typeout{SEGUE compatibility: patched supertabular restore}}
  {\PackageError{SEGUE}{Failed to patch supertabular restore}{Check supertabular package version}}
\makeatother
\definecolor{linkblue}{rgb}{0.12,0.36,0.63}
\usepackage[breaklinks,colorlinks,allcolors=linkblue]{hyperref}
\newcommand{\method}{\textsc{Segue}}
\newcommand{\cdmd}{\textsc{SpanDMD}}
\newcommand{\bench}{\textsc{OpenTrans-360}}

\title{No Corners Cut: State-Grounded Transitions for Mid-Stream Prompt Switches \\in Video Generation}
\author{%
  {\normalsize
  Zejing Rao\textsuperscript{1,3,*\,$\dagger$},
  Ketong Ren\textsuperscript{1,*},
  Xiaoqiang Liu\textsuperscript{2},
  Yiping Meng\textsuperscript{2},
  Guoxin Zhang\textsuperscript{2},
  Fan Tang\textsuperscript{3,$\ddagger$}}
  \\[0.8em]
  {\normalsize \textsuperscript{1}University of Chinese Academy of Sciences}\\[0.15em]
  {\normalsize \textsuperscript{2}Kling AI\qquad
     \textsuperscript{3}University of Science and Technology Beijing}\\[0.45em]
  {\normalsize Project Page: \href{https://jayjrao.github.io/SEGUE/}{https://jayjrao.github.io/SEGUE/}}
}
\hypersetup{
  pdftitle={No Corners Cut: State-Grounded Transitions for Mid-Stream Prompt Switches in Video Generation},
  pdfauthor={Zejing Rao, Ketong Ren, Xiaoqiang Liu, Yiping Meng, Guoxin Zhang, Fan Tang}
}


%% file: sections/teaser.tex
\includegraphics[width=0.91\linewidth]{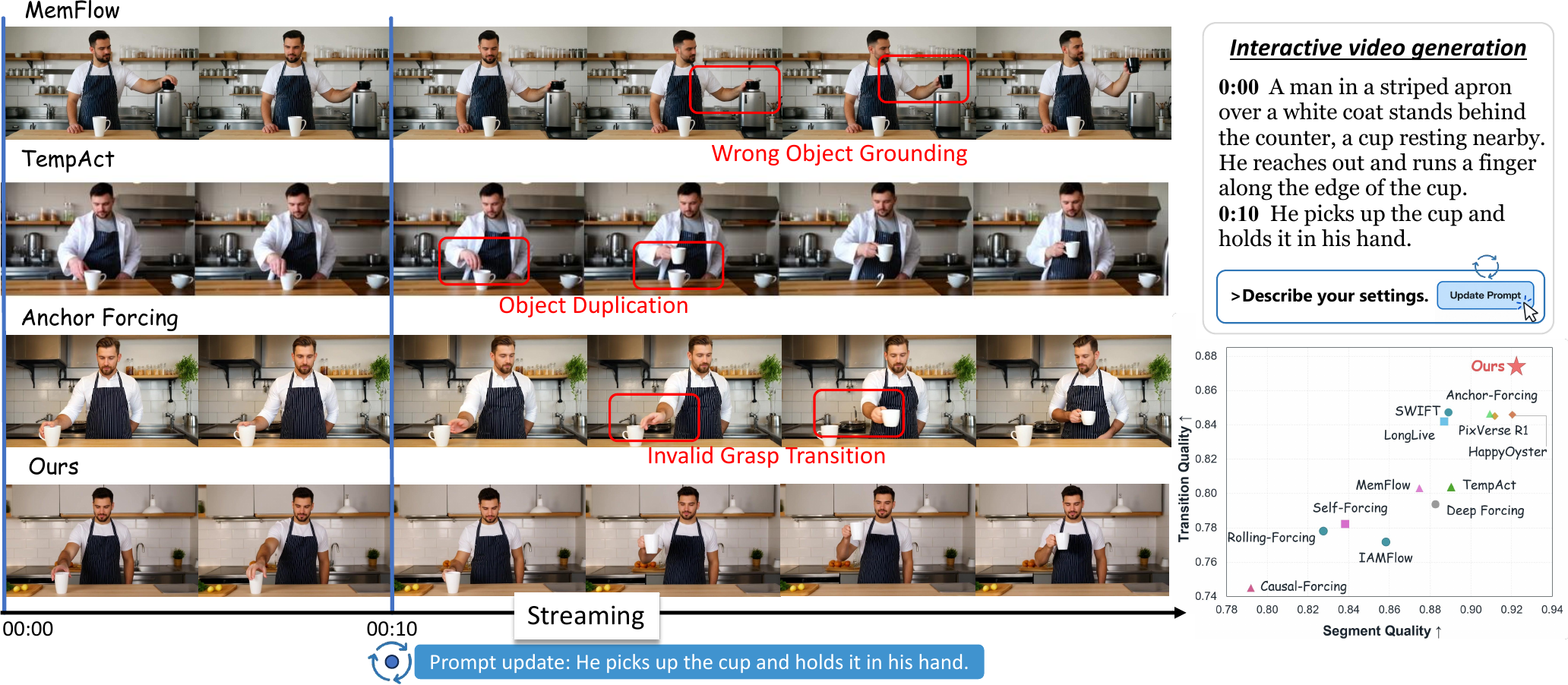}

\captionof{figure}{\textbf{State-grounded continuous transitions versus baseline ``corner cutting'' in interactive streaming video generation.}
\textbf{Left: Qualitative comparison at a mid-stream prompt switch.}
Existing streaming methods can still realize the new goal through implausible shortcuts, prematurely reaching the target while skipping intermediate states required for a coherent transition (e.g., object duplication and abrupt loss of contact, highlighted in red).
In contrast, \method{} (ours) grounds its dynamic planning on the realized visual boundary, synthesizing kinematically plausible preparatory phases prior to goal execution.
\textbf{Right: Quantitative trade-off on \bench{}.} \method{} (top-right)
achieves a better trade-off between segment quality and transition quality than causal autoregressive baselines and commercial video models.}

  \label{fig:teaser}

%% file: sections/abstract.tex
\begin{abstract}

Streaming video generators allow users to dynamically modulate video synthesis via mid-stream prompt switching.
Existing streaming methods can respond to the updated instruction while still \textit{cutting corners}, prematurely realizing goals or taking heuristic shortcuts that bypass necessary intermediate state changes needed for a plausible transition.
In this study, we present \method{}, a novel framework that makes this process explicit and trains the generator to execute these transitions faithfully.
At each switch, a training-free planner parses the latest frame and prompts, writes a few segue prompts with roles and durations, and then hands control back to the user's prompt.
Furthermore, to address the inherent difficulty of training causal models on short-lived temporal schedules without corrupting preparatory supervision, we introduce \cdmd{},
which evaluates each active prompt using the full rollout as temporal context while retaining its DMD residual only within the prompt's assigned span.
On \bench{}, a benchmark of 1,800 switches that scores how the old state exits and the new one begins, \method{} ranks first on all eight transition metrics and raises the overall score over the strongest baseline from 0.866 to 0.887.
It leads on four of six StreamAV-Bench instruction-response metrics and achieves the highest CLIP score in all five post-switch segments of VBench-Long evaluation, while the planner transfers to frozen autoregressive generators without retraining.

\end{abstract}

%% file: sections/intro.tex
\section{Introduction}

Recent breakthroughs in autoregressive video diffusion have catalyzed real-time, interactive streaming video    generation~\citep{yin2025causvid,huang2025selfforcing,cui2026selfforcingpp}.
Users can dynamically modify text prompts while the video stream unfolds.
A critical class of interactive updates involves continuing the narrative of the current scene rather than executing an abrupt cinematic cut.
Because previously rendered frames are immutable, the newly requested event must emerge organically from the prevailing visual context.

Recent streaming systems improve inference efficiency, visual consistency, and responsiveness to updated instructions through caching mechanisms or memory maintenance~\citep{yang2025longlive,ji2025memflow}. However, we observe that visual continuity does not necessarily imply a semantically correct transition path. 
As shown in Figure~\ref{fig:teaser}, existing models can maintain smooth appearance and object consistency while still exhibiting object duplication, invalid physical grasps, or abrupt state jumps during prompt transitions. 
Such failures are poorly captured by frame-level smoothness
metrics, which can remain high even when models fail to
properly exit the preceding state or enter the new one.
We attribute this limitation partly to prompt-conditioned
training, which primarily associates text prompts with
realized visual content rather than the intermediate state
changes needed to reach it.
Consequently, after a prompt switch, the generator tends to move the current visual state rapidly toward content that matches the new instruction, without explicitly accounting for a semantically plausible transition path. We refer to this behavior as ``corner cutting.''

Addressing such corner cutting requires both formulating a transition path conditioned on the runtime visual state, which existing pre-specified temporal planners are not designed to accommodate, and reliably executing short-lived prompts through distillation.
The latter is particularly non-trivial under standard distribution-matching distillation (DMD)\citep{yin2024dmd,yin2025causvid}: evaluating the complete rollout under the target prompt judges preparatory frames against future events and thus reinforces shortcuts, whereas cropping rollouts into isolated clips severs the necessary temporal context for coherent continuation.

In this study, we present \method{}, a framework that explicitly plans transition trajectories from realized visual states and trains causal generators to execute them.
At each switch, a vision-language planner observes the latest generated frame together with the previous and updated prompts.
Inspired by state-grounded action selection in embodied planning \citep{ahn2022saycan,rana2023sayplan}, it determines which of four preparatory roles in a Transition Dependency Schema (\mbox{TD-Schema}) are still required by the realized state: \textsc{Terminate}, \textsc{Release}, \textsc{Align}, and \textsc{Entry}. The selected roles are instantiated as a short sequence of segue prompts that bridge the realized state to the updated prompt, followed by an \textsc{Execute} handoff that returns control to the user's original prompt.
The planner needs no training and can be attached to existing streaming generators.
To address the execution challenge, we propose \cdmd{} to train the generator on these schedules under one principle: responsibility for a stage is local, but the evidence for judging it is contextual.
For each prompt in the schedule, both the real-score and fake-score models are evaluated on the complete noised rollout while the resulting DMD contribution is retained only within the temporal span assigned to that prompt.
To evaluate the transition process targeted by these components systematically, we introduce \bench{}, a benchmark comprising 360 prompt streams and 1,800 mid-stream switches, with metrics covering both within-segment generation quality and switch-level transition quality.
On \bench{} our method achieves the strongest overall performance across both segment-level and transition-level evaluation as summarized in the right panel of Figure~\ref{fig:teaser} and ranks first on all eight transition metrics.
It also ranks first on four of six instruction-adherence and
interactive-response metrics on StreamAV-Bench \citep{liu2026streamav}, and achieves the highest CLIP score in all five post-switch segments of the VBench-Long\citep{huang2024vbenchcomprehensiveversatilebenchmark} evaluation.
Our contributions are as follows:
\begin{itemize}
  \setlength{\itemsep}{1pt}\setlength{\parskip}{0pt}\setlength{\topsep}{2pt}
  
\item We formulate mid-stream prompt switching as
state-grounded transition planning and introduce a
training-free planner that identifies
unmet prerequisites at the realized visual boundary
and compiles them into a dependency-aware schedule.

\item We introduce \cdmd{} to train causal generators
to follow short-lived segue prompts, assigning each
condition's DMD residual to its temporal span while
retaining full-rollout context for both score networks.

  \item We build \bench{}, which jointly evaluates within-segment generation quality and switch-level transition quality, with checklist metrics achieving an average Spearman $\rho$ of 0.87 against human judgments.
\end{itemize}

%% file: sections/related-work.tex
\section{Related Work}

\paragraph{Autoregressive and interactive video generation.}

Recent advances in causal and autoregressive video diffusion have enabled
efficient streaming generation. CausVid~\citep{yin2025causvid} distills
bidirectional diffusion models into few-step causal generators, while
Self-Forcing~\citep{huang2025selfforcing} reduces the train--inference gap by
training on self-generated histories. Subsequent works further improve
long-horizon stability and causal distillation, including Rolling Forcing,
Self-Forcing++, Causal Forcing, and Causal Forcing++
\citep{liu2026rolling,cui2026selfforcingpp,zhu2026causal,zhao2026causal}.
Many of these approaches build on distribution-matching objectives such as
DMD and DMD2~\citep{yin2024dmd,yin2024dmd2}, providing the causal generation
backbones required for continuous video synthesis.

Building on these advances, recent systems support online interaction through
changing text conditions. LongLive~\citep{yang2025longlive} enables sequential
prompt updates through causal generation and KV recaching, while Anchor
Forcing~\citep{yang2026anchor} improves switching through anchor-guided cache
management. Memory-based methods such as MemFlow, IAMFlow, and
LongLive-RAG~\citep{ji2025memflow,liu2026iamflow,hu2026longlive} further
improve long-horizon consistency under evolving prompts. Related streaming
settings also include multi-shot generation~\citep{chen2026longlive2,meng2026causalcine}, human-object interaction
generation\citep{rao2026streamhoi},
interactive world models~\citep{he2025matrix,lingbot-world},
and causal video editing~\citep{liang2025looking,wang2026liveedit,zhao2026sana}.
We focus specifically on prompt updates that should preserve the
continuity of the current scene rather than introduce a cut or reset.

\paragraph{Temporal planning and multi-event generation.}

Language models have increasingly been used to make temporal structure explicit
in video generation. DirecT2V~\citep{hong2023direct2v} expands an abstract
prompt into time-varying prompts, while
VideoDirectorGPT~\citep{lin2023videodirectorgpt} constructs structured
multi-scene plans with entity and layout constraints. TempAct~\citep{wang2026tempact}
extends temporal planning to autoregressive generation by decomposing a
pre-specified temporal prompt into span-aware step prompts and jointly
optimizing its executor. VLM-based methods additionally use visual reasoning
to derive or refine generation plans~\citep{yang2025vlipp}.
Multi-event video generation binds prompts to time spans within one clip, by training with
time-based positional encodings~\citep{wu2025mind}, by routing prompts to spans at
inference~\citep{promptrelay2026}, or by controlling attention across
prompts~\citep{cai2025ditctrl}. Transition generation fills the gap between two given
clips~\citep{chen2024seine,zhang2024tvg,zhang2024mavin}.
While these methods provide explicit temporal control, they generally assume that the generation objective, prompt schedule, or boundary conditions are specified in advance. In contrast, we address online prompt switches, where the required transition must be determined from the realized visual state when a new prompt arrives. Our method plans state-dependent transitions on the fly, while SpanDMD trains a causal generator to execute the resulting span-level prompts.

\paragraph{Video generation benchmarks.}

Existing benchmarks evaluate complementary aspects of video generation.
VBench and EvalCrafter~\citep{huang2024vbench,liu2024evalcrafter} focus on
perceptual quality, motion, and text--video alignment, while T2VBench,
T2V-CompBench, and VBench-2.0
\citep{ji2024t2vbench,sun2025t2v,zheng2025vbench} extend evaluation toward
compositional semantics and state consistency. More recent long-form and
streaming benchmarks~\citep{liu2026iamflow,miao2026video,liu2026streamav}
further consider temporal consistency and responsiveness under evolving
generation. Our \bench{} instead explicitly evaluates the semantic transition
from the realized pre-switch state to the newly requested event, including
whether the preceding event is resolved and the target event is reached
through a plausible intermediate process.

%% file: sections/method.tex
\section{Method}
\label{sec:method}

\begin{figure*}[t]
  \centering
  \includegraphics[width=\linewidth]{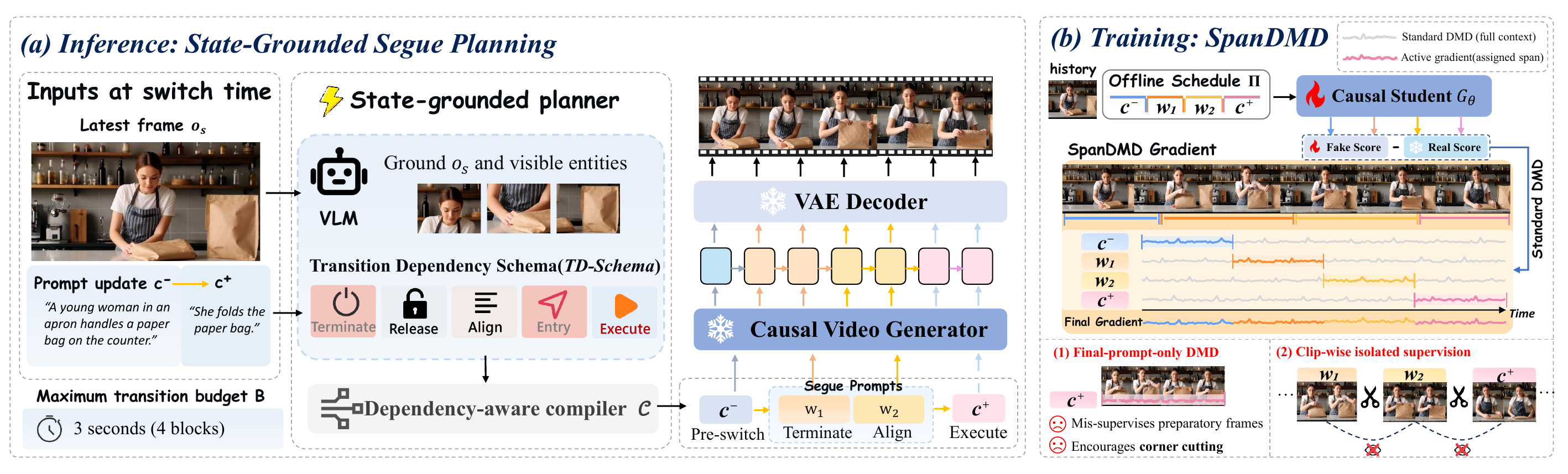}

  \caption{Overview of \method{}. At a switch, the planner reads the latest generated frame, the user prompts $c^{-}$ and $c^{+}$, and the transition budget $B$, and compiles a dependency-aware segue schedule under the TD-Schema leading to $c^{+}$; the causal generator executes this schedule block by block. During training, \cdmd{} evaluates the complete rollout under each active prompt, including $c^{-}$ and $c^{+}$, while retaining each prompt's DMD residual only on its assigned temporal span. The bottom-right illustrates two alternative supervision choices motivating \cdmd{}.}

  \label{fig:overview}
\end{figure*}

We study mid-stream prompt switches that should continue the current scene. \method{} has two components (Figure~\ref{fig:overview}): a state-grounded planner that writes the otherwise unspecified transition as a short schedule of segue prompts (Section~\ref{sec:tds}), and \cdmd{}, which trains the causal generator to execute such schedules (Section~\ref{sec:spandmd}).

\subsection{Problem Formulation}

Under prompt $c^{-}$, a causal generator $p_\theta$ emits fixed-length video blocks and maintains a bounded KV cache of past video latents~\citep{yang2025longlive}. When the user switches to $c^{+}$ after block $s$, the displayed history $H_s=(X_1,\ldots,X_s)$ is fixed. A direct switch applies $c^{+}$ to subsequent blocks but leaves unspecified how to reach the requested event from the current state.

Standard prompt-conditioned training aligns a prompt with the content of a clip. At a mid-stream switch, the realized state may still lack the prerequisites for $c^{+}$, leading to \emph{corner cutting}.
This challenge is reflected in a controlled 100-case study: the same autoregressive generators exhibit fewer transition failures when both events are specified in advance than
when the second event arrives online (Supplementary Sec.~A).
However, such advance knowledge is unavailable in the online setting.

We therefore formulate the switch as a state-grounded planning problem.
Given the last decoded frame $o_s$ of $H_s$, both prompts, and a budget of $B$ blocks, a vision--language planner proposes preparatory operations
\begin{equation}
\label{eq:rho}
\mathcal P = f_\phi(o_s,c^{-},c^{+},B).
\end{equation}
Each operation specifies a role, a \emph{segue prompt} describing an observable change, and a duration. A deterministic compiler $\mathcal C$ converts these operations into an executable schedule
\begin{equation}
\label{eq:plan}
\Pi = \mathcal C(\mathcal P)
= \big((w_k,b_k)\big)_{k=1}^{K},
\qquad
T_\Pi=\sum_{k=1}^{K}b_k\leq B,
\end{equation}
where $w_k$ is the $k$-th segue prompt and $b_k\in\mathbb N_{+}$ is its duration in blocks. After $T_\Pi$ blocks, generation resumes under $c^{+}$.

\subsection{State-Grounded Segue Planning}
\label{sec:tds}

\noindent\textbf{Grounding the transition.}
From $o_s$, the planner identifies the ongoing action and its progress, visible entities $\mathcal E_s$, object contacts, subject pose, and spatial relations. The observation takes precedence whenever it differs from $c^{-}$. Given the entities $\mathcal E^{+}$ required by $c^{+}$, the planner identifies which are missing, $\Delta\mathcal E=\mathcal E^{+}\setminus\mathcal E_s$.

\noindent\textbf{Transition Dependency Schema.}
TD-Schema defines four preparatory roles. \textsc{Terminate} ends or safely interrupts a conflicting action. \textsc{Release} removes an obstructing contact or interaction. \textsc{Align} establishes a required pose, spatial relation, interaction, or viewpoint. \textsc{Entry} brings a missing entity from $\Delta\mathcal E$ into view through an observable process. For example, switching from drinking to typing with both hands may require finishing the sip (\textsc{Terminate}), placing the cup on the desk (\textsc{Release}), and moving both hands over the keyboard (\textsc{Align}). The planner omits roles whose conditions already hold. \textsc{Execute} marks the handoff to $c^{+}$ and adds no segue prompt.

\noindent\textbf{Dependency-aware compilation.}
The compiler $\mathcal C$ orders operations by their dependencies, merges those achieved by the same observable change, and rejects schedules that violate structural constraints or exceed the budget after duration adjustment. Supplementary Sec.~G gives the planner prompts and compilation rules.

\noindent\textbf{Scheduled generation and execution.}
While the planner runs, generation continues under $c^{-}$. The schedule takes effect at the first block boundary after it returns. Counting from that boundary, let $\tau_0=0$, $\tau_k=\sum_{i=1}^{k}b_i$, and $I_k=\{\tau_{k-1}+1,\ldots,\tau_k\}$. The active prompt for block $j$ is
\begin{equation}
\label{eq:active-condition}
u(j)=
\begin{cases}
w_k, & j\in I_k,\\
c^{+}, & j>T_\Pi.
\end{cases}
\end{equation}
The planner changes only text conditioning, so it also works with frozen generators. We retain the underlying generator's KV re-cache operation at prompt boundaries~\citep{yang2025longlive}. 

\subsection{\cdmd{}: Span-Scoped Distillation with Full-Rollout Context}
\label{sec:spandmd}

\begin{figure}[t]
    \centering
    \includegraphics[width=\linewidth]{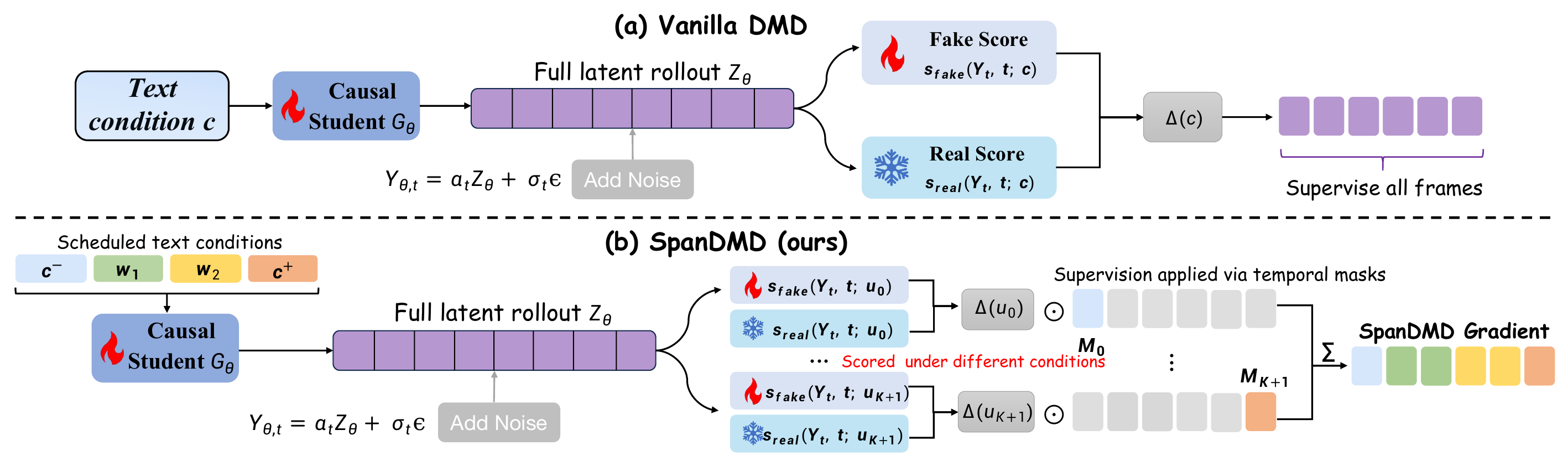}
    \caption{Vanilla DMD supervises the full rollout under one prompt. \cdmd{} scores the same rollout under each scheduled prompt and restricts its residual to the corresponding span.
    }
    \label{fig:dmd_comparison}
\end{figure}

Segue prompts last only one to a few blocks, so a generator
trained on longer-lived prompts may respond late, skip stages,
or blend them. Training the generator to follow such schedules
requires temporally aligned supervision. 
However, the pretrained teacher is conditioned on individual prompts rather than the full online schedule, and thus does not directly provide stage-aligned scores.
We therefore extend LongLive's
self-generated rollout and KV re-cache
training~\citep{yang2025longlive} using DMD~\citep{yin2024dmd,yin2025causvid}.
DMD updates the generator using the difference between a
learned fake score tracking the student distribution and a
frozen real score representing the teacher distribution.
Applying a single target or merged prompt to the full rollout
does not explicitly assign supervision to stages and can
encourage premature target realization. Scoring cropped
stages separately preserves their prompts but loses
cross-stage context. Instead, \cdmd{} evaluates both scores
on the full rollout under each stage's prompt and retains
their residual only within that stage's span
(Figure~\ref{fig:dmd_comparison}).

Formally, let $Z_\theta$ be a self-generated latent rollout
containing one instruction switch and $K$ segue stages.
We supervise all $K+2$ spans, including the pre-switch and
post-transition regions, with prompts
$u_0=c^{-}$, $u_\ell=w_\ell$ for $1\leq\ell\leq K$, and
$u_{K+1}=c^{+}$.
Each prompt $u_\ell$ is assigned a span $I_\ell$ and its
binary mask $M_\ell$.

At diffusion timestep $t$, let
$Y_{\theta,t}=\alpha_t Z_\theta+\sigma_t\epsilon$,
where $\epsilon\sim\mathcal{N}(0,I)$.
Evaluating both score networks on the complete
$Y_{\theta,t}$ under each $u_\ell$ gives the span-masked update
\begin{equation}
\label{eq:spandmd}
\begin{aligned}
g_{\mathrm{SpanDMD}}
  &= \mathbb{E}_{t,\epsilon}\!\left[
     J_{\theta,t}^{\top}
     \sum_{\ell=0}^{K+1}
     \lambda_\ell M_\ell \odot \Delta s_\ell
     \right], \\
\Delta s_\ell
  &= s_{\mathrm{fake}}(Y_{\theta,t},t;u_\ell)
     -s_{\mathrm{real}}(Y_{\theta,t},t;u_\ell), \\
J_{\theta,t}
  &= \partial Y_{\theta,t}/\partial\theta .
\end{aligned}
\end{equation}
Both scores are treated as constants (stop-gradient) when
updating the generator, and $\lambda_\ell$ weights span
$I_\ell$; standard DMD timestep weighting and gradient
normalization~\citep{yin2024dmd,yin2025causvid} are omitted
for clarity.
The fake-score network is trained on student-generated
rollouts using span-masked flow-matching losses under the
corresponding prompts, with the complete rollout as input
(Supplementary Sec.~H).


Crucially, span masking does not discard temporal
context. 
To interpret this, consider idealized exact scores
$s_a=\nabla_Y\log p_{a,t}(Y\mid u_\ell)$ associated with
noised full-rollout densities $p_{a,t}(\cdot\mid u_\ell)$, where $a\in\{\mathrm{fake},\mathrm{real}\}$.
Partition $Y_{\theta,t}$ into the span coordinates
$Y_{\theta,t,I_\ell}$ and the remaining context
$Y_{\theta,t,-I_\ell}$. Under this
exact-score assumption,
\begin{equation}
\label{eq:span_conditional_score}
\begin{aligned}
[\Delta s_\ell]_{I_\ell}
&=
\nabla_{Y_{\theta,t,I_\ell}}
\log
\frac{
p_{\mathrm{fake},t}
(Y_{\theta,t,I_\ell}
\mid Y_{\theta,t,-I_\ell},u_\ell)
}{
p_{\mathrm{real},t}
(Y_{\theta,t,I_\ell}
\mid Y_{\theta,t,-I_\ell},u_\ell)
}.
\end{aligned}
\end{equation}
This identity follows from joint-density factorization
and shows that the score residual restricted
to a span can still depend on the remaining rollout,
unlike scoring isolated clips.
Thus, the span masks specify where each prompt
contributes supervision, while full-rollout scoring
retains cross-stage context.
This provides a conditional-score
interpretation of \cdmd{} without requiring a teacher
that explicitly models the full prompt schedule.

%% file: sections/benchmark.tex
\section{\bench: Evaluating Mid-Stream Transitions}
\label{sec:benchmark}

\bench{} evaluates whether a generator can follow an updated prompt coherently from the state reached in its own rollout. It contains 360 streams of six 10-second prompts each, yielding five switches per 60-second video and 1,800 switches overall. 

\noindent\textbf{Construction.}
OpenTrans-360 is constructed independently from structured
combinations of subjects, settings, and events, expanded
into six-stage prompt streams by
Qwen2.5-72B-Instruct-AWQ~\cite{qwen2025qwen25technicalreport},
and filtered through automatic checks and human inspection
(Supplementary Sec.~I).
The streams remain within a persistent scene, covering
human-centric actions and interactions, object-centric
state or motion changes, and coordinated changes across
entities.

\noindent\textbf{Evaluation.}
We evaluate eleven metrics using InternVL3.5-38B~\citep{wang2025internvl35advancingopensourcemultimodal}, with each metric comprising ten binary criteria.
Three segment metrics assess prompt fidelity (SPF), motion continuity (SMC), and physical plausibility (SPP). Eight transition metrics assess visual and motion smoothness at the boundary (VBS, MBS), old-state exit (OSE), new-state entry (NSE), subject and scene preservation (SP, ScP), semantic update fidelity (SUF), and transition pacing (TP). 
The metrics evaluate observable state changes in generated videos, without relying on any segue prompts or planning schema.
In a blinded pairwise study on 30 stratified cases, benchmark scores correlated with method-level human win rates across all eleven metrics (mean Spearman $\rho=0.87$; Table~\ref{tab:human-alignment}). 
Supplementary Sec.~I details the sampling windows, Sec.~M lists the evaluation criteria, and Sec.~K describes the human alignment.

\begin{table}[t]
  \centering
  \footnotesize
  \setlength{\tabcolsep}{2pt}
  \caption{Spearman correlation between method-level human win rates
  and benchmark scores for each metric.}
  \label{tab:human-alignment}
  \resizebox{\columnwidth}{!}{%
    \begin{tabular}{@{}l*{11}{c}@{}}
      \toprule
      Metric         & SPF  & SMC  & SPP  & VBS  & MBS  & OSE  & NSE  & SP   & ScP  & SUF  & TP   \\
      \midrule
      Spearman $\rho$ & 0.82 & 0.88 & 0.78 & 0.83 & 0.85 & 0.93 & 0.90 & 0.88 & 0.91 & 0.90 & 0.89 \\
      \bottomrule
    \end{tabular}%
  }
\end{table}

%% file: sections/experiment.tex
\section{Experiments}
\label{sec:experiments}

\subsection{Settings}

\noindent\textbf{Implementation details.}
We initialize our 480p Wan2.1-T2V-1.3B~\citep{wan2025} causal generator from the base checkpoint that LongLive~\citep{yang2025longlive} trained on 5-second clips. 
LongLive uses this checkpoint for subsequent training on 60-second rollouts with mid-stream prompt switches. In the same
60-second rollouts setting, we train our generator with \cdmd{}.
A frozen Wan2.1-T2V-14B teacher provides the real score,
while a Wan2.1-T2V-1.3B fake-score network is trained jointly.
We use JoyAI-VL-Interaction~\citep{joyai2026vlinteraction}
as the planner without fine-tuning.
For training, we use Qwen3-VL-235B-A22B-Instruct\cite{qwen3technicalreport} to annotate 5-second VidProM clips~\citep{wang2024vidprom}. 
The model describes the first and last frames to obtain $c^{-}$ and $c^{+}$, respectively, and extracts ordered segue prompts from the observed intermediate state changes, yielding approximately 45k segue prompts in total (Supplementary Sec.~J).
Detailed parameters of sampling, planning, and training are provided in Supplementary Sec.~L.

\input{tables/main_comparison}
\input{tables/streamav_comparison_semantic}

\noindent\textbf{Benchmarks and baselines.}
We evaluate on \bench{} (Section~\ref{sec:benchmark}), StreamAV-Bench~\citep{liu2026streamav}, which measures streaming quality, instruction following, and responsiveness, and on VBench-Long~\citep{huang2024vbenchcomprehensiveversatilebenchmark} following MemFlow~\citep{ji2025memflow}. We compare with two closed-source systems, PixVerse R1~\citep{pixverse2026r1} and HappyOyster~\citep{happyoyster2026}, and ten open-source methods: Self-Forcing~\citep{huang2025selfforcing}, LongLive~\citep{yang2025longlive}, Rolling Forcing~\citep{liu2026rolling}, Deep Forcing~\citep{yi2025deepforcing}, MemFlow~\citep{ji2025memflow}, Causal Forcing~\citep{zhu2026causal}, Anchor Forcing~\citep{yang2026anchor}, SWIFT~\citep{tan2026swift}, IAMFlow~\citep{liu2026iamflow}, and TempAct~\citep{wang2026tempact}. They cover causal streaming, long-horizon stabilization, memory and cache adaptation, and temporal planning. On OpenTrans-360, all methods use the same prompt streams with nominal 10\,s prompt intervals. 

\subsection{Comparison Results}
\input{tables/vbench-clip}

\noindent\textbf{Results on \bench{}.}
Table~\ref{tab:OPENTRANS-360-results} shows that \method{} ranks first
on all eight transition metrics and achieves the highest overall score
(0.887 vs.\ 0.866). 
Paired 95\% case-bootstrap CIs further support these improvements,
with absolute gains of $\Delta=0.022$ on MBS
(CI $[\text{0.019},\text{0.026}]$), $\Delta=0.019$ on OSE
(CI $[\text{0.015},\text{0.027}]$), $\Delta=0.027$ on NSE
(CI $[\text{0.022},\text{0.035}]$), and $\Delta=0.021$ on Overall
(CI $[0.017,0.031]$).
NSE and OSE measure how the old event ends and the new one begins,
while MBS measures motion continuity across the switch.
Visual smoothness alone is less discriminative: all methods score
above 0.94 on VBS, whereas OSE ranges from 0.51 to 0.80.

\begin{figure*}[t]
\centering
\includegraphics[width=0.92\linewidth]{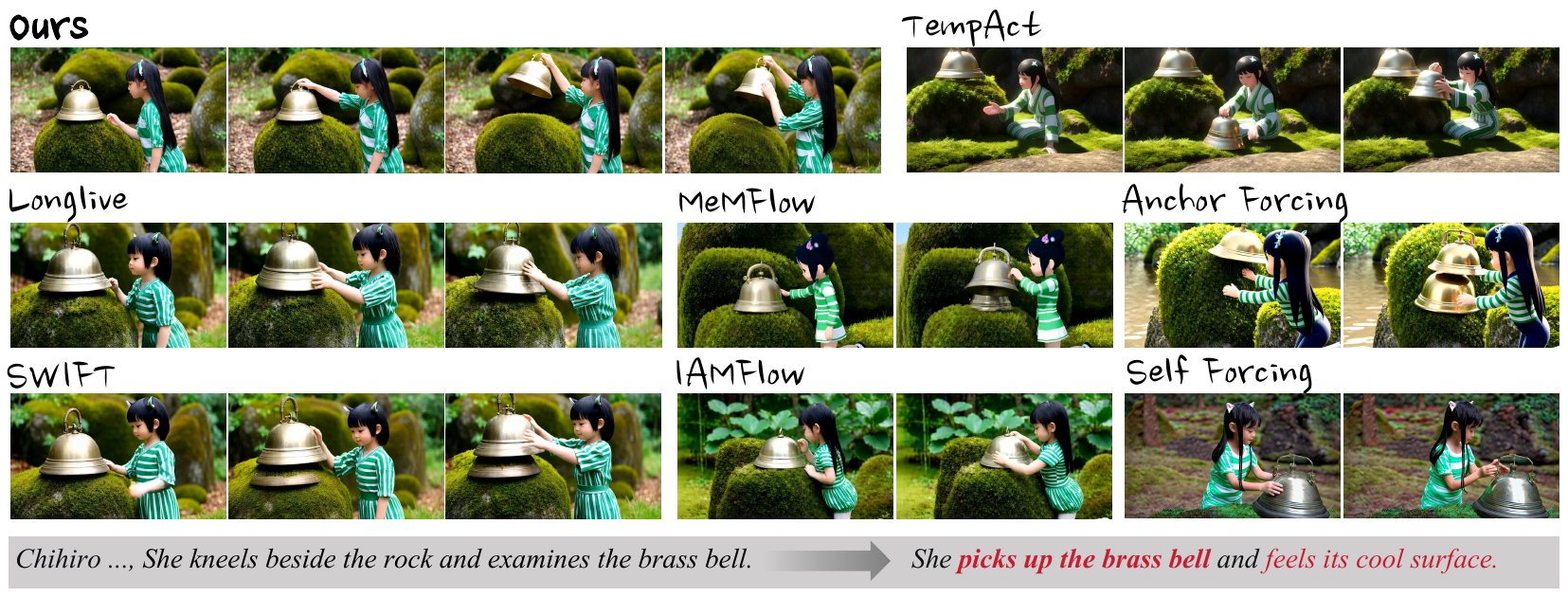}
\caption{Comparison at the switch from ``She kneels beside the rock and examines the brass bell'' to ``She picks up the brass bell and feels its cool surface.'' \method{} grasps and lifts the existing bell; TempAct and Anchor Forcing duplicate it, and several baselines never reach a lifted state.}
\label{fig:main-comparison}
\end{figure*}

Figure~\ref{fig:main-comparison} illustrates corner cutting when switching from examining a brass bell to picking it up.
\method{} correctly grasps and lifts the existing bell before touching its surface, whereas TempAct and Anchor Forcing duplicate the bell.
LongLive, IAMFlow, and Self-Forcing fail to complete the lift, while MemFlow and SWIFT exhibit unstable grasps and visible object deformation. 
\textit{In a blinded user study, \textsc{Segue} achieves overall-preference scores above 50\% against all twelve baselines (Supplementary Sec.~B).}

\noindent\textbf{Results on StreamAV-Bench and VBench.}
As reported in Table~\ref{tab:streamav-semantic}, \method{} ranks first on four of six instruction and response
metrics on StreamAV-Bench and improves PVC by 7.1\% over the
best baseline, while maintaining competitive visual quality
(0.603 VA, 2.838 VQ).
It also achieves the lowest VA-D and second-lowest VQ-D,
indicating strong long-horizon visual stability. Detailed results on StreamAV-Bench are reported in Supplementary Table~S3.
On VBench-Long~\citep{huang2024vbenchcomprehensiveversatilebenchmark}, as reported in Table~\ref{tab:vbench-multiprompt}, \method{} ranks first on three of four visual-quality
metrics and second on imaging quality, while achieving the highest
CLIP~\citep{radford2021learningtransferablevisualmodels} scores in all five segments after the first prompt switch.

\begin{figure*}[t]
\centering
\includegraphics[width=0.92\linewidth]{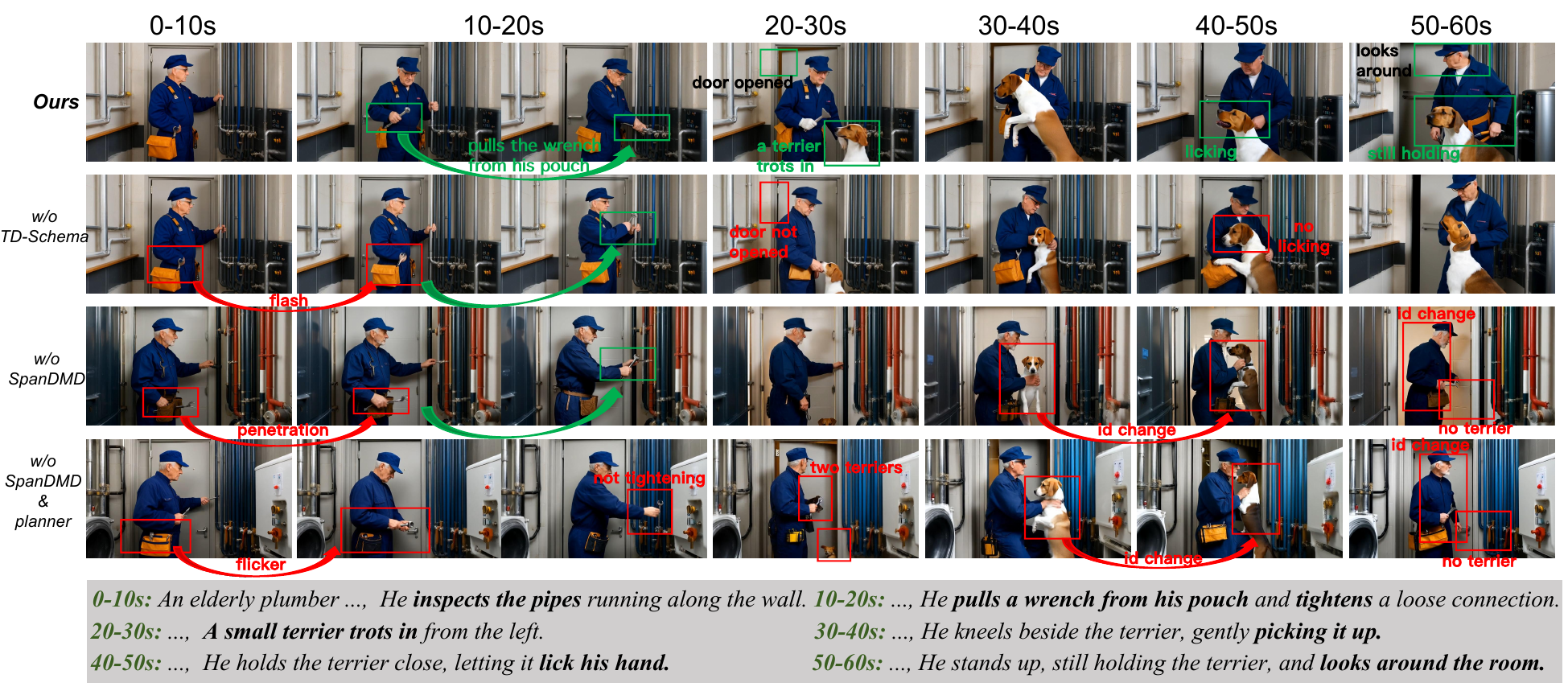}
\caption{Ablation on a six-prompt stream (prompts at the bottom). Green marks the intended progression; red marks failures such as flicker, interpenetration, duplicated or missing terriers, and identity changes.}
\label{fig:main-ablation}
\end{figure*}

\subsection{Ablation Study}
\label{sec:ablation}

\noindent\textbf{Variants.}
Table~\ref{tab:ablation} presents three groups of ablations.
The top group evaluates three component variants:
\emph{w/o TD-Schema} removes schema constraints while retaining
state-grounded planning;
\emph{w/o SpanDMD} retains the planned prompt schedule but
uses vanilla DMD with all conditions merged;
and \emph{w/o SpanDMD \& planner} trains with direct prompt
switches using vanilla DMD, without segue prompts during training or inference.
The middle group compares three alternative DMD supervision
strategies: final-prompt-only, cropped-span, and unmasked
per-condition DMD.
The bottom group compares the Self-Forcing and LongLive generators with their planner-augmented versions, keeping generator weights frozen.

\noindent\textbf{Effect of the planner.}
Introducing planned segue schedules during both training and inference, while retaining vanilla DMD, raises MBS from 0.824 to 0.869, OSE from 0.717 to 0.759, and SUF from 0.747 to 0.790, with smaller gains on the other transition metrics (Table~\ref{tab:ablation}). In Figure~\ref{fig:main-ablation}, the model without the planner switches to the next event before the pipe is tightened and later produces two terriers; with the planner, the sequence follows the intended progression from pulling out the wrench to the terrier's arrival and the interaction with it.
A controlled comparison shows that visual grounding improves
all six evaluated transition metrics and helps the same planner
identify state-dependent prerequisites missed by text-only
planning (Supplementary Sec.~F).
The planner also transfers to other generators without retraining: all six transition metrics improve for both Self-Forcing and LongLive, with SUF rising from 0.647 to 0.721 on Self-Forcing and OSE from 0.713 to 0.752 on LongLive.

\input{tables/ablation}

\begin{figure}[t]

    \centering
    \includegraphics[width=\linewidth]{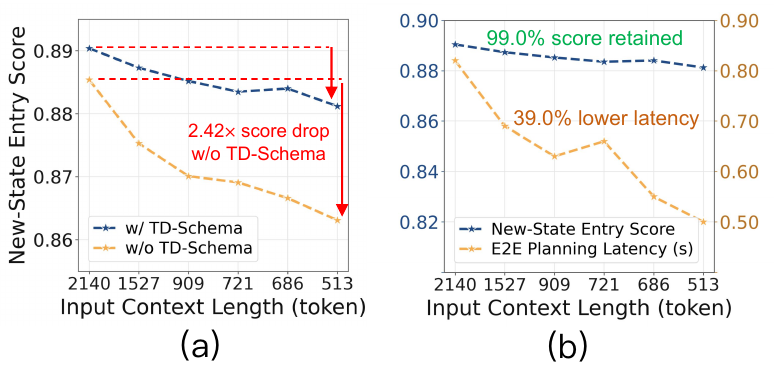}

    \caption{Planning quality and latency as the planner's input context shrinks, with and without TD-Schema.}
    \label{fig:td-schema}

\end{figure}

Performance is largely insensitive to the choice of planner VLM, while JoyAI-VL-Interaction offers the best quality--latency trade-off among the four tested planners.
Detailed quantitative comparisons are provided in
Supplementary Sec.~E.

\noindent\textbf{Effect of TD-Schema.}
The w/o TD-Schema variant gives the planner an unstructured natural-language description of the same content instead of the schema, with the same input length.
It is lower on all six transition metrics, most strongly on OSE (0.802 to 0.758). Here, input context length denotes the number of text tokens provided to the planner per call, including planning instructions, the preceding and updated prompts and chat-template tokens. TD-Schema also keeps the planner accurate with a short context (Figure~\ref{fig:td-schema}): reducing the context from 2,140 to 513 tokens lowers NSE only from 0.890 to 0.881, whereas without TD-Schema the same reduction lowers NSE 2.42 times as much.
The shorter context cuts planning latency from about 0.82\,s to 0.50\,s, below the time to generate one 12-frame block (approximately 0.58\,s at a generation throughput of 20.7 FPS).

\begin{figure}[h]
    \centering
      \includegraphics[width=\linewidth]
      {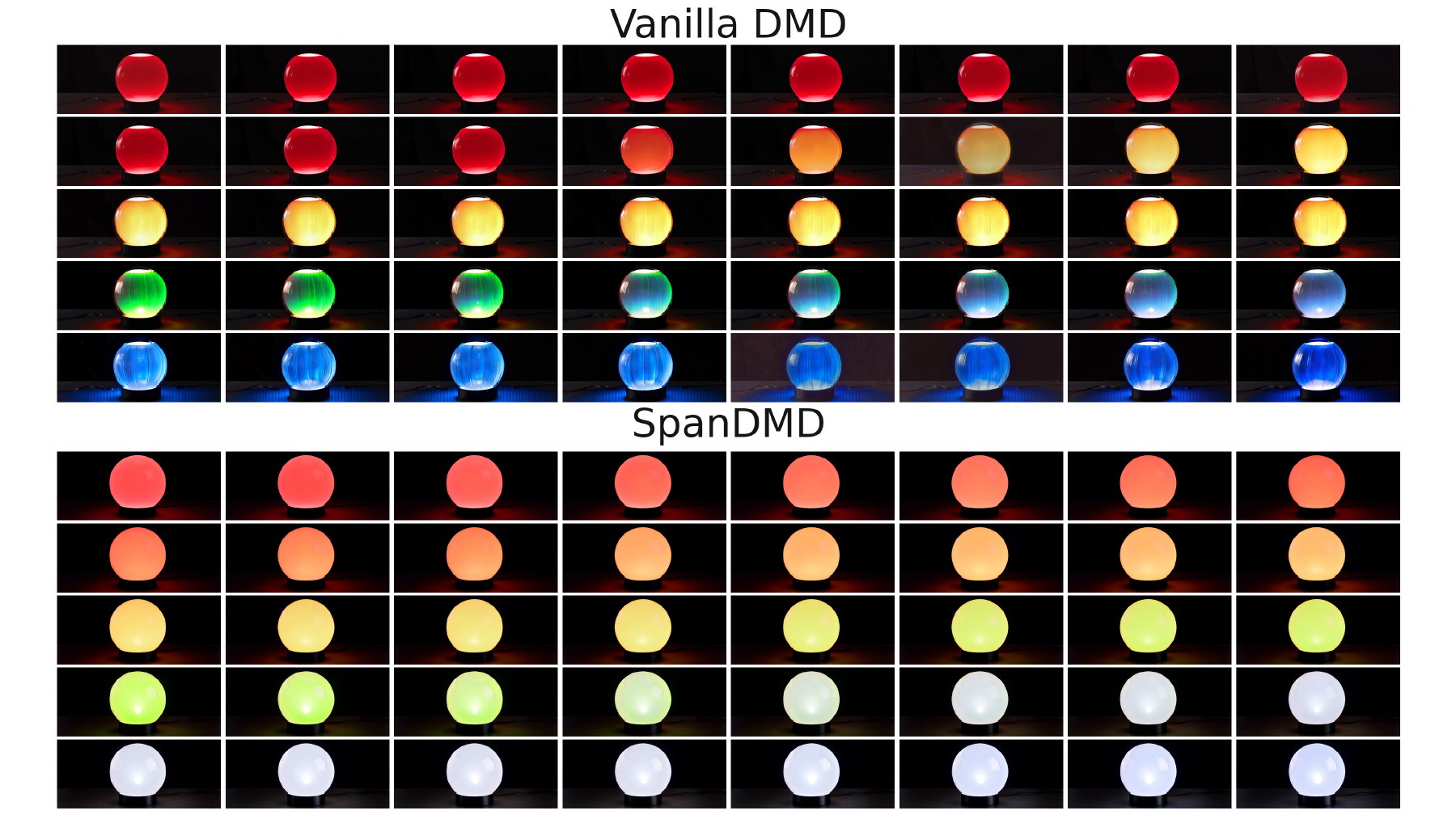}
      \caption{
  Color-transition comparison between vanilla DMD and \cdmd{}. 
      }
      \label{fig:color-long}
\end{figure}

\noindent\textbf{Effect of \cdmd{}.}
The w/o \cdmd{} variant retains the student's planned prompt
schedule but uses vanilla DMD with $c^-$, all segue prompts,
and $c^+$ merged into one condition for both score networks
(Section~\ref{sec:spandmd}).
This reduces all six transition metrics, particularly OSE
and MBS. Figure~\ref{fig:main-ablation} shows interpenetration,
identity changes, and disappearing objects.
We further evaluate a color-transition task where a sphere changes from red to blue through intermediate colors.
As shown in Figure~\ref{fig:color-long}, \cdmd{} produces smoother and more timely transitions than vanilla DMD, with fewer artifacts (Supplementary Sec.~H.1).

We further compare three alternative supervision strategies:
final-prompt-only DMD, cropped-span DMD, and unmasked
per-condition DMD (Table~\ref{tab:ablation}).
\cdmd{} leads on all six metrics.
Compared with unmasked per-condition DMD, which applies
each residual across all spans, span-specific masking
improves OSE from 0.742 to 0.802 by avoiding conflicting
supervision.
Compared with cropped-span DMD, retaining full-rollout
context improves OSE from 0.755 to 0.802 and NSE
from 0.859 to 0.890, preserving cross-stage coherence.

%% file: tables/main_comparison.tex
\begin{table*}[t]
\centering
\caption{
Comparison on OPENTRANS-360.
The Overall score is the unweighted mean of the 11 metric scores.
Higher is better for all metrics.
}
\label{tab:OPENTRANS-360-results}

\resizebox{\linewidth}{!}{%
\begin{tabular}{lcccccccccccc}
\toprule
& \multicolumn{3}{c}{\textbf{Segment Quality}}
& \multicolumn{8}{c}{\textbf{Transition Quality}}
& \\
\cmidrule(lr){2-4}
\cmidrule(lr){5-12}
Method
& SPF$\uparrow$
& SMC$\uparrow$
& SPP$\uparrow$
& VBS$\uparrow$
& MBS$\uparrow$
& OSE$\uparrow$
& NSE$\uparrow$
& SP$\uparrow$
& ScP$\uparrow$
& SUF$\uparrow$
& TP$\uparrow$
& Overall$\uparrow$ \\
\midrule
\multicolumn{13}{l}{\textit{Closed-source Models}} \\
PixVerse R1
& 0.899
& 0.852
& 0.984
& 0.984
& 0.872
& \underline{0.783}
& 0.853
& 0.751
& 0.809
& 0.773
& 0.936
& 0.863 \\
HappyOyster
& \underline{0.901}
& 0.864
& \textbf{0.996}
& 0.961
& 0.861
& 0.762
& 0.854
& \underline{0.780}
& 0.840
& 0.769
& 0.941
& \underline{0.866} \\
\midrule
\multicolumn{13}{l}{\textit{Open-source Models}} \\
Self-Forcing
& 0.793
& 0.799
& 0.923
& \underline{0.993}
& 0.865
& 0.628
& 0.779
& 0.726
& 0.773
& 0.647
& 0.847
& 0.798 \\
LongLive
& 0.881
& 0.801
& 0.952
& 0.973
& 0.821
& 0.713
& 0.851
& 0.736
& \underline{0.845}
& 0.745
& 0.930
& 0.841 \\
Rolling-Forcing
& 0.758
& 0.786
& 0.939
& 0.962
& 0.805
& 0.680
& 0.734
& 0.694
& 0.799
& 0.720
& 0.831
& 0.792 \\
Deep Forcing
& 0.803
& 0.864
& 0.981
& 0.989
& \underline{0.875}
& 0.643
& 0.758
& 0.772
& 0.789
& 0.691
& 0.832
& 0.818 \\
MemFlow
& 0.856
& 0.793
& 0.975
& 0.992
& 0.860
& 0.667
& 0.805
& 0.746
& 0.819
& 0.684
& 0.853
& 0.823 \\
Causal-Forcing
& 0.712
& 0.744
& 0.920
& 0.942
& 0.794
& 0.575
& 0.722
& 0.662
& 0.698
& 0.688
& 0.879
& 0.758 \\
Anchor-Forcing
& 0.882
& \underline{0.866}
& 0.980
& 0.989
& 0.866
& 0.758
& 0.859
& 0.767
& 0.830
& 0.775
& 0.928
& 0.864 \\
SWIFT
& 0.892
& 0.813
& 0.962
& 0.990
& 0.826
& 0.781
& \underline{0.863}
& 0.765
& 0.844
& 0.764
& \underline{0.945}
& 0.859 \\
IAMFlow
& 0.821
& 0.766
& 0.988
& 0.978
& 0.868
& 0.505
& 0.776
& 0.748
& 0.754
& 0.689
& 0.855
& 0.795 \\
TempAct
& 0.842
& 0.858
& 0.971
& 0.976
& 0.769
& 0.682
& 0.815
& 0.704
& 0.795
& \underline{0.778}
& 0.911
& 0.827 \\
\midrule
Ours
& \textbf{0.909}
& \textbf{0.869}
& \underline{0.992}
& \textbf{0.995}
& \textbf{0.897}
& \textbf{0.802}
& \textbf{0.890}
& \textbf{0.783}
& \textbf{0.863}
& \textbf{0.796}
& \textbf{0.961}
& \textbf{0.887} \\
\bottomrule
\end{tabular}%
}
\end{table*}

%% file: tables/streamav_comparison_semantic.tex
\begin{table}[t]
  \centering
  \small
  \setlength{\tabcolsep}{2pt}
  \caption{Instruction adherence and interactive response metrics on
  StreamAV-Bench. Bold and underline indicate the best and second-best scores, respectively.}
  \label{tab:streamav-semantic}
  \resizebox{\columnwidth}{!}{%
    \begin{tabular}{@{}l*{6}{c}@{}}
      \toprule
      Method & VIF$\uparrow$ & VID$\downarrow$ & VUF$\uparrow$
             & PVC$\uparrow$ & PVUAR$\uparrow$ & PVRL$\downarrow$ \\
      \midrule
      PixVerse R1
        & 2.976 & 0.484 & 2.573
        & \underline{4.278} & \underline{0.350} & 12.992 \\
      HappyOyster
        & \underline{3.565} & 0.510 & \textbf{2.975}
        & 3.744 & \textbf{0.518} & 14.484 \\
      \midrule
      Self-Forcing
        & 3.019 & 0.373 & 2.237 & 3.912 & 0.185 & 14.164 \\
      LongLive
        & 3.164 & 0.317 & 2.341 & 3.631 & 0.241 & 11.792 \\
      Rolling-Forcing
        & 3.009 & 0.425 & 2.315 & 4.086 & 0.217 & 13.234 \\
      Deep Forcing
        & 3.041 & 0.359 & 2.295 & 3.879 & 0.220 & 13.433 \\
      MemFlow
        & 3.142 & 0.333 & 2.309 & 3.630 & 0.217 & 13.698 \\
      Causal-Forcing
        & 2.834 & 0.364 & 2.284 & 4.218 & 0.219 & 14.032 \\
      SWIFT
        & 3.152 & 0.305 & 2.357 & 3.615 & 0.246 & \underline{10.340} \\
      IAMFlow
        & 3.192 & \underline{0.302} & 2.319 & 3.655 & 0.222 & 11.683 \\
      \midrule
      \textbf{Ours}
        & \textbf{3.667} & \textbf{0.301} & \underline{2.894}
        & \textbf{4.581} & 0.308 & \textbf{10.311} \\
      \bottomrule
    \end{tabular}%
  }
\end{table}

%% file: tables/vbench-clip.tex
\begin{table}[t]
  \centering
  
  \small
  \setlength{\tabcolsep}{3pt}
  \renewcommand{\arraystretch}{1.15}
  \caption{VBench-Long and CLIP results on 60-second multi-prompt videos.
  CLIP scores are computed for each 10-second segment using its
  corresponding prompt. Bold and underline indicate the best and second-best results,
  respectively.}
  \label{tab:vbench-multiprompt}

  \resizebox{\linewidth}{!}{%
  \begin{tabular}{l*{10}{c}}
    \toprule
    \multirow{2}{*}{Method}
    & \multirow{2}{*}{\shortstack{Aesthetic\\Quality$\uparrow$}}
    & \multirow{2}{*}{\shortstack{Imaging\\Quality$\uparrow$}}
    & \multirow{2}{*}{\shortstack{Motion\\Smooth.$\uparrow$}}
    & \multirow{2}{*}{\shortstack{Temporal\\Flicker.$\uparrow$}}
    & \multicolumn{6}{c}{CLIP Score$\uparrow$} \\
    \cmidrule(lr){6-11}
    & & & & &
    0--10\,s & 10--20\,s & 20--30\,s &
    30--40\,s & 40--50\,s & 50--60\,s \\
    \midrule
    Self-Forcing
    & 58.25 & 71.02 & 98.53 & 97.40
    & 25.44 & 24.77 & 23.41 & 22.14 & 21.88 & 21.07 \\
    LongLive
    & 59.91 & \textbf{71.79}
    & 98.89 & \underline{98.23}
    & 25.67 & 25.35 & \underline{24.49}
    & \underline{24.31} & 23.28 & 23.10 \\
    MemFlow
    & 60.05 & 71.66 & 98.73 & 97.62
    & 25.72 & 24.88 & 23.70 & 23.64 & 23.43 & 23.26 \\
    Anchor Forcing
    & \underline{60.11} & 71.48 & 98.79 & 98.22
    & 25.59 & 25.27 & 24.40 & 24.22 & 23.37 & 23.63 \\
    SWIFT
    & 60.04 & 71.34 & \underline{99.01} & 98.19
    & 25.66 & 25.32 & 24.45 & 23.92
    & \underline{24.48} & \underline{23.77} \\
    IAMFlow
    & 59.94 & 71.43 & 98.88 & 97.80
    & 25.90 & \underline{25.50} & 24.33
    & 23.61 & 23.59 & 23.09 \\
    TempAct
    & 59.87 & 71.09 & 98.40 & 97.98
    & \textbf{25.98} & 25.12 & 24.21
    & 23.83 & 23.70 & 23.18 \\
    \midrule
    \textbf{Ours}
    & \textbf{60.18}
    & \underline{71.74}
    & \textbf{99.26}
    & \textbf{98.52}
    & \underline{25.92}
    & \textbf{25.86}
    & \textbf{25.20}
    & \textbf{24.89}
    & \textbf{24.91}
    & \textbf{24.36} \\
    \bottomrule
  \end{tabular}%
  }
\end{table}

%% file: tables/ablation.tex
\begin{table}[t]
  \centering
  \footnotesize
  \setlength{\tabcolsep}{2pt}
  \caption{Ablations on six selected transition metrics of \bench{}.
  Top: components of \method{}.
  Middle: alternative DMD supervision strategies.
  Bottom: the planner added to existing generators without retraining.}
  \label{tab:ablation}
  \resizebox{\columnwidth}{!}{%
  \begin{tabular}{@{}lcccccc@{}}
  \toprule
  Method
  & VBS$\uparrow$
  & MBS$\uparrow$
  & OSE$\uparrow$
  & NSE$\uparrow$
  & SUF$\uparrow$
  & TP$\uparrow$ \\
  \midrule
  \rowcolor{orange!12}
  \method{} (full)
  & 0.995 & 0.897 & 0.802 & 0.890 & 0.796 & 0.961 \\
  w/o TD-Schema
  & 0.989 & 0.881 & 0.758 & 0.885 & 0.788 & 0.954 \\
  w/o \cdmd{}
  & 0.988 & 0.869 & 0.759 & 0.865 & 0.790 & 0.950 \\
  w/o \cdmd{} \& planner
  & 0.976 & 0.824 & 0.717 & 0.861 & 0.747 & 0.931 \\
  \midrule
  \rowcolor{orange!12}
  \cdmd{} (\method{})
  & 0.995
  & 0.897
  & 0.802
  & 0.890
  & 0.796
  & 0.961 \\
  Final-prompt-only
  & 0.986
  & 0.866
  & 0.749
  & 0.864
  & 0.785
  & 0.945 \\
  Cropped-span
  & 0.983
  & 0.871
  & 0.755
  & 0.859
  & 0.783
  & 0.951 \\
  Unmasked
  & 0.982
  & 0.862
  & 0.742
  & 0.853
  & 0.768
  & 0.933 \\
  \midrule
  Self-Forcing
  & 0.993 & 0.865 & 0.628 & 0.779 & 0.647 & 0.847 \\
  \quad + planner
  & 0.995 & 0.877 & 0.684 & 0.829 & 0.721 & 0.880 \\
  LongLive
  & 0.973 & 0.821 & 0.713 & 0.851 & 0.745 & 0.930 \\
  \quad + planner
  & 0.986 & 0.853 & 0.752 & 0.864 & 0.775 & 0.945 \\
  \bottomrule
  \end{tabular}%
  }
\end{table}

%% file: sections/conclusion.tex
\section{Conclusion}

Mid-stream prompt changes specify new goals but leave
the transition from the current state implicit.
\method{} plans this transition from the realized visual
state using TD-Schema segue prompts, while \cdmd{}
provides span-specific supervision with full-rollout context.
Our method leads on all transition metrics on OpenTrans-360,
with improved instruction following and competitive visual
quality on external benchmarks.
Although planning may overlook physical prerequisites and
execution remains limited by the underlying generator
(Supplementary Sec.~C), our planner can also
improve existing generators without additional training.
Together, these results highlight the value of combining
state-grounded planning with span-specific supervision
for more faithful interactive video generation.

%% file: sections/llm_statement.tex
\section*{LLM Usage Statement}
We used Qwen3-VL-235B-A22B-Instruct to extract training segue prompts from VidProM clips and JoyAI-VL-Interaction for online transition planning. For \bench{}, Qwen2.5-72B-Instruct-AWQ generated prompt streams and InternVL3.5-38B scored videos. Qwen2-72B-Instruct generated prompts for the VBench-Long evaluation. We used GPT-6.1 to help develop and debug the experimental code and to polish the language of the manuscript. We reviewed the AI-assisted work and take responsibility for the submission.

%% file: appendix.tex
\appendix
\setcounter{figure}{0}
\setcounter{table}{0}
\setcounter{equation}{0}
\renewcommand{\thefigure}{S\arabic{figure}}
\renewcommand{\thetable}{S\arabic{table}}
\renewcommand{\theequation}{S\arabic{equation}}
\renewcommand{\theHfigure}{supp.\arabic{figure}}
\renewcommand{\theHtable}{supp.\arabic{table}}
\renewcommand{\theHequation}{supp.\arabic{equation}}

\twocolumn[{%
  \begin{center}
    {\LARGE\bfseries Supplementary Material}\par\vspace{0.4em}
  \end{center}
  \vspace{0.8em}
}]
\suppressfloats[t]
\input{appendix_sections/appendix_corner_cutting}
\input{appendix_sections/appendix_user_study}
\input{appendix_sections/appendix_limitation}
\input{appendix_sections/streamav-bench-results}

\input{appendix_sections/appendix_sensitivity_vlm}

\input{appendix_sections/appendix_visual_grounding}
\input{appendix_sections/appendix_agent_inference}
\input{appendix_sections/spandmd-analysis}
\input{appendix_sections/OpenTrans-360-Details}

\input{appendix_sections/training-data-curation}
\input{appendix_sections/appendix_human_alignment}
\input{appendix_sections/appendix_additional_training_hyperpara}
\input{tables/metric_checklists}

%% file: appendix_sections/appendix_corner_cutting.tex
\begingroup

\section{Analysis of Corner Cutting}
\label{app:corner_cutting}

\begin{table}[t]
  \centering
  \footnotesize
  \setlength{\tabcolsep}{2pt}
  \caption{Transition quality on 100 cases over a 5-second horizon,
  with two generation seeds per case. No-switch controls receive both
  events in a fixed prompt from the start. Online methods receive the
  new event through a mid-stream prompt update. Higher is better for
  all metrics. Bold indicates the best score within the online-switching group.}
  \label{tab:short_horizon_transitions}
  \resizebox{\columnwidth}{!}{%
    \begin{tabular}{@{}l*{6}{c}@{}}
      \toprule
      Method & VBS$\uparrow$ & MBS$\uparrow$ & OSE$\uparrow$
             & NSE$\uparrow$ & SUF$\uparrow$ & TP$\uparrow$ \\
      \midrule
      \multicolumn{7}{@{}l}{\textit{Offline bidirectional generation}} \\
      Wan2.1-1.3B    & 0.991 & 0.885 & 0.791 & 0.884 & 0.807 & 0.962 \\
      Wan2.1-14B     & 0.992 & 0.896 & 0.801 & 0.892 & 0.819 & 0.966 \\
      MiniMax-H3-33B & 0.997 & 0.890 & 0.815 & 0.904 & 0.820 & 0.969 \\
      \midrule
      \multicolumn{7}{@{}l}{\textit{Autoregressive, no prompt switch}} \\
      LongLive-1.3B  & 0.993 & 0.879 & 0.786 & 0.879 & 0.795 & 0.955 \\
      MemFlow-1.3B   & 0.992 & 0.871 & 0.770 & 0.871 & 0.786 & 0.939 \\
      \midrule
      \multicolumn{7}{@{}l}{\textit{Autoregressive, online prompt switch}} \\
      LongLive-1.3B  & 0.989 & 0.831 & 0.755 & 0.859 & 0.768 & 0.943 \\
      MemFlow-1.3B   & 0.991 & 0.851 & 0.666 & 0.815 & 0.690 & 0.861 \\
      SEGUE-1.3B (Ours)
        & \textbf{0.996} & \textbf{0.895} & \textbf{0.808}
        & \textbf{0.895} & \textbf{0.814} & \textbf{0.968} \\
      \bottomrule
    \end{tabular}%
  }
\end{table}

\paragraph{Experimental setup.}
We evaluate 100 event-transition cases over a 5-second horizon with two generation seeds for each case, to examine whether corner cutting is associated with online
prompt switching beyond the generators' intrinsic limitations.
We compare three settings.
First, offline bidirectional models
(Wan2.1-1.3B, Wan2.1-14B, and MiniMax-H3-33B)
receive both events before generation in a single temporally
annotated prompt, specifying $c^{-}$ for the initial interval
and $c^{+}$ for the remaining interval.
Second, LongLive-1.3B and MemFlow-1.3B generate each video under
a fixed multi-event prompt that describes both events from the
outset, such as ``A happens, then B happens,'' without any
mid-stream prompt update.
Third, in the online setting, generation starts under $c^{-}$
and receives the update to $c^{+}$ at the nominal 1-second
boundary, with the generated prefix already committed.
LongLive and MemFlow switch directly to the new prompt,
whereas SEGUE inserts planned segue prompts.
The no-switch controls allow comparisons within the same
autoregressive generators on the same requested event sequences.
The short horizon reduces the opportunity for long-video
error accumulation.

\begin{figure*}[t]
    \centering
    \includegraphics[width=0.84\linewidth]{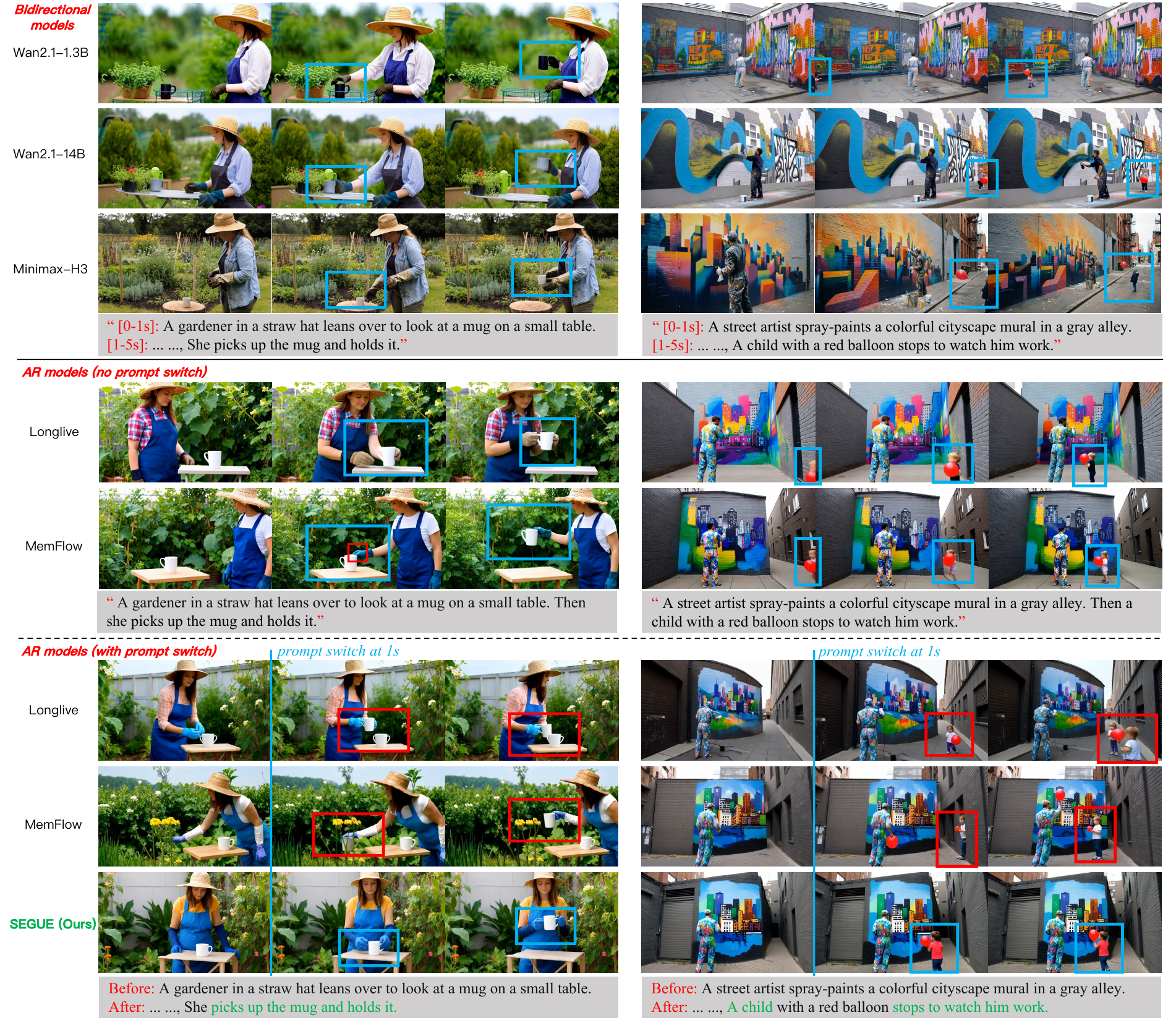}
    \caption{
        Qualitative comparison with and without online prompt switching.
        Left: a gardener picks up an existing mug.
        Right: a child carrying a red balloon approaches a street artist.
        Top: offline bidirectional models receive both events in a
        temporally annotated prompt before generation.
        Middle: LongLive and MemFlow receive a fixed multi-event
        prompt describing both events from the outset, without
        a mid-stream update.
        Bottom: LongLive, MemFlow, and SEGUE receive the new event
        through an online prompt update at the nominal 1-second boundary.
        The baseline examples exhibit mug duplication and abrupt
        child appearance under online switching, while their
        no-switch counterparts show more coherent event progression.
        SEGUE produces coherent transitions under online updates.
        Cyan boxes highlight coherent object or subject progression;
        red boxes mark artifacts.
    }
    \label{fig:short_horizon_transitions}
\end{figure*}

\paragraph{Quantitative results.}
Table~\ref{tab:short_horizon_transitions} shows that both
LongLive and MemFlow obtain higher scores on all six metrics
with fixed multi-event prompts than with online prompt switching.
For LongLive, switching lowers MBS from 0.879 to 0.831,
OSE from 0.786 to 0.755, and SUF from 0.795 to 0.768.
For MemFlow, OSE drops from 0.770 to 0.666,
NSE from 0.871 to 0.815, and SUF from 0.786 to 0.690.
These within-generator comparisons show that transition quality
deteriorates under online updates even over a short 5-second
horizon.

The bidirectional models also demonstrate strong transition
capability under offline conditioning, where both events are
specified in advance.
These results provide a complementary reference, while the
no-switch autoregressive controls establish that the same
generators used for online switching can produce more coherent
event progressions when both events are known from the outset.

Among the three methods evaluated with online prompt switching,
SEGUE achieves the highest scores on all six metrics.
Relative to direct switching in LongLive, it improves
MBS by 0.064, OSE by 0.053, NSE by 0.036, and SUF by 0.046.
Its results approach those of the offline bidirectional references
while retaining the online setting in which the target event is
provided only after the initial video prefix has been generated.

\paragraph{Qualitative results.}
Figure~\ref{fig:short_horizon_transitions} compares the three
settings on transitions involving an existing object and the
introduction of a new subject.
In the no-switch examples, LongLive and MemFlow depict the
existing mug being lifted from the table.
Under online prompt switching, however, a mug appears in the
subject's hand while another remains on the table.
Similarly, the no-switch examples show the child entering
from the edge of the frame and moving into the scene,
whereas the switched examples show the child appearing
abruptly within the scene without an observable entrance.
SEGUE depicts the existing mug being lifted and the child
walking into the scene and approaching the artist.
These examples illustrate how online updates can lead the
baselines to realize the newly requested event while skipping
the intermediate changes needed to connect it to the preceding
visual state.

\endgroup

%% file: appendix_sections/appendix_user_study.tex
\section{User Study}
\label{app:user-study}

\begin{figure*}[t]
    \centering
    \includegraphics[width=0.82\linewidth]{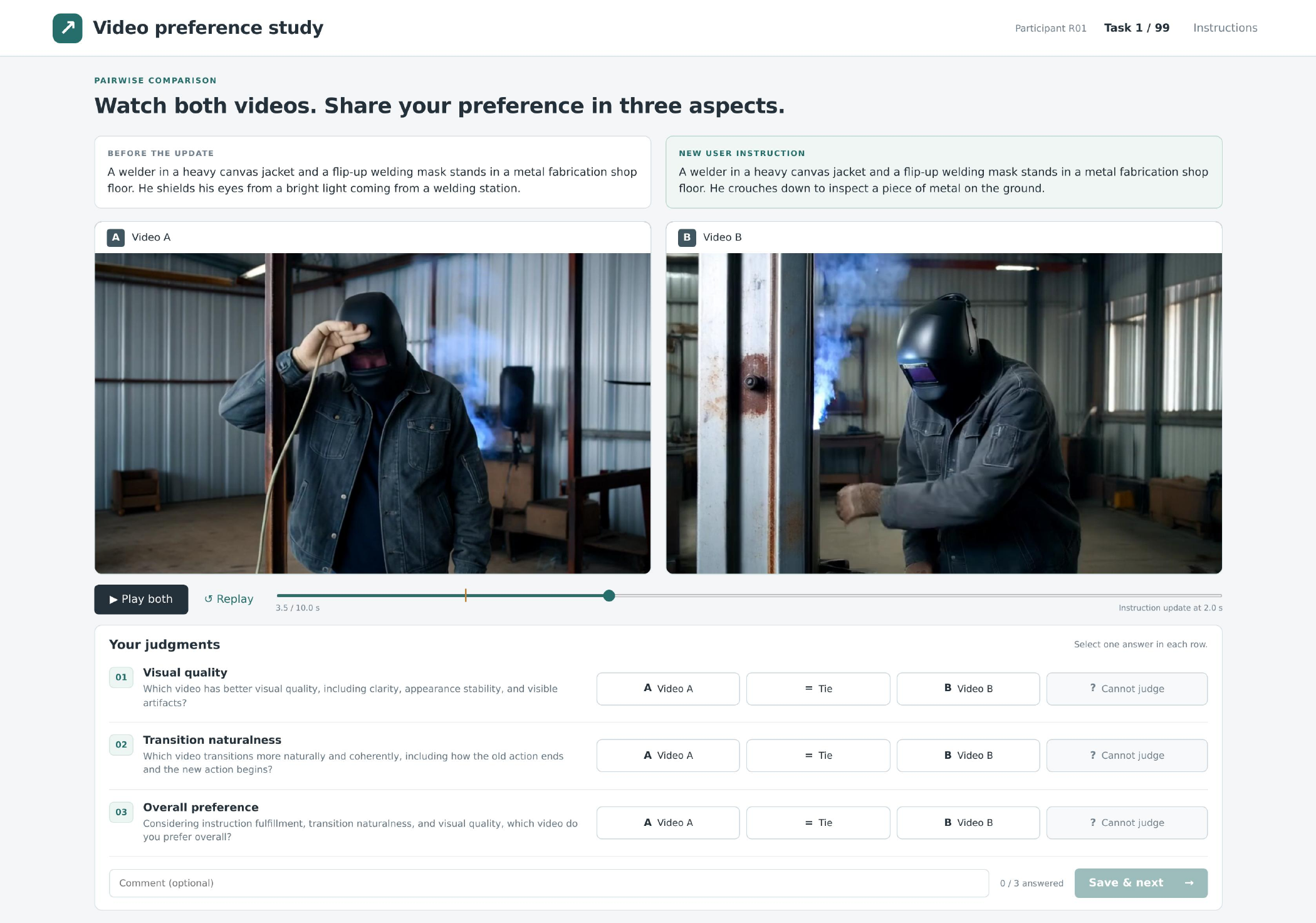}
    \caption{User-study interface. Participants compare anonymized clips under
    the same instruction update and separately judge visual quality,
    transition naturalness, and overall preference.}
    \label{fig:user-study-interface}
\end{figure*}

\begin{table}[t]
\centering
\small
\caption{Pairwise user preferences for \textsc{Segue} against each baseline.
Each entry reports the preference score (\%; ties receive half credit).
Higher scores favor \textsc{Segue}. Overall preference is evaluated separately.}
\label{tab:user-study-preferences}
\setlength{\tabcolsep}{7pt}
\resizebox{\linewidth}{!}{\begin{tabular}{lccc}
\toprule
Baseline
& \shortstack{Visual\\quality}
& \shortstack{Transition\\naturalness}
& \shortstack{Overall\\preference} \\
\midrule
Self-Forcing    & 76.7\% & 73.3\% & 80.0\% \\
LongLive        & 61.7\% & 58.3\% & 65.0\% \\
Rolling Forcing & 78.3\% & 75.0\% & 81.7\% \\
Deep Forcing    & 66.7\% & 63.3\% & 70.0\% \\
MemFlow         & 63.3\% & 60.0\% & 66.7\% \\
Causal Forcing  & 83.3\% & 80.0\% & 85.0\% \\
Anchor Forcing  & 63.3\% & 63.3\% & 65.0\% \\
SWIFT           & 68.3\% & 65.0\% & 66.7\% \\
IAMFlow         & 71.7\% & 68.3\% & 75.0\% \\
TempAct         & 63.3\% & 61.7\% & 66.7\% \\
PixVerse R1     & 53.3\% & 51.7\% & 56.7\% \\
HappyOyster     & 53.3\% & 56.0\% & 53.3\% \\
\bottomrule
\end{tabular}}
\end{table}

We conduct a blinded pairwise user study comparing \textsc{Segue} with twelve
baselines on visual quality, transition naturalness, and overall preference.
We recruit 12 participants and sample 30 cases stratified by benchmark
category. One switch per case is selected before inspecting model outputs,
with switch positions balanced across cases. All comparisons use the same
selected switches and clip windows, from 2\,s before to 8\,s after each update.

Figure~\ref{fig:user-study-interface} shows the annotation interface.
Participants view synchronized clips A and B at normal speed, together with
the preceding and updated instructions and the switch time. Method identities,
benchmark scores, and planner-generated waypoints are hidden. Task order and
left--right placement are randomized, with placement fixed across the three
questions within a task. Participants separately judge visual quality
(clarity, appearance stability, and artifacts), transition naturalness
(how the preceding event ends and the new event begins), and overall
preference (considering instruction fulfillment, transition naturalness,
and visual quality), selecting A, B, a tie, or \emph{cannot judge} for each.
Overall preference is judged directly, not averaged from the other criteria.

\begin{figure*}[t]
    \centering
    \includegraphics[width=\linewidth]{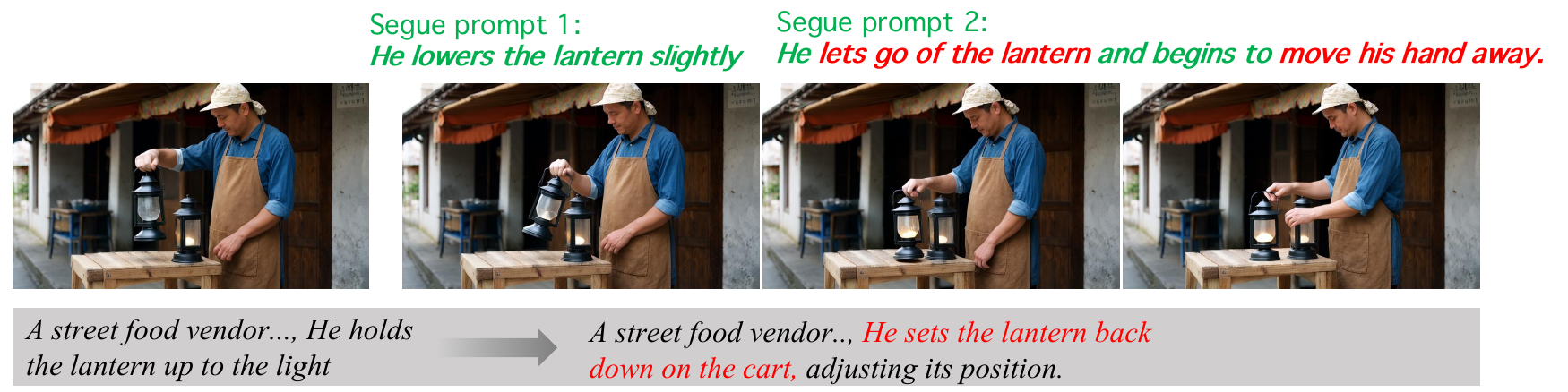}
    \caption{
    A missing support prerequisite in transition planning.
    The plan proceeds from lowering the lantern slightly
    to releasing it without explicitly establishing support
    from the cart, although the generated continuation
    largely completes the placement.
    }
    \label{fig:failure-case}
\end{figure*}

Each of the $30\times12=360$ clip pairs is evaluated by three distinct
participants, who answer all three questions, yielding 1,080 assessment tasks
and 3,240 criterion-specific responses. We add 108 hidden repeat tasks
(324 responses) for consistency checks, giving each participant 90 original
and 9 repeat tasks. For each clip pair and criterion, a majority among three
valid original responses determines the outcome; incomplete or unresolved
comparisons are excluded. Repeats do not enter the preference scores.
For baseline $b$ and criterion $c$, let $W_{b,c}$, $T_{b,c}$, and $L_{b,c}$
count resolved comparisons favoring \textsc{Segue}, tied, and favoring the
baseline, respectively. Table~\ref{tab:user-study-preferences} reports
\begin{equation}
    \begin{aligned}
      P_{b,c} &= 100\,\frac{W_{b,c}+0.5T_{b,c}}{n_{b,c}}, \\
      n_{b,c} &= W_{b,c}+T_{b,c}+L_{b,c}.
    \end{aligned}
\end{equation}

Of the 1,080 planned criterion-specific comparisons
(360 clip pairs $\times$ three criteria), 1,039 yielded a resolved
majority judgment; the remaining 41 were excluded because of
incomplete or unresolved judgments.
Of the 324 hidden repeat responses, 311 formed valid pairs with their original responses, with a repeat-consistency rate of 89\%.
Pairwise agreement among original annotators was 83\%, computed using valid original responses, including those
from comparisons without a resolved majority judgment.
Both agreement measures include ties and compare method
identities rather than screen positions.

%% file: appendix_sections/appendix_limitation.tex
\section{Limitations and Future Work}
\label{app:limitations}

\paragraph{Physical prerequisites in planning.}
Figure~\ref{fig:failure-case} illustrates a missing support
prerequisite. The planner moves from lowering the lantern
slightly to releasing it without explicitly establishing
support from the cart. Role-level ordering alone does not
ensure that such physical prerequisites are satisfied:
a supporting contact must persist until another support
is established. Although the generated continuation largely
completes the placement, the generator may be compensating
for an underspecified plan. Future work could explicitly
represent support and contact conditions and verify them
before advancing to the next stage.

\paragraph{Partial observation and execution reliability.}
The planner observes only the latest decoded frame, which
may not reveal motion history or occluded interactions.
Even a valid plan may not be faithfully executed, as physical
consistency and identity preservation remain limited by the
underlying generator. Incorporating short video histories
could improve state grounding, while feedback during
transitions could help adjust subsequent segue prompts
when execution deviates from the plan.

\paragraph{Training cost.}
\textsc{SpanDMD} requires separate score-network evaluations
for each active condition, so its distillation cost increases
with the number of supervised stages. More efficient
multi-condition score estimation and adaptive supervision
schedules could reduce this overhead while preserving
stage-specific guidance.


%% file: appendix_sections/streamav-bench-results.tex
\section{Results on StreamAV-Bench}\label{app:streamav}
\input{tables/streamav_comparison}

Table~\ref{tab:streamav_selected} reports the full results on StreamAV-Bench. Baseline results are taken from the original benchmark, while \textit{Ours} is evaluated following the same protocol.
For instruction adherence and interactive response, our method achieves the best
VIF (3.667), VID (0.301), PVC (4.581), and PVRL (10.311), ranking first on four
of the six semantic-related metrics. In particular, the PVC score improves by
7.1\% over the previous best result, indicating substantially smoother visual
transitions under prompt updates.

At the same time, the improvement in semantic transition quality does not come
at the expense of visual quality. Our method achieves 0.603 VA and 2.838 VQ,
closely matching the best results of 0.605 and 2.840, respectively. It further
achieves the lowest VA-D (0.021) and the second-lowest VQ-D (0.247), showing
strong visual stability over long-horizon generation.

%% file: tables/streamav_comparison.tex
\begin{table*}[t]
\centering
\caption{
Evaluation results on StreamAV-Bench.
The best and second-best results are highlighted in \textbf{bold}
and \underline{underlined}, respectively.
}
\label{tab:streamav_selected}

\resizebox{\textwidth}{!}{%
\begin{tabular}{lcccccccccc}
\toprule
& \multicolumn{4}{c}{\textbf{Visual Quality}}
& \multicolumn{6}{c}{\textbf{Instruction Adherence and Interactive Response}} \\
\cmidrule(lr){2-5}
\cmidrule(lr){6-11}
Method
& VA$\uparrow$
& VQ$\uparrow$
& VA-D$\downarrow$
& VQ-D$\downarrow$
& VIF$\uparrow$
& VID$\downarrow$
& VUF$\uparrow$
& PVC$\uparrow$
& PVUAR$\uparrow$
& PVRL$\downarrow$ \\
\midrule
PixVerse R1
& 0.529
& 2.428
& 0.032
& 0.253
& 2.976
& 0.484
& 2.573
& \underline{4.278}
& \underline{0.350}
& 12.992 \\
HappyOyster
& 0.529
& 2.785
& 0.032
& 0.417
& \underline{3.565}
& 0.510
& \textbf{2.975}
& 3.744
& \textbf{0.518}
& 14.484 \\
\midrule
Self-Forcing
& 0.585
& 2.753
& 0.025
& 0.323
& 3.019
& 0.373
& 2.237
& 3.912
& 0.185
& 14.164 \\
LongLive
& \textbf{0.605}
& 2.768
& 0.023
& 0.367
& 3.164
& 0.317
& 2.341
& 3.631
& 0.241
& 11.792 \\
Rolling-Forcing
& 0.572
& 2.757
& 0.043
& 0.427
& 3.009
& 0.425
& 2.315
& 4.086
& 0.217
& 13.234 \\
Deep Forcing
& 0.597
& 2.632
& 0.023
& 0.334
& 3.041
& 0.359
& 2.295
& 3.879
& 0.220
& 13.433 \\
MemFlow
& 0.600
& 2.836
& 0.028
& 0.502
& 3.142
& 0.333
& 2.309
& 3.630
& 0.217
& 13.698 \\
Causal-Forcing
& 0.556
& 2.397
& 0.030
& \textbf{0.189}
& 2.834
& 0.364
& 2.284
& 4.218
& 0.219
& 14.032 \\
SWIFT
& \textbf{0.605}
& 2.735
& \underline{0.022}
& 0.411
& 3.152
& 0.305
& 2.357
& 3.615
& 0.246
& \underline{10.340} \\
IAMFlow
& 0.602
& \textbf{2.840}
& 0.023
& 0.425
& 3.192
& \underline{0.302}
& 2.319
& 3.655
& 0.222
& 11.683 \\
\midrule
\textbf{Ours}
& \underline{0.603}
& \underline{2.838}
& \textbf{0.021}
& \underline{0.247}
& \textbf{3.667}
& \textbf{0.301}
& \underline{2.894}
& \textbf{4.581}
&  0.308
& \textbf{10.311} \\
\bottomrule
\end{tabular}%
}
\end{table*}

%% file: appendix_sections/appendix_sensitivity_vlm.tex
\section{Sensitivity to the planner VLM}
\label{app:sensitivity_vlm}
\input{tables/planner-vlm}

We compare four planner VLMs while keeping the trained generator,
TD-Schema, deterministic compiler, transition budget, and
evaluation cases fixed (Table~\ref{tab:vlm-ablation}).
The small variance in quality scores across these VLMs indicates
limited sensitivity to planner choice within the tested set.
JoyAI-VL-Interaction-INT4 ranks first or ties for first on nine
of the eleven quality metrics and achieves the lowest measured
planning latency (5th--95th percentile: $0.44$--$0.85$\,s),
motivating its use as our default planner.

%% file: tables/planner-vlm.tex
\begin{table*}[t]
    
    \centering
    \caption{
Comparison of different planner VLMs.
Latency is reported as the 5th--95th percentile range
across evaluation cases.
Quality scores vary little across the four VLMs,
with a mean per-metric population variance of approximately
$1.33 \times 10^{-5}$.
Bold indicates the best result in each column, including ties.
}
    \label{tab:vlm-ablation}
    \small
    \setlength{\tabcolsep}{3pt}
    \resizebox{\linewidth}{!}{%
    \begin{tabular}{l*{12}{c}}
        \toprule
        Planner VLM
        & \shortstack{Latency (s)$\downarrow$}
        & SPF$\uparrow$
        & SMC$\uparrow$
        & SPP$\uparrow$
        & VBS$\uparrow$
        & MBS$\uparrow$
        & OSE$\uparrow$
        & NSE$\uparrow$
        & SP$\uparrow$
        & ScP$\uparrow$
        & SUF$\uparrow$
        & TP$\uparrow$ \\
        \midrule
        JoyAI-VL-Interaction-INT4
        & \textbf{[0.44, 0.85]}
        & \textbf{0.909}
        & \textbf{0.869}
        & \textbf{0.992}
        & 0.995
        & \textbf{0.897}
        & 0.802
        & \textbf{0.890}
        & \textbf{0.783}
        & \textbf{0.863}
        & \textbf{0.796}
        & \textbf{0.961} \\
        Qwen3-VL-2B-Instruct
        & [4.81, 9.82]
        & 0.884
        & 0.861
        & 0.990
        & 0.993
        & 0.890
        & 0.794
        & 0.884
        & 0.781
        & 0.858
        & 0.791
        & \textbf{0.961} \\
        Qwen3-VL-8B-Instruct
        & [5.40, 10.63]
        & 0.904
        & 0.866
        & 0.989
        & \textbf{0.996}
        & 0.894
        & \textbf{0.805}
        & 0.887
        & \textbf{0.783}
        & 0.860
        & 0.793
        & 0.958 \\
        InternVL3.5-8B
        & [2.65, 4.58]
        & 0.899
        & 0.867
        & \textbf{0.992}
        & 0.995
        & 0.896
        & 0.800
        & 0.886
        & 0.778
        & 0.861
        & 0.789
        & 0.956 \\
        \bottomrule
    \end{tabular}%
    }
\end{table*}

%% file: appendix_sections/appendix_visual_grounding.tex
\section{Effect of Visual Grounding}
\label{app:visual-grounding}

We compare the same planner with and without the current-frame
input, keeping the generator, TD-Schema, and transition budget
fixed. Each pair receives the same prompt update and continues
from the same generated history with identical generation noise.
We evaluate five prompt switches with two seeds on each of
25 selected OpenTrans-360 cases, retaining 244 of 250 pairs
and excluding six with invalid planner outputs.
The six transition metrics are computed at the intervened
boundary using identical temporal windows and averaged first
within each case and then across cases.

As shown in Table~\ref{tab:visual-grounding}, visual grounding
improves all six evaluated metrics, with the largest relative gains in TP (4.9\%), OSE (4.2\%), and NSE (3.7\%). These improvements
indicate more coherent exits from the old state, more natural
entries into the new state, and better transition pacing,
supporting the value of conditioning the planner on the realized
visual state.

\begin{table*}[t]
    \centering
    \small
    \setlength{\tabcolsep}{5pt}
    \caption{Effect of visual grounding on six transition metrics
    across 244 matched pairs from 25 OpenTrans-360 cases.
    Each pair shares the same pre-switch history, prompt update,
    and generation noise.}
    \label{tab:visual-grounding}
    \begin{tabular}{lcccccc}
        \toprule
        Planner input
        & VBS$\uparrow$ & MBS$\uparrow$ & OSE$\uparrow$
        & NSE$\uparrow$ & SUF$\uparrow$ & TP$\uparrow$ \\
        \midrule
        Text only            & 0.984 & 0.867 & 0.756 & 0.860 & 0.774 & 0.920 \\
        Text + current frame & 0.991 & 0.890 & 0.788 & 0.892 & 0.783 & 0.965 \\
        \bottomrule
    \end{tabular}
\end{table*}

\begin{figure*}[t]
    \centering
    \includegraphics[width=0.82\linewidth]{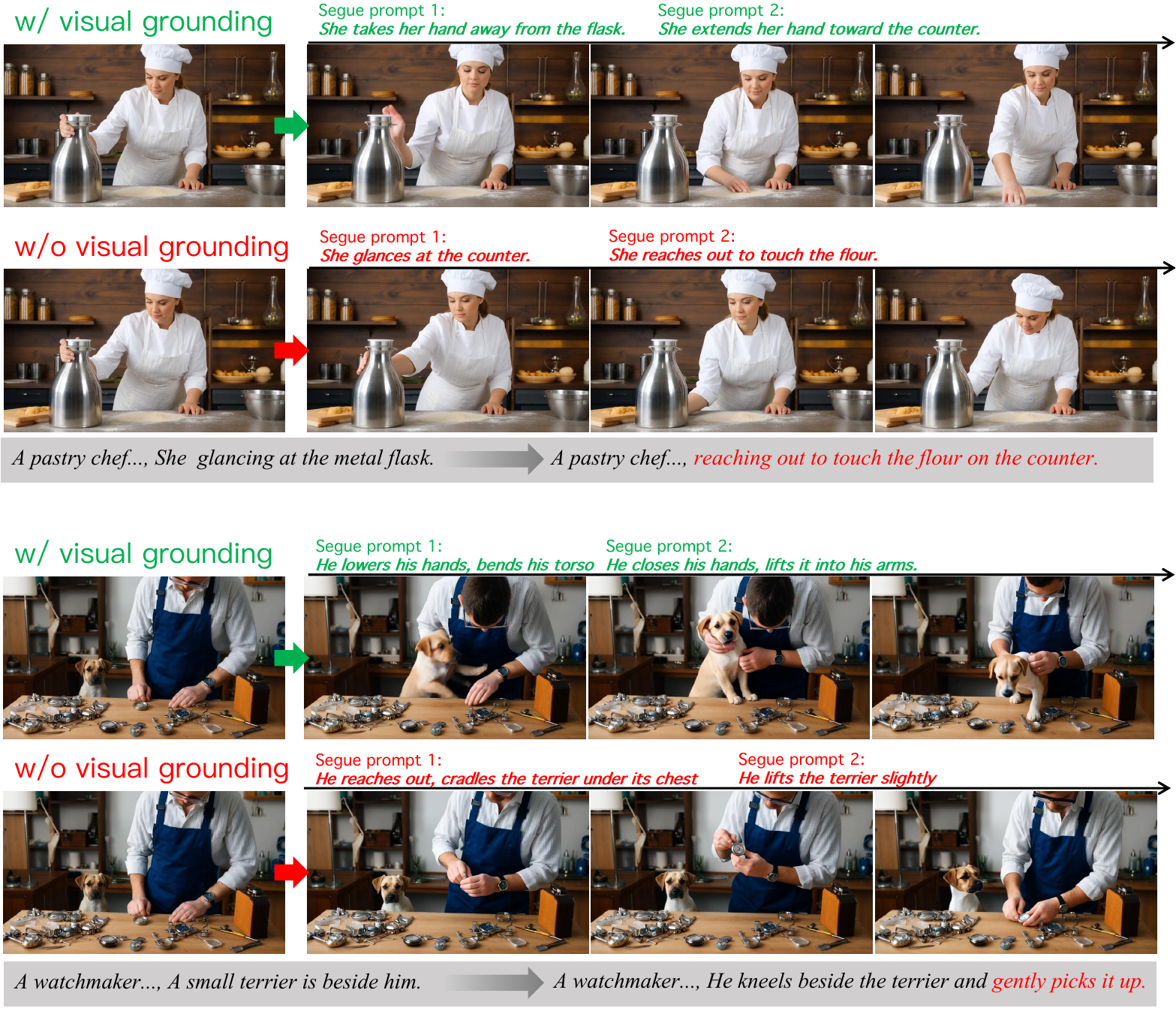}
    \caption{
    Qualitative effect of visual grounding under the same prompt
    update and pre-switch history. The grounded planner specifies
    preparatory actions based on the realized state: releasing
    the flask before reaching toward the flour (top) and bending
    toward the terrier before attempting the lift (bottom).
    The text-only plans omit these steps, and their continuations
    do not complete the requested interaction in the shown frames.
    }
    \label{fig:visual-grounding}
\end{figure*}

Figure~\ref{fig:visual-grounding} illustrates how visual input
changes the preparatory actions specified by the planner.
In the pastry-chef example, the preceding prompt describes
looking at the flask, while the realized frame also shows
a hand holding it. The grounded plan explicitly releases
this contact before reaching toward the flour; the text-only
plan omits the release, and the hand remains on the flask
in the shown frames.
In the watchmaker example, the grounded plan lowers the hands
and bends the torso toward the terrier before attempting the
lift. The text-only plan starts directly with cradling the
terrier, while its continuation remains focused on handling tools.
Together with the quantitative results, these examples suggest
that visual grounding helps the planner identify prerequisites
absent from the prompt pair and produce segue prompts that
better connect the current state to the requested action.

%% file: appendix_sections/appendix_agent_inference.tex
\providecommand{\AgentBudgetBlocks}{4}
\providecommand{\AgentMaxWaypoints}{4}

\section{Agentic Inference: Inputs, Prompts, and Execution}
\label{app:agent_inference}

The agent bridges a prompt update by reasoning about the difference between the realized visual state and the requirements of the next event. At each switch, it grounds its plan in the last generated frame and selects intermediate actions that resolve conflicting ongoing activity, release obstructing interactions, establish necessary prerequisites, or introduce missing entities. These actions prepare the transition to the target prompt while preserving the continuity of the generated scene.

\subsection{Inputs and Planning Output}
\label{app:agent_interface}

We use JoyAI-VL-Interaction as the planner. Its input consists of the last generated frame $o_{s}$, the preceding prompt $c^{-}$, the updated prompt $c^{+}$, and a transition budget $B$. The image supplies evidence of what has actually occurred, while the two prompts describe the intended semantic update. Planning is causal: the agent observes no future frames. Table~\ref{tab:agent_interface} summarizes the inputs and the resulting plan.

\begin{table}[!htbp]
\centering
\small
\caption{Inputs and outputs of transition planning.}\label{tab:agent_interface}
\begin{tabular}{@{}>{\raggedright\arraybackslash}p{0.28\linewidth}@{\hspace{0.6em}}>{\raggedright\arraybackslash}p{\dimexpr0.72\linewidth-0.6em\relax}@{}}
\toprule
Component & Description \\
\midrule
Visual observation $o_{s}$ & The final frame of the generated segment, providing evidence of ongoing actions, subject pose, physical contacts, held objects, visible entities, and spatial relations. \\
prompt update $(c^{-},c^{+})$ & The preceding and updated prompts, specifying the change in the desired event. The observed state takes precedence when it differs from the preceding prompt. \\
Transition budget $B$ & The available duration for intermediate actions before control returns to the target prompt. \\
Grounded source state & A concise description of the observed conditions relevant to planning the transition. \\
Intermediate plan & An ordered sequence of segue prompts, each specifying its role, action description, duration, and intended start and end states. \\
\bottomrule
\end{tabular}
\end{table}

\subsection{State-Dependent Constraints in TD-Schema}
\label{app:agent_execution}

TD-Schema organizes a transition around the conditions that must change before the target event can proceed coherently. It defines four preparatory roles for ending incompatible activity, removing interaction dependencies, establishing prerequisites, and introducing required entities, together with an \textsc{Execute} handoff symbol that returns control to the target prompt. Their definitions and applicability are given in Table~\ref{tab:agent_roles}.

\begin{table*}[!htbp]
\centering
\small
\caption{\textbf{Preparatory-role semantics and the \textsc{Execute} handoff in TD-Schema.} The four preparatory roles describe possible intermediate actions; \textsc{Execute} marks the handoff to $c^{+}$ and does not instantiate an additional segue prompt.}\label{tab:agent_roles}
\begin{tabular}{@{}>{\raggedright\arraybackslash}p{0.17\linewidth}>{\raggedright\arraybackslash}p{0.79\linewidth}@{}}
\toprule
Schema element & Definition and applicability \\
\midrule
\textsc{Terminate} & Concludes or safely interrupts an ongoing event when its continuation conflicts with the updated prompt. It resolves the incompatible activity through an observable completion or interruption, avoiding an unexplained discontinuity in the subject's behavior. \\
\textsc{Release} & Removes interaction dependencies that prevent the next event, such as held objects, physical contacts, occupied hands, or incompatible subject--object relations. It is invoked only for dependencies that obstruct the target event; compatible interactions can persist. \\
\textsc{Align} & Establishes prerequisites of the target event, including subject pose, spatial arrangement, interaction configuration, or viewing condition. It may comprise several distinct preparatory actions when multiple prerequisites remain unsatisfied. \\
\textsc{Entry} & Establishes required entities absent from the current visual state through a plausible observable process. It accounts for how an entity becomes available for the target event, maintaining continuity with the visible scene. \\
\textsc{Execute} & Denotes the beginning of the target event specified by $c^{+}$. It is not instantiated as an additional segue prompt; instead, it marks the handoff from transition planning back to the original target prompt. \\
\bottomrule
\end{tabular}
\end{table*}

The planner selects only preparatory roles justified by the grounded state. TD-Schema therefore specifies conditional dependencies rather than a fixed chain: an absent conflict requires no \textsc{Terminate}, an unobstructed interaction requires no \textsc{Release}, and an already visible entity requires no \textsc{Entry}. Where applicable, the role dependencies follow
\[
\begin{aligned}
\textsc{Terminate}&\;\prec\;\textsc{Release}\\
&\;\prec\;
\{\textsc{Align},\textsc{Entry}\}\;\prec\;
\underbrace{\textsc{Execute}}_{\text{handoff to }c^{+}}.
\end{aligned}
\]
Here, $\prec$ denotes precedence among instantiated preparatory roles and the final \textsc{Execute} handoff; it does not require every preparatory role to be instantiated.  \textsc{Align} may repeat to establish different prerequisites. \textsc{Align} and \textsc{Entry} occupy the same stage, allowing the order to reflect the scene: a subject may orient toward a doorway before an entity appears, or respond to that entity after its arrival.

Each segue prompt describes an observable change whose resulting state should support the next action. The first segue prompt starts from the grounded source state, and consecutive segue prompts are planned to have compatible end and start states. Subject identity, scene structure, and interactions unrelated to the requested update are preserved. In the standard setting, a transition uses one to \AgentMaxWaypoints{} intermediate segue prompts within a budget of \AgentBudgetBlocks{} generation blocks (approximately 3 seconds). Segue prompt durations divide this budget, and their descriptions condition the corresponding video spans. After the intermediate plan, the original prompt $c^{+}$ resumes control to initiate the target event.

\subsection{Planner Prompt Specification}
\label{app:agent_prompts}

Tables~\ref{tab:agent_system_prompt} and~\ref{tab:agent_user_prompt} give a condensed prompt specification of the TD-Schema formulation above. The system message defines the planning roles and continuity requirements. The user-message table presents the planning prompts applied to the inputs summarized in Table~\ref{tab:agent_interface}. The templates express the planning procedure at the method level, with implementation-specific serialization and bookkeeping omitted.

\begin{table*}[!htbp]
\centering
\small
\caption{System-message specification for grounded transition planning with TD-Schema.}\label{tab:agent_system_prompt}
\begin{tabular}{@{}>{\raggedright\arraybackslash}p{\linewidth}@{}}
\toprule
\textbf{System message} \\
\midrule
You are a transition planner for streaming video generation. Given the last generated frame, the preceding prompt, and an updated prompt, plan the intermediate actions needed to reach the conditions for the target event. \\[1pt]
\addlinespace[2pt]
Ground the source state in the image. Identify ongoing activity, held objects, physical contacts, occupied hands, subject pose, spatial relations, and visible entities. Use the observed state when it differs from the preceding prompt. Do not assume any future visual outcome. \\[1pt]
\addlinespace[2pt]
Select only preparatory roles justified by the difference between the observed state and the target event: \\[1pt]
TERMINATE: Conclude or safely interrupt an ongoing event when its continuation conflicts with the updated prompt. \\[1pt]
RELEASE: Remove interaction dependencies that prevent the target event, including obstructing held objects, contacts, occupied hands, or incompatible subject--object relations. Preserve compatible interactions. \\[1pt]
ALIGN: Establish missing prerequisites of the target event, including subject pose, spatial arrangement, interaction configuration, or viewing condition. Use separate preparatory actions when needed. \\[1pt]
ENTRY: Establish a required entity absent from the current visual state through a plausible observable process. \\[1pt]
EXECUTE: Hand control back to the original target prompt to begin its event. This is a handoff, not an additional segue prompt. Do not output an EXECUTE segue prompt or paraphrase the target event as an intermediate action. \\[1pt]
\addlinespace[2pt]
TD-Schema is not a fixed chain. Omit preparatory roles whose conditions are already satisfied. Resolve conflicting activity and obstructing interactions before the actions that depend on their resolution. ALIGN may repeat, and ALIGN and ENTRY may occur in either order when justified by the scene. \\[1pt]
\addlinespace[2pt]
Describe each segue prompt as an observable change, not merely a completed state. Begin from the grounded source state and make each segue prompt's resulting state compatible with the next action. Preserve subject identity and scene elements unrelated to the prompt update. Avoid unexplained appearances, disappearances, or changes in physical contact. \\[1pt]
\addlinespace[2pt]
Allocate the available budget among the selected intermediate actions. Return the grounded source state and an ordered plan, specifying each segue prompt's role, action description, duration, start state, and end state. Leave the target event itself to the original updated prompt. \\[1pt]
\bottomrule
\end{tabular}
\end{table*}

\begin{table}[!htbp]
\centering
\small
\caption{Planning instructions in the user message.}\label{tab:agent_user_prompt}
\begin{tabular}{@{}>{\raggedright\arraybackslash}p{\linewidth}@{}}
\toprule
\textbf{Planning instructions} \\
\midrule
1. Ground the state. Describe what the image shows that matters for the target event, including unfinished actions, interaction dependencies, subject pose, spatial configuration, and available entities. \\[1pt]
\addlinespace[2pt]
2. Identify unmet conditions. Determine whether an ongoing event conflicts with the update, whether an interaction obstructs the target, which prerequisites remain unsatisfied, and whether a required entity is absent. Do not introduce a transition action without a corresponding need in the observed state. \\[1pt]
\addlinespace[2pt]
3. Construct the intermediate plan. Select the necessary TERMINATE, RELEASE, ALIGN, and ENTRY actions and order them according to their dependencies. Each action should start from the preceding action's resulting state and move toward the conditions needed for the target event. \\[1pt]
\addlinespace[2pt]
4. Allocate durations. Distribute the transition budget among the intermediate actions. For each segue prompt, provide its role, action description, duration, and brief start and end states. \\[1pt]
\addlinespace[2pt]
After the intermediate plan, return control to the original target prompt. EXECUTE denotes this handoff and must not appear as an additional segue prompt. \\[1pt]
\bottomrule
\end{tabular}
\end{table}

\subsection{Deterministic Compilation and Failure Handling}
\label{app:agent_compiler}

The compiler converts the planner output into a block-level
conditioning schedule through dependency ordering, action
merging, validation, and duration normalization.
Semantic prerequisites and physical plausibility are assessed
by the visually grounded planner; the compiler enforces
structural and temporal constraints without a separate
symbolic simulator.

\paragraph{Dependency ordering and action merging.}
After parsing the output and normalizing role names, the
compiler checks for recognized roles, nonempty action
descriptions, and positive integer block durations.
It orders the proposed operations according to the precedence
constraints in Table~\ref{tab:agent_roles}.
\textsc{Align} and \textsc{Entry} share a precedence level,
so their relative order follows the planner's proposal
when either order is admissible.
An ordering violation in the proposed sequence is therefore
handled by reordering rather than immediate rejection.

The compiler also merges compatible operations that can be
realized by the same observable change into a single segue
prompt, preserving the required changes and their dependencies.
Distinct preparatory actions remain separate; in particular,
\textsc{Align} may repeat when different prerequisites require
different actions.
The compiled plan is then checked for the allowed segue
prompt count, role repetitions, and dependency consistency.
Plans that remain inadmissible after ordering and merging
are rejected.

\paragraph{Budget normalization and scheduling.}
Let $K$ be the number of segue prompts after ordering and
merging, $d_i$ their associated positive integer block durations,
and $B$ the maximum available transition budget.
A plan with $K>B$ is rejected because each segue prompt
requires at least one block.
Durations are retained when $\sum_i d_i \le B$.
Only over-budget plans, with $\sum_i d_i>B$, undergo
duration normalization. For these plans, the compiler computes
\[
  \tilde d_i =
  \max\!\left(
      1,\,
      \operatorname{round}
      \left(\frac{B d_i}{\sum_j d_j}\right)
  \right).
\]
It assigns the residual $B-\sum_i\tilde d_i$ to the longest
rescaled segue prompt, breaking ties by the compiled order
and maintaining a minimum duration of one block.
For an over-budget plan, if this adjustment cannot produce
positive integer durations summing to $B$, the plan is rejected.
Duration normalization operates on the compiled actions
without further deleting or merging them.
An accepted plan is expanded into a block-level schedule
by repeating each segue prompt for its assigned duration.
Control returns to the unchanged target prompt $c^{+}$
immediately after the resulting schedule ends, even when
its total duration $T_{\Pi}$ is less than $B$.

\paragraph{Rejected and empty plans.}
Malformed outputs, empty segue prompt lists, unresolved
structural violations, and unsuccessful budget normalization
trigger the same fallback: the updated prompt $c^{+}$
immediately conditions all reserved transition blocks.
The implementation represents this fallback internally as
a single \textsc{Execute} record; it introduces no additional
intermediate action or transition-stage delay.
An empty plan invokes this fallback rather than serving as
a verified claim that no preparation is required.

The system does not automatically re-query the planner
after rejection.
Planner inference or communication failures use the same
fallback, and the reason is recorded in the plan log.
As a final defensive check, an oversized schedule is
truncated to $B$ blocks.
For an accepted plan with $T_{\Pi}<B$, any remaining reserved
blocks are conditioned on $c^{+}$ and do not extend the
segue-prompt schedule.

%% file: appendix_sections/spandmd-analysis.tex
\section{Analysis of \cdmd{}}
\label{app:spandmd}

\paragraph{Training details.}
We train on 60-second self-generated rollouts, applying
DMD supervision every 5 seconds (12 evaluations per rollout).
Each evaluation uses a 7-block window (21 latent frames,
corresponding to 81 decoded frames), with overlapping
windows keeping the 14B teacher within its native temporal
context.
One selected switch-containing window per rollout uses
multi-condition \cdmd{} supervision, while the remaining
11 windows use single-condition DMD.
For each span condition $u_\ell$, both score networks
observe the complete noised window, while the corresponding
DMD residual is retained only within its assigned span.
The fake-score network is trained on student-generated
rollouts using span-conditioned flow-matching losses,
masked to each span and averaged over nonempty supervised
spans.
We set $\lambda_\ell=1$ and retain per-span DMD gradient
normalization.
The switch window requires $L=2$--$5$ guided teacher
evaluations ($2L$ forward passes with classifier-free
guidance), yielding a modest overall teacher-evaluation
cost of $(11+L)/12=1.08$--$1.33\times$ relative to
single-condition DMD.
\cdmd{} introduces no additional inference-time cost.

\paragraph{Merged-prompt DMD ablation.}
For \emph{w/o SpanDMD}, we keep the planner-generated waypoint
schedule and the student's span-wise prompt conditioning unchanged.
We concatenate all conditions in temporal order into a single teacher
prompt,
$\tilde{c}=c^{-}\Vert w_1\Vert\cdots\Vert w_K\Vert c^{+}$,
where $\Vert$ denotes text concatenation.
Vanilla DMD conditions both the real-score and fake-score networks
on $\tilde{c}$ and applies their residual to the supervised rollout
without span-specific masks.

\paragraph{Conditional-score view.}
Let $p_{t,\ell}$ be the teacher's density over the complete noised rollout $y$ under prompt $u_\ell$, and write $y=(y_{I_\ell},y_{-I_\ell})$. Since $\log p_{t,\ell}(y)=\log p_{t,\ell}(y_{I_\ell}\mid y_{-I_\ell})+\log p_{t,\ell}(y_{-I_\ell})$ and the second term does not depend on $y_{I_\ell}$,
\begin{equation}
  \big[\nabla_{y}\log p_{t,\ell}(y)\big]_{I_\ell}
  =\nabla_{y_{I_\ell}}\log p_{t,\ell}\big(y_{I_\ell}\mid y_{-I_\ell}\big).
  \label{eq:conditional-score}
\end{equation}
The same identity holds for the fake score. The residual that \cdmd{} keeps on $I_\ell$ is therefore the difference between the student's and the teacher's conditional scores of the span given the rest of the rollout. With exact scores and the rest of the rollout held fixed, the corresponding update moves the span toward the teacher's conditional distribution under $u_\ell$, rather than toward the distribution of an isolated clip. This view requires neither independent stages nor a single teacher distribution over the whole schedule. The context $y_{-I_\ell}$ contains other stages and is therefore not typical of clips described by $u_\ell$ alone, so the teacher's conditional score is an approximation; the cropped alternative in Sec.~3.3 of the main paper avoids this mismatch but loses the context entirely.

\paragraph{Why use span masks?}
A natural alternative retains separate score evaluations for each condition but removes the temporal masks, applying every condition's residual to all supervised frames. This can introduce conflicting supervision when adjacent prompts describe incompatible states, such as holding versus releasing an object.

To isolate this effect, we evaluate \emph{unmasked per-condition DMD} in Table~5 (component and supervision ablations) of the main paper, keeping the same initialization, student schedule, per-condition conditioning, and full-rollout inputs as \cdmd{}, but removing span-specific residual assignment. The corresponding fake-score supervision is also applied to all supervised frames. Compared with this alternative, \cdmd{} improves OSE from 0.742 to 0.802 and MBS from 0.862 to 0.897, demonstrating the benefit of assigning each condition's supervision to its corresponding span while retaining full temporal context.

\subsection{Color-Sequence Diagnostic}
\label{app:spandmd-color}
\begin{figure}[h]
    \centering
    \includegraphics[width=\linewidth]{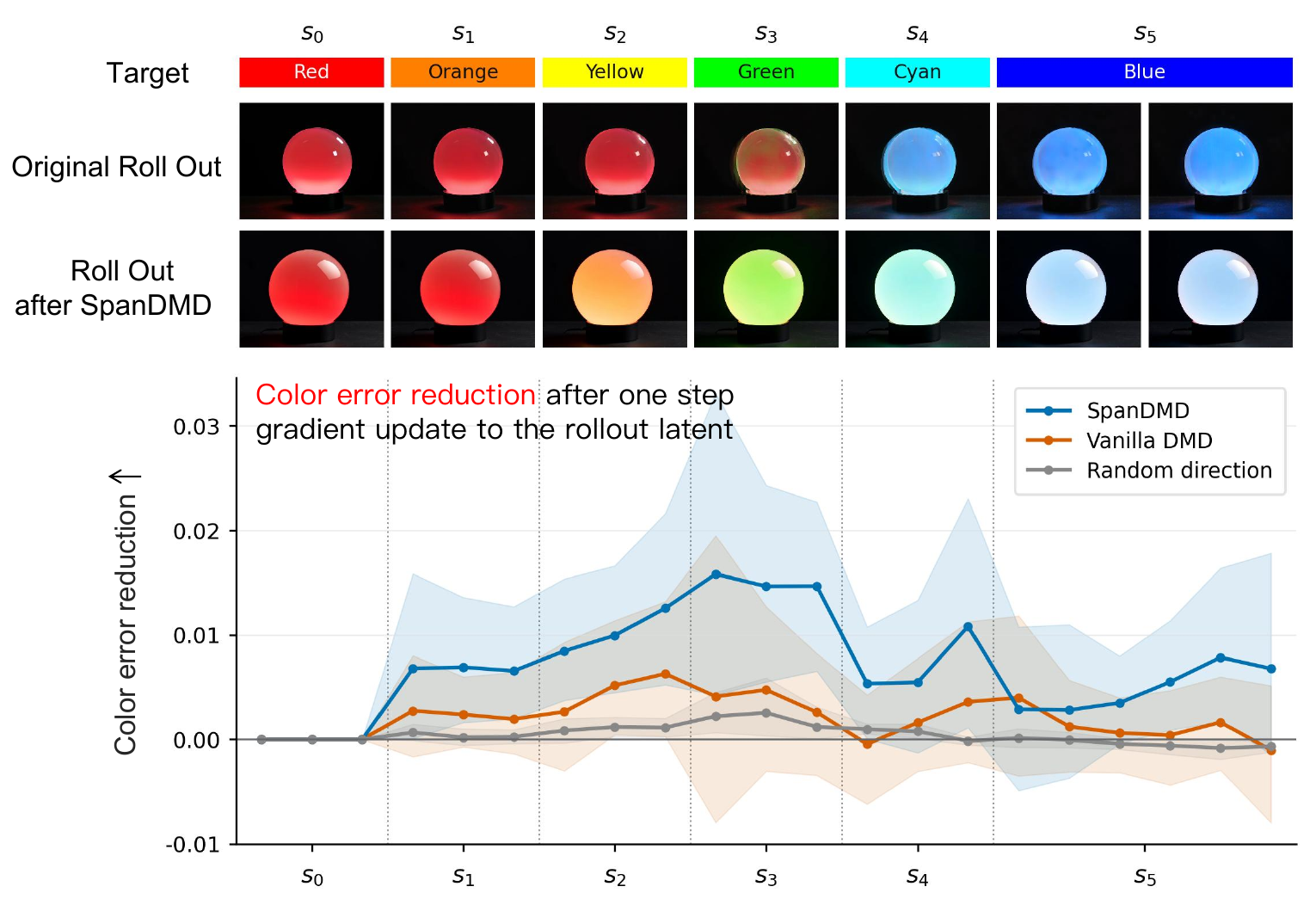}
    \caption{
Color-sequence evaluation over 5-second videos.
\textbf{Top:} Qualitative examples of generator outputs
before and after SpanDMD training, shown alongside the
target color of each span.
\textbf{Bottom:} A latent-update diagnostic measuring
mean color-error reduction after one equal-magnitude
update over 200 videos, starting from shared initial
latents with all model parameters frozen.
Shaded bands show pointwise 95\% paired hierarchical
bootstrap confidence intervals.
}
    \label{fig:color}
\end{figure}

To examine how \cdmd{} guides intermediate state transitions,
we design a 5-second color-sequence task in which a sphere
must change from red to blue through orange, yellow, green,
and cyan, following a prescribed temporal schedule
(Figure~\ref{fig:color}).
The top panel provides a qualitative comparison of
generator outputs before and after \cdmd{} training;
the bottom panel presents the latent-update diagnostic
described below.
Vanilla DMD receives the entire color sequence and its timing in one prompt, whereas \cdmd{} uses the color prompt of each span. Starting from the same generated videos, we freeze the models, apply one latent update of equal magnitude for each method, and measure how much closer the colors move to their targets.
Across 50 color-sequence plans with four rollout seeds per plan
(200 videos in total), the \cdmd{} update reduces the color error throughout $s_1$--$s_5$ and exceeds the vanilla DMD update at most sampled times, especially in the intermediate stages $s_1$--$s_4$, while a random update stays close to zero. \cdmd{} thus guides the rollout toward the intermediate targets as well as the final one.

We measure color error in normalized RGB chromaticity.
For each original decoded frame, we construct a soft spatial
weight map proportional to the product of pixel saturation,
brightness (the maximum RGB channel), and a Gaussian centered
on the image, with horizontal and vertical standard deviations
of $0.28W$ and $0.35H$, respectively.
The weights are normalized to sum to one and held fixed across
all methods and before/after comparisons; if the weighted mass
is below $10^{-6}$, we use the Gaussian alone.
This emphasizes the central colored object without requiring
a segmentation model.
Given these weights $w_t(p)$, we compute
\begin{equation}\begin{aligned}
    \mathbf{r}_t &= \sum_p w_t(p)\mathbf{I}_t(p),
    \\
    \mathbf{c}_t &=
    \frac{\mathbf{r}_t}
    {\max(\mathbf{1}^{\top}\mathbf{r}_t,10^{-8})},
    \\
    E_t &= \|\mathbf{c}_t-\mathbf{c}^{*}_t\|_2^2,
\end{aligned}\end{equation}
where RGB intensities lie in $[0,1]$ and $\mathbf{c}^{*}_t$
is the prescribed span color normalized by its channel sum.
The RGB targets for red, orange, yellow, green, cyan, and blue
are $(1,0,0)$, $(1,0.5,0)$, $(1,1,0)$, $(0,1,0)$,
$(0,1,1)$, and $(0,0,1)$, respectively.
The plotted quantity is $E_t^{\mathrm{before}}-
E_t^{\mathrm{after}}$, so positive values indicate improvement.
We evaluate color error at decoded frames aligned with
the latent temporal positions.

For this diagnostic, all methods start from the same latent
video and use an identical update mask.
Each method's latent direction is rescaled so that the RMS
displacement over the selected entries equals $\alpha$ times
the original latent RMS over those same entries, with
$\alpha=0.10$.
Each method applies a single update from the original latent video.
This matches the total update magnitude across methods
without constraining how the correction is distributed
across temporal spans.
DMD directions are evaluated at diffusion timestep $950$,
using two shared noise draws per video.
The random control uses a Gaussian direction subject to
the same mask and RMS normalization.
All model parameters remain frozen; this experiment measures
local latent correction, not a model-parameter optimization step.

We first average the two noise draws within each video,
then construct 10,000 paired hierarchical bootstrap replicates.
For each replicate, we sample 50 color-sequence plans with
replacement and, within each selected plan, sample four rollout
seeds with replacement.
The same resampling indices are used across methods and
temporal positions.
Shaded bands are the pointwise 2.5th and 97.5th percentiles
of the bootstrap mean, rather than simultaneous confidence
bands over the entire trajectory.

We further compare Vanilla DMD and \cdmd{} over
20-second color-transition sequences
(Figure~7), with approximately
3--4.5 seconds per post-switch color span.
Both models follow identical span-wise prompts and
switching schedules, with matched initial and denoising
noise.
Vanilla DMD eventually reaches some target colors but
still exhibits delayed responses and transient dimming
or flicker.
In contrast, \cdmd{} produces more gradual transitions
through intermediate hues with more stable appearances.


%% file: appendix_sections/OpenTrans-360-Details.tex
\providecommand{\bench}{\textsc{OpenTrans-360}}

\section{Benchmark Design, Construction, and Evaluation}
\label{app:benchmark}

\bench{} evaluates semantic transitions under runtime prompt updates. As summarized in Table~\ref{tab:benchmark_comparison}, general video-generation benchmarks primarily assess the quality and semantic correctness of generated content, while long-form and streaming benchmarks extend evaluation to prompt updates, state retention, and boundary continuity. Our benchmark examines the process connecting the state already generated to the newly requested event: whether the preceding activity is resolved, the transition is plausible, necessary prerequisites are established, and the new event begins coherently. These aspects complement prompt fulfillment and visual smoothness, which alone do not establish that the transition is semantically valid.

\input{tables/benchmark_comparison}

\subsection{Data Distribution}
\label{app:benchmark_distribution}

\providecommand{\bench}{\textsc{OpenTrans-360}}
\begin{figure*}[t]
\centering
\includegraphics[width=\textwidth]{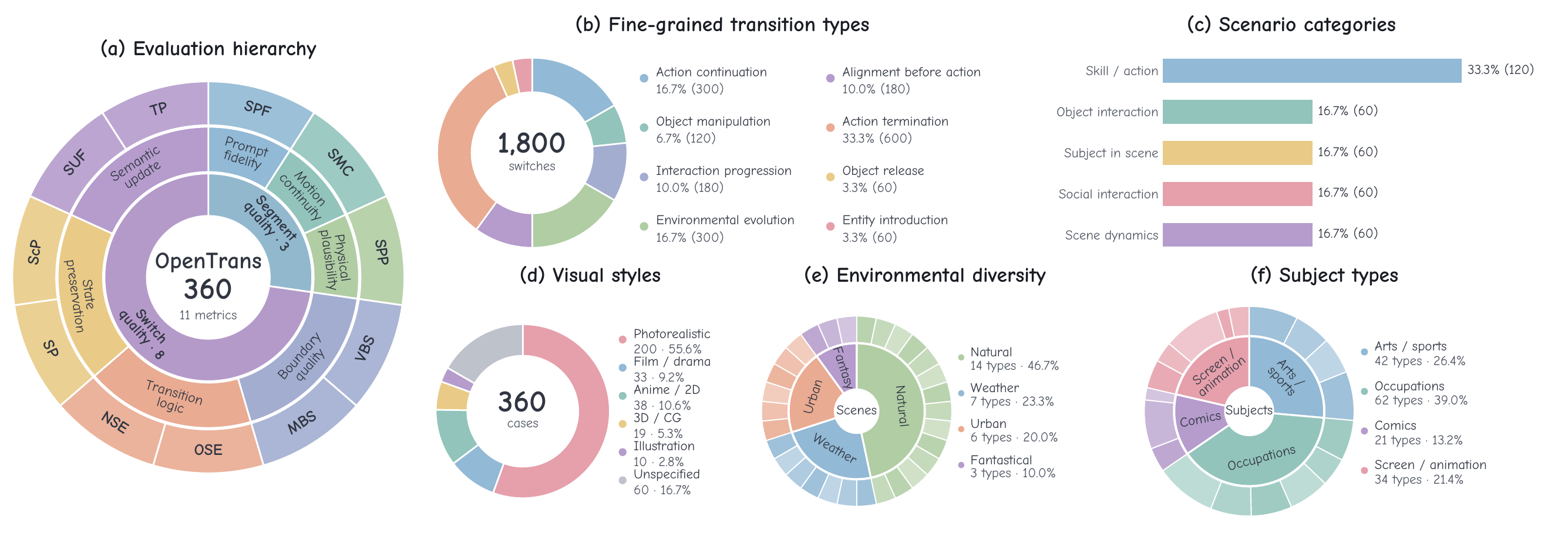}
\caption{Overview of \bench{}.
(a) Evaluation hierarchy.
(b) Eight design-derived semantic transition types.
(c) Scenario categories.
(d) Visual styles.
(e) Environmental diversity.
(f) Subject-type diversity.
Transition statistics cover 1,800 prompt switches; scenario and style statistics cover 360 cases.
Environmental and subject-type statistics count distinct pool entries, not cases.
In (e)--(f), inner sectors indicate broad families, while outer sectors represent individual scenes and subject subcategories, respectively.}
\label{fig:benchmark_diversity}
\end{figure*}

As shown in Figure~\ref{fig:benchmark_diversity}, \bench{} comprises 360 cases and 1,800 prompt switches, covering eight design-derived transition types. Each case contains six sequential prompts, active for 10 seconds each, yielding a 60-second video with five updates. The three broad construction categories in Sec.~4 (\bench{}) of the main paper comprise five scenario templates: human-centric (120 skill/action and 60 subject-in-scene cases), object-centric (60 object-interaction cases), and compositional (60 social-interaction and 60 subject-free scene-dynamics cases). The subject-free templates coordinate changes among elements of a persistent environment. Overall, the cases include 200 real-world subjects, 100 fictional characters, and 60 subject-free scenarios, giving a 2:1 real-to-fictional ratio among subject-bearing cases.

We constructed the pools by giving an LLM predefined category ranges and asking it (Qwen2.5-72B-Instruct-AWQ~\cite{qwen2025qwen25technicalreport}) to generate entries within each category. The construction pools contain 55 fictional characters, 42 skill-oriented and 62 additional occupation entries, 30 props, 30 secondary entities, and 30 subject-free scenes. Real-world subjects have occupation-specific appearances, while fictional characters span cinematic, animated, and illustrated styles. Secondary entities include people, dogs, cats, birds, and other animals. Subject-free scenes cover natural landscapes, weather, urban environments, and fantastical settings. Balanced allocation cycles through shuffled pools to distribute coverage across entries, and recurring subjects are assigned alternative settings where available.

The 360-case design comprises 2,160 prompts and 1,800 prompt switches. To characterize transition coverage, we subdivide switches using their expected step annotations and scenario tags into eight mutually exclusive types: action continuation (16.7\%), object manipulation (6.7\%), interaction progression (10.0\%), environmental evolution (16.7\%), alignment before action (10.0\%), action termination (33.3\%), object release (3.3\%), and entity introduction (3.3\%). Preparatory steps identify the latter four types; switches requiring only execution are grouped by scenario, with skill and subject-in-scene cases combined as action continuation. These labels describe the construction templates, not observed transitions in generated videos. The benchmark targets updates within a persistent setting and does not include scene cuts.

\subsection{LLM-Assisted Data Construction}
\label{app:benchmark_construction}

We first assemble structured case specifications by combining a scenario category with a subject and appearance, a compatible setting, and category-specific objects or secondary entities. These specifications determine a six-stage event outline. For example, an object-interaction sequence places a prop in the scene before it is handled and eventually released; a social-interaction sequence starts with the main subject alone, introduces another entity in the third segment, and continues with their interaction. Subject-free sequences describe environmental changes in one persistent scene. The benchmark focuses on continuity-preserving updates within a setting.

Qwen2.5-72B-Instruct-AWQ expands each specification into a shared identity-and-scene description and six short, present-tense action captions. Each final prompt combines the shared description with its segment-specific action, keeping persistent context explicit across updates. For subject-free cases, the fixed scene description supplies this shared context. The generation prompt requests visible events, consistent appearance, and plausible entity arrival, while avoiding repeated introductions and unnecessary scene changes.

Automatic validation checks the six-caption structure, caption length, scene compatibility, and category-specific constraints. For instance, object-interaction cases must retain their assigned prop, and social cases must depict an explicit arrival followed by interaction without repeatedly introducing the same entity. Invalid candidates are regenerated with the rejection reasons supplied as feedback, for up to three attempts in total. The accepted prompts are stored with their scenario metadata. Expected transition annotations are derived from the event templates; the visual evaluator scores the generated frames and prompts without requiring a model to reproduce a prescribed waypoint chain.

\paragraph{Construction prompt.}
Table~\ref{tab:benchmark_construction_prompt} gives a condensed template; the event outline and constraints are instantiated for the selected category.

\begin{table}[!htbp]
\centering
\small
\caption{Condensed Qwen prompt for benchmark construction.}\label{tab:benchmark_construction_prompt}
\begin{tabular}{@{}>{\raggedright\arraybackslash}p{\linewidth}@{}}
\toprule
\textbf{Prompt template} \\
\midrule
\textbf{System message} \\[1pt]
Write short, concrete video captions describing visible events. Use present-tense sentences without metaphors, camera jargon, or narration of intentions and emotions. Return only the requested JSON object. \\[1pt]
\midrule
\textbf{User message} \\[1pt]
Subject and appearance: \texttt{\{subject\}}. Visual style: \texttt{\{style\}}. Setting: \texttt{\{scene\}}. Category: \texttt{\{category\}}. Relevant prop or secondary entity: \texttt{\{entity\}}. \\[1pt]
\addlinespace[2pt]
Construct six consecutive steps, each lasting approximately 10 seconds, following this event outline: \texttt{\{category\_outline\}}. Preserve the assigned appearance and scene. Describe a distinct visible event in each step. Where a secondary entity is introduced, show it arriving and keep it present for subsequent interaction. \\[1pt]
\addlinespace[2pt]
Return a JSON object with \texttt{scene}, \texttt{identity}, and six \texttt{actions}. The identity sentence describes the persistent subject and setting; each action refers back to the subject with a pronoun. For subject-free scenarios, return only six actions describing the supplied scene. \\[1pt]
\bottomrule
\end{tabular}
\end{table}

\subsection{Metric Definitions and Computation}
\label{app:benchmark_metrics}

We evaluate three segment-quality metrics (SPF, SMC, and SPP) and eight switch-quality metrics (VBS, MBS, OSE, NSE, SP, ScP, SUF, and TP). The five semantic-transition dimensions in Table~\ref{tab:benchmark_comparison} are conceptual aspects covered by these metrics, rather than five additional scores. OSE assesses prior-event resolution; NSE assesses preparation, reachability, and event initiation; SPP and MBS provide evidence of physical and motion plausibility; and SP, ScP, SUF, and TP characterize preservation, update fidelity, and pacing across the transition.

\paragraph{Evaluation input.}
We use InternVL3.5-38B~\cite{wang2025internvl35advancingopensourcemultimodal} with deterministic decoding. For each case, we evaluate all six segments and all five switches. Metrics with the same temporal window and frame count receive the same sampled frames across methods. Each query supplies temporally ordered frames, the metric-specific checklist, and the relevant prompt text where required. Table~\ref{tab:benchmark_sampling} gives the temporal windows. A switch occurs at time $t$; uniform sampling of $n$ frames inside $(a,b)$ uses times $a+(b-a)j/(n+1)$ for $j=1,\ldots,n$.

\begin{table*}[t]
\centering
\small
\caption{Metric-specific temporal sampling. All times are in seconds. Reference frames establish the preceding visual state; subsequent frames show the transition to be judged.}
\label{tab:benchmark_sampling}
\begin{tabular}{@{}l>{\raggedright\arraybackslash}p{0.58\linewidth}l@{}}
\toprule
Metric & Frames supplied & prompt text \\
\midrule
SPF, SPP & 3 uniformly spaced frames within the 10-second segment. & Current prompt \\
SMC & 6 consecutive frames near the segment midpoint. & None \\
VBS, MBS & 2 frames in $(t-2,t)$ and 2 in $(t,t+2)$. & None \\
OSE & 1 reference at $t-0.5$ and 4 in $(t,t+1.5)$. & Both prompts \\
NSE & 1 reference at $t-0.5$ and 4 in $(t,t+5)$. & Both prompts \\
SP, ScP & 2 frames in $(t-1,t)$ and 2 in $(t,t+1)$. & Both prompts \\
SUF & 2 frames in $(t-1,t)$ and 4 in $(t,t+5)$. & Both prompts \\
TP & 2 frames in $(t-2,t)$ and 6 in $(t,t+5)$. & Both prompts \\
\bottomrule
\end{tabular}

\end{table*}

\paragraph{Scoring and aggregation.}
Each metric uses ten binary criteria, reproduced below in their evaluator order. Let $y_{imqj}\in\{0,1\}$ denote the judgment for criterion $j$ of metric $m$ on sampled segment or boundary $q$ in case $i$. The query score, case score, and benchmark score are
\[
\begin{aligned}
s_{imq}&=\frac{1}{10}\sum_{j=1}^{10}y_{imqj},
\\
s_{im}&=\frac{1}{|\mathcal{Q}_{im}|}\sum_{q\in\mathcal{Q}_{im}}s_{imq},
\\
S_m&=\frac{1}{N_m}\sum_{i\in\mathcal{V}_m}s_{im},
\end{aligned}
\]
where $\mathcal{Q}_{im}$ contains the sampled evaluation units and $\mathcal{V}_m$ contains the cases with available metric scores, with $N_m=|\mathcal{V}_m|$. For SP, $\mathcal{V}_{\mathrm{SP}}$ contains only
subject-bearing cases, giving $N_{\mathrm{SP}}=300$ for complete evaluations. Each case has equal weight.
All scores lie in $[0,1]$, with higher values indicating better performance. The segment and switch aggregates are the unweighted means of their three and eight metrics, respectively. For complete evaluations, the Overall score is
\[
S_{\mathrm{Overall}}=\frac{1}{11}\sum_{m\in\mathcal{M}}S_m,
\]
where $\mathcal{M}$ contains all eleven metrics. The evaluator returns an ordered binary vector and its sum in JSON. The same metric checklist is applied across applicable cases and methods. For environmental sequences, the evaluator is instructed to judge the changing visual content without assuming a human subject. The following tables provide the exact ten criteria for each metric, together with its definition and evaluation context.

%% file: tables/benchmark_comparison.tex
\begin{table*}[t]
\centering
\scriptsize
\setlength{\tabcolsep}{3.2pt}
\renewcommand{\arraystretch}{1.12}
\caption{
Comparison with representative video-generation benchmarks.
$\checkmark$, $\triangle$, and $\times$ denote explicit, partial/indirect,
and absent evaluation, respectively.
}
\resizebox{\textwidth}{!}{
\begin{tabular}{lcc|ccc|ccccc}
\toprule
&
\multicolumn{2}{c|}{\textbf{Generation Setting}}
&
\multicolumn{3}{c|}{\textbf{Existing Evaluation Focus}}
&
\multicolumn{5}{c}{\textbf{Semantic Transition Evaluation}}
\\
\cmidrule(lr){2-3}
\cmidrule(lr){4-6}
\cmidrule(lr){7-11}
\textbf{Benchmark}
&
\textbf{Multi-}
&
\textbf{Runtime}
&
\textbf{Update}
&
\textbf{State}
&
\textbf{Boundary}
&
\textbf{Prior-Event}
&
\textbf{Transition}
&
\textbf{State}
&
\textbf{Event}
&
\textbf{Overall}
\\[-2pt]
&
\textbf{Prompt}
&
\textbf{Switch}
&
\textbf{Fulfillment}
&
\textbf{Retention}
&
\textbf{Smoothness}
&
\textbf{Resolution}
&
\textbf{Plausibility}
&
\textbf{Readiness}
&
\textbf{Initiation}
&
\textbf{Coherence}
\\
\midrule
\multicolumn{11}{l}{\emph{General video generation benchmarks}} \\
VBench~\citep{huang2024vbench}
& $\times$ & $\times$ & $\times$ & $\times$ & $\times$
& $\times$ & $\times$ & $\times$ & $\times$ & $\times$ \\
EvalCrafter~\citep{liu2024evalcrafter}
& $\times$ & $\times$ & $\times$ & $\times$ & $\times$
& $\times$ & $\times$ & $\times$ & $\times$ & $\times$ \\
T2VBench~\citep{ji2024t2vbench}
& $\times$ & $\times$ & $\checkmark$ & $\times$ & $\times$
& $\triangle$ & $\triangle$ & $\times$ & $\triangle$ & $\times$ \\
T2V-CompBench~\citep{sun2025t2v}
& $\times$ & $\times$ & $\checkmark$ & $\times$ & $\times$
& $\times$ & $\triangle$ & $\times$ & $\times$ & $\times$ \\
VBench-2.0~\citep{zheng2025vbench}
& $\times$ & $\times$ & $\checkmark$ & $\checkmark$ & $\times$
& $\times$ & $\triangle$ & $\times$ & $\times$ & $\triangle$ \\
\midrule
\multicolumn{11}{l}{\emph{Long-form and streaming video benchmarks}} \\
NarraStream-Bench~\citep{liu2026iamflow}
& $\checkmark$ & $\triangle$ & $\checkmark$ & $\checkmark$ & $\checkmark$
& $\times$ & $\triangle$ & $\times$ & $\times$ & $\triangle$ \\
StateBench~\citep{miao2026video}
& $\checkmark$ & $\triangle$ & $\checkmark$ & $\checkmark$ & $\times$
& $\times$ & $\triangle$ & $\triangle$ & $\times$ & $\triangle$ \\
StreamAV-Bench~\citep{liu2026streamav}
& $\checkmark$ & $\checkmark$ & $\checkmark$ & $\checkmark$ & $\checkmark$
& $\triangle$ & $\triangle$ & $\times$ & $\triangle$ & $\triangle$ \\
\midrule
\textbf{\bench\ (Ours)}
& $\checkmark$ & $\checkmark$ & $\checkmark$ & $\checkmark$ & $\checkmark$
& $\checkmark$ & $\checkmark$ & $\checkmark$ & $\checkmark$ & $\checkmark$ \\
\bottomrule
\end{tabular}
}
\vspace{2pt}
\label{tab:benchmark_comparison}

\end{table*}

%% file: appendix_sections/training-data-curation.tex
\section{Video-Grounded Construction of Transition Training Data}
\label{app:training_data_construction}

We use Qwen3-VL-235B-A22B-Instruct~\cite{qwen3technicalreport} to extract approximately 45k intermediate \emph{segue prompts} from 5-second video clips associated with VidProM~\cite{wang2024vidprom}. Given a clip, the model describes the visible state at the start ($t=0$) and end ($t=5$\,s, represented by the last available frame). These descriptions serve as the source prompt $c^{-}$ and target prompt $c^{+}$, respectively. The model then examines the intervening frames to identify how the scene moves from $c^{-}$ to $c^{+}$. For example, if an object ends up in a person's hand, the clip provides evidence of whether the person picked it up or received it from someone else.

The model expresses the observed transition as a sequence of segue prompts ordered by when the changes occur in the clip. Each prompt describes an observable change, such as ending an action, releasing contact, or moving into position, and is assigned its corresponding TD-Schema role. We use the same schedule format as in the main paper: up to four segue prompts with positive integer durations totaling at most four generation blocks. The 5-second clip provides the visual evidence for extracting the transition, while the block durations specify how long each prompt conditions the generator.

During training, the student generates its own rollout under the extracted schedule, and \cdmd{} assigns each prompt's supervision to its corresponding temporal span. At inference, the online planner uses the latest generated frame and the user's prompt update to produce a schedule in the same format.

\paragraph{Prompt template.}
Table~\ref{tab:training_data_prompt} shows the prompt template used with
Qwen3-VL-235B-A22B-Instruct to extract the source state $c^{-}$, the
target state $c^{+}$, and the intervening segue prompts from each
5-second video clip.

\begin{table}[t]
  \centering
  \small
  \caption{Prompt template for extracting endpoint states and intermediate
segue prompts from 5-second videos with Qwen3-VL-235B-A22B-Instruct.}
\label{tab:training_data_prompt}
  \begin{tabular}{@{}>{\raggedright\arraybackslash}p{\linewidth}@{}}
    \toprule
    \textbf{System message} \\
    You annotate state transitions in videos. Ground each description
    in visible evidence and use consistent names for the same subjects
    and objects throughout the clip. \\
    \midrule
    \textbf{User message} \\
    Input: a 5-second video clip. Watch the complete clip and extract
    the following in order. \\[2pt]

    \textbf{1. Endpoint states.}
    Describe the state visible in the first frame as
    \texttt{source\_prompt} and the state visible in the last frame as
    \texttt{target\_prompt}. Record the action, pose, object contacts,
    and spatial relations relevant to the change. \\[2pt]

    \textbf{2. Observed transition.}
    Trace what happens between the two states. Select up to four
    intermediate changes and describe each in a short, self-contained
    segue prompt. Preserve their observed order. Include a change only
    when the clip provides evidence for it. Leave the final state to
    \texttt{target\_prompt}. \\[2pt]

    \textbf{3. Roles and durations.}
    Assign each change a role:
    \textsc{Terminate} for ending an ongoing action,
    \textsc{Release} for removing a contact or dependency,
    \textsc{Align} for establishing a prerequisite, or
    \textsc{Entry} for a visible entity arrival.
    Assign each prompt a positive integer duration in generation
    blocks, with at most four blocks in total. \\[2pt]

    \textbf{Output.}
    Return a JSON object with \texttt{source\_prompt},
    \texttt{target\_prompt}, and an ordered list
    \texttt{segue\_prompts}. Each list entry contains
    \texttt{role}, \texttt{prompt}, and \texttt{duration\_blocks}. \\
    \bottomrule
  \end{tabular}
\end{table}

%% file: appendix_sections/appendix_human_alignment.tex
\section{Human Alignment of the Evaluation Metrics}
\label{app:human-alignment}

\begin{figure}[t]
\centering
\includegraphics[width=\linewidth]{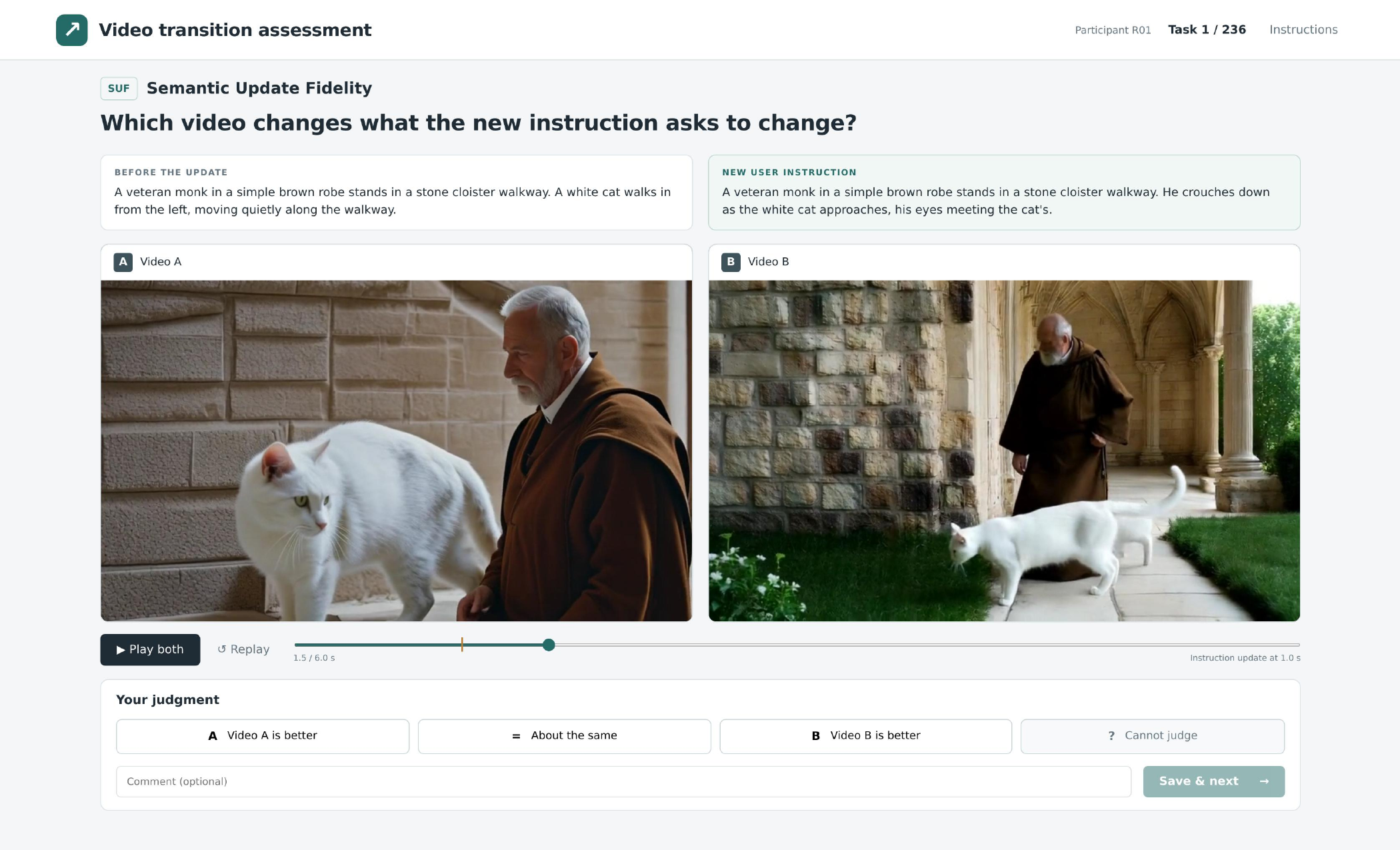}
\caption{Implemented interface for the human-alignment study.
Annotators compare synchronized clips alongside the preceding and updated
instructions and answer a metric-specific question, selecting A, B, a tie, or \emph{cannot judge}.}
\label{fig:human-alignment-interface}
\end{figure}

Following StreamAV-Bench~\citep{liu2026streamav}, we conduct
a blinded pairwise study to assess the agreement between
our eleven evaluation metrics and human judgments.
The cases and annotators used in this study are disjoint
from those in the user study (Sec.~\ref{app:user-study}).
We sample 30 cases, stratified by benchmark category, from cases with valid outputs from all 13 methods. For each metric, we assign each of the 78 unordered method pairs to one case and temporal unit, yielding 78 comparisons per metric and 858 comparisons in total. For SP, comparisons are assigned only to subject-bearing cases. Case and temporal-position assignments are balanced independently of benchmark scores. From a pool of 12 annotators, three distinct annotators judge each comparison, yielding 2,574 original judgments. We include 258 additional hidden repeat judgments (approximately 10\% of the original judgments) for consistency checks, bringing the planned total to 2,832 judgments.

Figure~\ref{fig:human-alignment-interface} shows the annotation interface. Each task presents one metric-specific question, the relevant instructions, and synchronized clips labeled A and B, with method identities and benchmark scores hidden. Task order and left--right placement are randomized. Annotators judge clips at normal speed and select A, B, a tie, or \emph{cannot judge}. Segment metrics use the corresponding instruction segment; transition metrics use windows around the instruction update. Human judgments use continuous playback, whereas our benchmark evaluates sampled frames.

A majority among three valid original judgments determines each task's label; incomplete or unresolved tasks are excluded. For each metric, we aggregate human preferences into method-level win rates, assigning 1 to a win, 0 to a loss, and 0.5 to a tie. We report Spearman correlations between these human win rates and mean benchmark scores computed on the same resolved, score-matched cases and temporal units. Hidden repeats serve only as consistency checks and do not enter the reported correlations.
Of the 858 planned comparisons, 831 yielded a resolved majority judgment; the remaining 27 were excluded because of incomplete or unresolved judgments.
Of the 258 planned hidden repeat judgments, 254 formed valid
pairs with their original judgments, with a repeat-consistency
rate of 93\%.
Pairwise agreement among original annotators was 88\%,
computed using valid original judgments, including those
from comparisons without a resolved majority judgment.
Both agreement measures include ties and compare method
identities rather than screen positions.

\begin{table*}[t]
\centering
\small
\setlength{\tabcolsep}{3pt}
\caption{Human alignment of the eleven evaluation metrics: Spearman correlation between method-level human win rates and the corresponding benchmark scores.}
\label{tab:human-alignment-supp}
\begin{tabular}{@{}l*{11}{c}@{}}
\toprule
Metric & SPF & SMC & SPP & VBS & MBS & OSE & NSE & SP & ScP & SUF & TP \\
\midrule
Spearman $\rho$ & 0.82 & 0.88 & 0.78 & 0.83 & 0.85 & 0.93 & 0.90 & 0.88 & 0.91 & 0.90 & 0.89 \\
\bottomrule
\end{tabular}
\end{table*}

%% file: appendix_sections/appendix_additional_training_hyperpara.tex
\section{Additional Implementation Details}
\label{app:training_hyperparameters}

Table~\ref{tab:training_hyperparameters} summarizes the sampling,
planning-budget, and training settings.
The generator is initialized from the LongLive base checkpoint
without long-video tuning.
LoRA adapters are applied to all linear layers within the
student and fake-score transformer blocks; their backbone weights,
the text encoder, and the video autoencoder remain frozen.
Each generation block contains three latent frames,
corresponding to 12 video frames
(approximately 0.75\,s at 16\,fps).
The four-block planning budget therefore allows approximately
3\,s of segue prompts before control returns to the updated prompt.

\begin{table*}[t]
\centering
\small
\setlength{\tabcolsep}{4pt}
\renewcommand{\arraystretch}{1.08}
\caption{Implementation settings. Optimizer and LoRA settings
apply to both trainable networks unless otherwise specified.}
\label{tab:training_hyperparameters}
\begin{tabular}{@{}>{\raggedright\arraybackslash}p{0.46\linewidth}
                  >{\raggedright\arraybackslash}p{0.50\linewidth}@{}}
\toprule
Hyperparameter & Setting \\
\midrule
\multicolumn{2}{@{}l}{\textit{Sampling and planning}} \\
Student sampler & Four-step causal sampling \\
Sampling timesteps & $\{1000,750,500,250\}$ \\
Timestep shift & $5$ \\
Generation block size & $3$ latent frames \\
Video frame rate & $16$\,fps \\
Local attention window / sink frames & $12$ / $3$ latent frames \\
Planner fine-tuning & None \\
Maximum transition budget $B$ & $4$ generation blocks \\
Maximum number of segue prompts & $4$ \\
\midrule
\multicolumn{2}{@{}l}{\textit{Training}} \\
Optimizer & AdamW \\
Training iterations & $3{,}000$ \\
Learning rates: student / fake score & $1\times10^{-5}$ / $2\times10^{-6}$ \\
Adam coefficients $(\beta_1,\beta_2)$ & $(0,0.999)$ \\
Weight decay / gradient-norm clipping & $0.01$ / $10$ \\
Learning-rate schedule & Constant; no warm-up or decay \\
Update frequency & Fake score: every iteration; student: every fifth iteration \\
Number of GPUs & $32$ \\
Precision & bfloat16 \\
Effective batch size & $32$ \\
Batch size per GPU / accumulation steps & $1$ / $1$ \\
LoRA rank / scaling parameter $\alpha$ & $256$ / $256$ \\
LoRA dropout & $0$ \\
Exponential moving average & Disabled \\
Memory optimization & Hybrid-sharded FSDP and gradient checkpointing \\
Configured rollout-length cap & $240$ latent frames \\
Training switch position (latent frames) &
Uniform over $\{21+18j\mid j=0,\ldots,10\}$ \\
Random seed & $0$ \\
\bottomrule
\end{tabular}
\end{table*}

The end-to-end planning latency reported in
the TD-Schema context-length plot in Sec.~5.3 of the main paper is measured from the start of
input-image decoding to the completion of segue-prompt
generation, including image decoding and the entire
planning process. The planner runs asynchronously in
parallel with the video generator. Generation continues
under $c^{-}$ until the first generation-block boundary
after the schedule returns, when the schedule takes effect.

For score evaluation, we sample an integer timestep $u$ uniformly
from $\{0,\ldots,999\}$ and use
$t=\operatorname{clip}\!\left(
1000\frac{5(u/1000)}{1+4(u/1000)},20,980
\right)$,
shared across the frames and span conditions of a rollout.
The teacher guidance convention is
$\hat{x}_{\mathrm{real}}=\hat{x}_{\mathrm{cond}}
+3(\hat{x}_{\mathrm{cond}}-\hat{x}_{\mathrm{uncond}})$;
the fake-score network uses its conditional prediction.

%% file: tables/metric_checklists.tex
\section{Metric Checklists}\label{app:checklists}
\Needspace{10\baselineskip}
\paragraph{SPF: Segment Prompt Fidelity.}
Whether the segment realizes its assigned instruction, including entities, attributes, actions, and scene context. Three uniformly spaced frames within the segment; the segment prompt is provided.

\begingroup
\small
\captionof{table}{Segment Prompt Fidelity (SPF): ten binary evaluation criteria.}\label{tab:benchmark_checklist_SPF}
\par
\tablefirsthead{\toprule
No. & Criterion (YES = 1, NO = 0) \\
\midrule}
\tablehead{\multicolumn{2}{@{}l@{}}{\tablename~\thetable{} (continued)}\\
\toprule
No. & Criterion (YES = 1, NO = 0) \\
\midrule}
\tabletail{\midrule}
\tablelasttail{\bottomrule}
\noindent\begin{supertabular}{@{}r@{\hspace{0.6em}}>{\raggedright\arraybackslash}p{\dimexpr\linewidth-3.1em\relax}@{}}
1 & The main content described in the prompt is visible in the frames. \\[1.5pt]
2 & The appearance attributes stated in the prompt (clothing, colours, materials, textures, forms) match what is shown. \\[1.5pt]
3 & The core action or change described in the prompt is taking place in the frames. \\[1.5pt]
4 & Every entity or object named in the prompt is present in the frames, with none missing. \\[1.5pt]
5 & The entities and objects in the frames behave, or are used, in the way the prompt describes. \\[1.5pt]
6 & The scene or environment shown matches the prompt description. \\[1.5pt]
7 & The atmosphere and lighting shown match the prompt description. \\[1.5pt]
8 & The frames contain no prominent content that the prompt does not mention. \\[1.5pt]
9 & The progression across the 3 frames is consistent with the temporal order implied by the prompt. \\[1.5pt]
10 & The overall visual content is highly consistent with the core semantics of the prompt, with no significant omission or deviation. \\[1.5pt]
\end{supertabular}\par
\endgroup

\Needspace{10\baselineskip}
\paragraph{SMC: Segment Motion Continuity.}
Whether motion within a segment remains temporally continuous, with stable trajectories, shapes, and background. Six consecutive native-rate frames starting 0.12 seconds before the segment midpoint; no text instruction is supplied.

\begingroup
\small
\captionof{table}{Segment Motion Continuity (SMC): ten binary evaluation criteria.}\label{tab:benchmark_checklist_SMC}
\par
\tablefirsthead{\toprule
No. & Criterion (YES = 1, NO = 0) \\
\midrule}
\tablehead{\multicolumn{2}{@{}l@{}}{\tablename~\thetable{} (continued)}\\
\toprule
No. & Criterion (YES = 1, NO = 0) \\
\midrule}
\tabletail{\midrule}
\tablelasttail{\bottomrule}
\noindent\begin{supertabular}{@{}r@{\hspace{0.6em}}>{\raggedright\arraybackslash}p{\dimexpr\linewidth-3.1em\relax}@{}}
1 & The position of the moving content changes gradually across frames with no sudden teleportation. \\[1.5pt]
2 & The shape or posture of the moving content transitions smoothly across frames with no abrupt jumps. \\[1.5pt]
3 & No part of the moving content shows clipping, fracturing or abnormal distortion. \\[1.5pt]
4 & There is no flickering between frames (no sudden jumps in brightness or colour tone). \\[1.5pt]
5 & There are no frame skips or scene resets between the 6 frames. \\[1.5pt]
6 & Every moving element in the scene has a continuous trajectory with no sudden appearance or disappearance. \\[1.5pt]
7 & The speed of motion is consistent across frames with no unreasonable sudden stop or acceleration. \\[1.5pt]
8 & There are no visual artifacts such as tearing, ghosting or block-level corruption in any frame. \\[1.5pt]
9 & The background remains stable across the 6 frames with no unexplained jitter or jump. \\[1.5pt]
10 & The overall sequence conveys a natural sense of temporal flow with no local time reversal or freeze. \\[1.5pt]
\end{supertabular}\par
\endgroup

\Needspace{10\baselineskip}
\paragraph{SPP: Segment Physical Plausibility.}
Whether depicted structure, contact, support, scale, and material dynamics are physically coherent. Three uniformly spaced frames within the segment; the segment prompt is provided.

\begingroup
\small
\captionof{table}{Segment Physical Plausibility (SPP): ten binary evaluation criteria.}\label{tab:benchmark_checklist_SPP}
\par
\tablefirsthead{\toprule
No. & Criterion (YES = 1, NO = 0) \\
\midrule}
\tablehead{\multicolumn{2}{@{}l@{}}{\tablename~\thetable{} (continued)}\\
\toprule
No. & Criterion (YES = 1, NO = 0) \\
\midrule}
\tabletail{\midrule}
\tablelasttail{\bottomrule}
\noindent\begin{supertabular}{@{}r@{\hspace{0.6em}}>{\raggedright\arraybackslash}p{\dimexpr\linewidth-3.1em\relax}@{}}
1 & The structure of every depicted body or form is coherent, with no parts bent or deformed beyond a natural range. \\[1.5pt]
2 & Weight and support relationships in the frames are physically reasonable, with nothing floating or unsupported without explanation. \\[1.5pt]
3 & Contact with the ground or supporting surfaces is physically plausible, with no clipping through surfaces and no unexplained hovering. \\[1.5pt]
4 & Every object in the frames is positioned and oriented in a physically plausible way. \\[1.5pt]
5 & The relative size and scale of all elements in the frames are realistic and mutually consistent. \\[1.5pt]
6 & Contact points between elements show no clipping or interpenetration. \\[1.5pt]
7 & Applied forces and the resulting deformation or displacement are consistent with each other. \\[1.5pt]
8 & Materials behave according to their nature (cloth drapes and folds, water flows, smoke disperses, rigid objects stay rigid). \\[1.5pt]
9 & Secondary dynamics (clothing, hair, foliage, ripples, dust) are consistent with the intensity of the motion shown. \\[1.5pt]
10 & There are no duplicated, missing or malformed parts in any element in the frames. \\[1.5pt]
\end{supertabular}\par
\endgroup

\Needspace{10\baselineskip}
\paragraph{VBS: Visual Boundary Smoothness.}
Whether imaging properties remain continuous across an instruction boundary, assessed relative to variation within each side. Frames 1--2: two samples in $(t-2,t)$; frames 3--4: two in $(t,t+2)$. No text instruction is supplied.

\begingroup
\small
\captionof{table}{Visual Boundary Smoothness (VBS): ten binary evaluation criteria.}\label{tab:benchmark_checklist_VBS}
\par
\tablefirsthead{\toprule
No. & Criterion (YES = 1, NO = 0) \\
\midrule}
\tablehead{\multicolumn{2}{@{}l@{}}{\tablename~\thetable{} (continued)}\\
\toprule
No. & Criterion (YES = 1, NO = 0) \\
\midrule}
\tabletail{\midrule}
\tablelasttail{\bottomrule}
\noindent\begin{supertabular}{@{}r@{\hspace{0.6em}}>{\raggedright\arraybackslash}p{\dimexpr\linewidth-3.1em\relax}@{}}
1 & There is no flash, white frame or black frame between frames 2 and 3. \\[1.5pt]
2 & There is no visible tearing, blocking or encoding artifact between frames 2 and 3. \\[1.5pt]
3 & There is no abrupt drop or rise in resolution or sharpness between frames 2 and 3. \\[1.5pt]
4 & The camera focal length or viewing angle between frames 2 and 3 stays on the trend established by frames 1 and 4, with no sudden change. \\[1.5pt]
5 & Overall, the change in brightness, colour tone and contrast from frame 2 to frame 3 does not feel more severe or more sudden than the change from frame 1 to frame 2 or from frame 3 to frame 4. \\[1.5pt]
6 & Overall, the four frames look like continuous footage from one take rather than two clips edited together. \\[1.5pt]
7 & The image noise or grain in frames 2 and 3 is at the same level as in frames 1 and 4, with no sudden cleaning up or coarsening. \\[1.5pt]
8 & There is no global colour-space or white-balance jump between frames 2 and 3 (for example a sudden yellow or blue cast). \\[1.5pt]
9 & Image quality is stable across all four frames, with no single frame suddenly blurred, over-exposed or under-exposed. \\[1.5pt]
10 & Taken together, there is no visual trace at frame 2 - 3 that could be attributed to two pieces of footage being cut together. \\[1.5pt]
\end{supertabular}\par
\endgroup

\Needspace{10\baselineskip}
\paragraph{MBS: Motion Boundary Smoothness.}
Whether trajectories, poses, and contact relationships continue coherently across the boundary. Frames 1--2: two samples in $(t-2,t)$; frames 3--4: two in $(t,t+2)$. No text instruction is supplied.

\begingroup
\small
\captionof{table}{Motion Boundary Smoothness (MBS): ten binary evaluation criteria.}\label{tab:benchmark_checklist_MBS}
\par
\tablefirsthead{\toprule
No. & Criterion (YES = 1, NO = 0) \\
\midrule}
\tablehead{\multicolumn{2}{@{}l@{}}{\tablename~\thetable{} (continued)}\\
\toprule
No. & Criterion (YES = 1, NO = 0) \\
\midrule}
\tabletail{\midrule}
\tablelasttail{\bottomrule}
\noindent\begin{supertabular}{@{}r@{\hspace{0.6em}}>{\raggedright\arraybackslash}p{\dimexpr\linewidth-3.1em\relax}@{}}
1 & The relative motion of scene elements across frames 2 and 3
is consistent with the apparent camera movement, allowing for
independent object motion and depth-dependent parallax. \\[1.5pt]
2 & Shape and posture change gradually from frame 2 to frame 3, rather than jumping straight to a completely different configuration. \\[1.5pt]
3 & No part of the content shows clipping, fracturing or implausible distortion in frames 2 and 3. \\[1.5pt]
4 & The motion keeps a consistent speed and direction from frame 2 to frame 3, with no sudden reversal or stall. \\[1.5pt]
5 & The facing direction of the content does not turn implausibly between frames 2 and 3. \\[1.5pt]
6 & The spatial structure of the background stays consistent between frames 2 and 3, with no jump to a differently laid-out space (judge layout only, not colour or style). \\[1.5pt]
7 & Contact relationships between the content and objects in the scene stay coherent from frame 2 to frame 3, with nothing detaching or attaching instantaneously. \\[1.5pt]
8 & There is no semi-transparent residue, ghosting or double-exposure blending of the old content into frames 3 and 4. \\[1.5pt]
9 & Across the four frames the motion reads as one single coherent trajectory rather than two independent motions joined together. \\[1.5pt]
10 & Taken together, the amount of motion or pose change at frame 2 - 3 stays within the natural rhythm of this content, with no teleport-like or break-like motion artifact. \\[1.5pt]
\end{supertabular}\par
\endgroup

\Needspace{10\baselineskip}
\paragraph{OSE: Old-State Exit.}
Whether the preceding process ends in a complete and understandable way, without abrupt erasure or unresolved activity. Frame 1: a reference at $t-0.5$; frames 2--5: four samples in $(t,t+1.5)$. Both adjacent prompts are provided.

\begingroup
\small
\captionof{table}{Old-State Exit (OSE): ten binary evaluation criteria.}\label{tab:benchmark_checklist_OSE}
\par
\tablefirsthead{\toprule
No. & Criterion (YES = 1, NO = 0) \\
\midrule}
\tablehead{\multicolumn{2}{@{}l@{}}{\tablename~\thetable{} (continued)}\\
\toprule
No. & Criterion (YES = 1, NO = 0) \\
\midrule}
\tabletail{\midrule}
\tablelasttail{\bottomrule}
\noindent\begin{supertabular}{@{}r@{\hspace{0.6em}}>{\raggedright\arraybackslash}p{\dimexpr\linewidth-3.1em\relax}@{}}
1 & The process under way in frame 1 has a recognisable ending in frames 2-5, rather than vanishing between two adjacent frames. \\[1.5pt]
2 & The old state is not frozen in an obviously unfinished intermediate form. \\[1.5pt]
3 & The form in which the old state ends is one in which that kind of process could plausibly stop in reality. \\[1.5pt]
4 & The ending of the old state leaves nothing suspended and unresolved (no change stopping halfway with no follow-through). \\[1.5pt]
5 & Every element taking part in the old state in frame 1 has an explicable fate in frames 2-5, with nothing vanishing or resetting out of nowhere. \\[1.5pt]
6 & The way the rate of change winds down is explicable, with no unattributable instant drop to zero. \\[1.5pt]
7 & By frames 4--5, any ongoing activity that conflicts with
the updated instruction has ended; compatible activity
may continue. \\[1.5pt]
8 & Traces left by the old state (dust, ripples, moving objects, residual light) settle or disperse plausibly rather than being wiped out at once. \\[1.5pt]
9 & No extra change unrelated to either description interrupts the exit of the old state. \\[1.5pt]
10 & Overall, the exit of the old state is a complete and understandable process rather than being cut off by force. \\[1.5pt]
\end{supertabular}\par
\endgroup

\Needspace{10\baselineskip}
\paragraph{NSE: New-State Entry.}
Whether the new event begins from the realized preceding state through visible preparation and reachable intermediate configurations. Frame 1: a reference at $t-0.5$; frames 2--5: four samples in $(t,t+5)$. Both adjacent prompts are provided.

\begingroup
\small
\captionof{table}{New-State Entry (NSE): ten binary evaluation criteria.}\label{tab:benchmark_checklist_NSE}
\par
\tablefirsthead{\toprule
No. & Criterion (YES = 1, NO = 0) \\
\midrule}
\tablehead{\multicolumn{2}{@{}l@{}}{\tablename~\thetable{} (continued)}\\
\toprule
No. & Criterion (YES = 1, NO = 0) \\
\midrule}
\tabletail{\midrule}
\tablelasttail{\bottomrule}
\noindent\begin{supertabular}{@{}r@{\hspace{0.6em}}>{\raggedright\arraybackslash}p{\dimexpr\linewidth-3.1em\relax}@{}}
1 & The new state starts from the state shown in frame 1, rather than starting from a completely different starting point out of nowhere. \\[1.5pt]
2 & The spatial configuration of the elements when the new state begins is reached from the configuration in frame 1 through a visible process of change. \\[1.5pt]
3 & The build-up or preparation that the new state requires is visible in frames 2-5 rather than skipped. \\[1.5pt]
4 & The first visible stage of the new state is genuinely its beginning, rather than opening from the middle or from an already completed form. \\[1.5pt]
5 & No necessary intermediate step is missing between frames 2-5 (the viewer does not have to imagine an unshown transition). \\[1.5pt]
6 & Where an element present in frame 1 changes its position or form, that change is visible as a process rather than happening instantaneously. \\[1.5pt]
7 & The relationship of the new state to the existing surroundings (ground, background, nearby objects, water surface, light sources) carries over from frame 1 without changing out of nowhere. \\[1.5pt]
8 & The new state does not rely on any element that is absent in frame 1 and never shown entering the scene. \\[1.5pt]
9 & The spatial framing and layout established in frame 1 still hold while the new state starts, with no reset. \\[1.5pt]
10 & Overall, the start of the new state satisfies all the preconditions it needs and is genuinely reachable from the state in frame 1. \\[1.5pt]
\end{supertabular}\par
\endgroup

\Needspace{10\baselineskip}
\paragraph{SP: Subject Preservation.}
Whether the persistent subject retains its identity across the update, accounting for changes explained by pose, lighting, or the instructions. Frames 1--2: two samples in $(t-1,t)$; frames 3--4: two in $(t,t+1)$. Both adjacent prompts are provided. Newly introduced subjects are ignored. SP is evaluated only on the 300 subject-bearing cases;
the 60 subject-free cases are excluded.

\begingroup
\small
\captionof{table}{Subject Preservation (SP): ten binary evaluation criteria.}\label{tab:benchmark_checklist_SP}
\par
\tablefirsthead{\toprule
No. & Criterion (YES = 1, NO = 0) \\
\midrule}
\tablehead{\multicolumn{2}{@{}l@{}}{\tablename~\thetable{} (continued)}\\
\toprule
No. & Criterion (YES = 1, NO = 0) \\
\midrule}
\tabletail{\midrule}
\tablelasttail{\bottomrule}
\noindent\begin{supertabular}{@{}r@{\hspace{0.6em}}>{\raggedright\arraybackslash}p{\dimexpr\linewidth-3.1em\relax}@{}}
1 & Facial features (face shape, proportions of the features, skin tone) stay consistent across the four frames. \\[1.5pt]
2 & Hairstyle and hair colour stay consistent across the four frames. \\[1.5pt]
3 & Clothing style stays consistent across the four frames. \\[1.5pt]
4 & Clothing colour stays consistent across the four frames. \\[1.5pt]
5 & Body shape and proportions stay consistent across the four frames. \\[1.5pt]
6 & Identity attributes such as gender, apparent age and ethnicity stay consistent across the four frames. \\[1.5pt]
7 & Signature accessories (glasses, hat, belt, held items) stay consistent across the four frames, with none appearing or disappearing without reason. \\[1.5pt]
8 & The difference in appearance between frame 2 and frame 3 is no larger than the natural pose- and lighting-driven difference between frames 1 and 2 or between frames 3 and 4. \\[1.5pt]
9 & It is the same recognisable individual throughout all four frames, not replaced by someone else. \\[1.5pt]
10 & Any difference between frames 3-4 and frames 1-2 can be explained by a change of pose, a change of lighting, or the content of the two descriptions, rather than being unexplained drift. \\[1.5pt]
\end{supertabular}\par
\endgroup

\Needspace{10\baselineskip}
\paragraph{ScP: Scene Preservation.}
Whether environmental changes match the instruction update while the spatial context and unrelated scene elements remain coherent. Frames 1--2: two samples in $(t-1,t)$; frames 3--4: two in $(t,t+1)$. Both adjacent prompts are provided.

\begingroup
\small
\captionof{table}{Scene Preservation (ScP): ten binary evaluation criteria.}\label{tab:benchmark_checklist_ScP}
\par
\tablefirsthead{\toprule
No. & Criterion (YES = 1, NO = 0) \\
\midrule}
\tablehead{\multicolumn{2}{@{}l@{}}{\tablename~\thetable{} (continued)}\\
\toprule
No. & Criterion (YES = 1, NO = 0) \\
\midrule}
\tabletail{\midrule}
\tablelasttail{\bottomrule}
\noindent\begin{supertabular}{@{}r@{\hspace{0.6em}}>{\raggedright\arraybackslash}p{\dimexpr\linewidth-3.1em\relax}@{}}
1 & The amount by which the spatial structure of the background changes across the four frames matches the environmental difference implied by the two descriptions. \\[1.5pt]
2 & The amount by which the lighting environment (direction, intensity, colour temperature) changes across the four frames matches the environmental difference implied by the two descriptions. \\[1.5pt]
3 & No environmental element appears in frames 3-4 that neither description mentions. \\[1.5pt]
4 & No environmental element present in frames 1-2 disappears in frames 3-4 in a way neither description can explain. \\[1.5pt]
5 & Any environmental change across the four frames happens progressively, with no wholesale instantaneous replacement. \\[1.5pt]
6 & The spatial relationship of the content to background landmarks stays coherent across the four frames, with no teleporting into another space. \\[1.5pt]
7 & The positions and the number of pre-existing objects in the environment stay coherent across the four frames. \\[1.5pt]
8 & The framing and depth of field stay coherent across the four frames, with no jump to a completely different view. \\[1.5pt]
9 & The degree of environmental change is proportional to the semantic difference between the two descriptions, neither clearly excessive nor clearly insufficient. \\[1.5pt]
10 & Overall, the environment across the four frames is consistent with the combined narrative logic of the two descriptions. \\[1.5pt]
\end{supertabular}\par
\endgroup

\Needspace{10\baselineskip}
\paragraph{SUF: Semantic Update Fidelity.}
Whether the visible change matches the semantic difference between the instructions, preserving what they share. Frames 1--2: two samples in $(t-1,t)$; frames 3--6: four in $(t,t+5)$. Both adjacent prompts are provided.

\begingroup
\small
\captionof{table}{Semantic Update Fidelity (SUF): ten binary evaluation criteria.}\label{tab:benchmark_checklist_SUF}
\par
\tablefirsthead{\toprule
No. & Criterion (YES = 1, NO = 0) \\
\midrule}
\tablehead{\multicolumn{2}{@{}l@{}}{\tablename~\thetable{} (continued)}\\
\toprule
No. & Criterion (YES = 1, NO = 0) \\
\midrule}
\tabletail{\midrule}
\tablelasttail{\bottomrule}
\noindent\begin{supertabular}{@{}r@{\hspace{0.6em}}>{\raggedright\arraybackslash}p{\dimexpr\linewidth-3.1em\relax}@{}}
1 & Frames 3-6 show an observable change compared with frames 1-2. \\[1.5pt]
2 & What changed is exactly what the semantic difference between the two descriptions refers to. \\[1.5pt]
3 & The parts the two descriptions share (content, scene, style) stay unchanged across the six frames and were not altered along with the rest. \\[1.5pt]
4 & The magnitude of the visual change is commensurate with the semantic difference, without being clearly excessive. \\[1.5pt]
5 & The magnitude of the visual change is commensurate with the semantic difference, without being clearly insufficient. \\[1.5pt]
6 & What the new description refers to becomes established in frames 3-6, rather than appearing only as a brief trace that disappears again. \\[1.5pt]
7 & Content specific to the old description has receded in frames 3-6, without coexisting with the new content long enough to be confusing. \\[1.5pt]
8 & No additional, unattributable change appears in the six frames that neither description covers. \\[1.5pt]
9 & The change unfolds progressively between frames 2 and 6, rather than being completed in a single frame and then remaining static. \\[1.5pt]
10 & Overall, the six frames faithfully express the move from the previous description to the new one, missing nothing that should change and altering nothing that should not. \\[1.5pt]
\end{supertabular}\par
\endgroup

\Needspace{10\baselineskip}
\paragraph{TP: Transition Pacing.}
Whether response onset, transition duration, and progress toward the new state have a natural temporal rhythm. Frames 1--2: two samples in $(t-2,t)$; frames 3--8: six in $(t,t+5)$. Both adjacent prompts are provided.

\begingroup
\small
\captionof{table}{Transition Pacing (TP): ten binary evaluation criteria.}\label{tab:benchmark_checklist_TP}
\par
\tablefirsthead{\toprule
No. & Criterion (YES = 1, NO = 0) \\
\midrule}
\tablehead{\multicolumn{2}{@{}l@{}}{\tablename~\thetable{} (continued)}\\
\toprule
No. & Criterion (YES = 1, NO = 0) \\
\midrule}
\tabletail{\midrule}
\tablelasttail{\bottomrule}
\noindent\begin{supertabular}{@{}r@{\hspace{0.6em}}>{\raggedright\arraybackslash}p{\dimexpr\linewidth-3.1em\relax}@{}}
1 & The new content begins to appear soon after the switch, with no long lag spent still showing the old content. \\[1.5pt]
2 & The change is not completed instantaneously between two adjacent frames; a transition process is visible. \\[1.5pt]
3 & The duration of the transition suits the magnitude of the change, without feeling rushed. \\[1.5pt]
4 & The duration of the transition suits the magnitude of the change, without feeling drawn out. \\[1.5pt]
5 & The pace of motion around the switch stays natural, with no unreasonable sudden acceleration. \\[1.5pt]
6 & The pace of motion around the switch stays natural, with no unreasonable sudden deceleration or stall. \\[1.5pt]
7 & There is no idle gap during the transition in which nothing meaningful happens. \\[1.5pt]
8 & The change advances monotonically towards the new content, with no flip-flopping back and forth between old and new. \\[1.5pt]
9 & The new content is stably established before the end of the eight-frame window, rather than remaining in an unfinished transitional state. \\[1.5pt]
10 & Overall, the timing of this switch reads as an event unfolding naturally rather than being forced in or artificially stretched out. \\[1.5pt]
\end{supertabular}\par
\endgroup

%% file: main.bbl
\begin{thebibliography}{52}
\providecommand{\natexlab}[1]{#1}
\providecommand{\url}[1]{\texttt{#1}}
\expandafter\ifx\csname urlstyle\endcsname\relax
  \providecommand{\doi}[1]{doi: #1}\else
  \providecommand{\doi}{doi: \begingroup \urlstyle{rm}\Url}\fi

\bibitem[Ahn et~al.(2022)Ahn, Brohan, Brown, Chebotar, Cortes, David, Finn, Fu, Gopalakrishnan, Hausman, et~al.]{ahn2022saycan}
Michael Ahn, Anthony Brohan, Noah Brown, Yevgen Chebotar, Omar Cortes, Byron David, Chelsea Finn, Chuyuan Fu, Keerthana Gopalakrishnan, Karol Hausman, et~al.
\newblock Do as i can, not as i say: Grounding language in robotic affordances.
\newblock \emph{arXiv preprint arXiv:2204.01691}, 2022.

\bibitem[Cai et~al.(2025)Cai, Cun, Li, Liu, Zhang, Zhang, Shan, and Yue]{cai2025ditctrl}
Minghong Cai, Xiaodong Cun, Xiaoyu Li, Wenze Liu, Zhaoyang Zhang, Yong Zhang, Ying Shan, and Xiangyu Yue.
\newblock Ditctrl: Exploring attention control in multi-modal diffusion transformer for tuning-free multi-prompt longer video generation.
\newblock In \emph{2025 IEEE/CVF Conference on Computer Vision and Pattern Recognition (CVPR)}, 2025.

\bibitem[Chen et~al.(2026{\natexlab{a}})Chen, Huang, and Liu]{promptrelay2026}
Gordon Chen, Ziqi Huang, and Ziwei Liu.
\newblock Prompt relay: Inference-time temporal control for multi-event video generation.
\newblock \emph{arXiv preprint arXiv:2604.10030}, 2026{\natexlab{a}}.

\bibitem[Chen et~al.(2024)Chen, Wang, Zhang, Zhuang, Ma, Yu, Wang, Lin, Qiao, and Liu]{chen2024seine}
Xinyuan Chen, Yaohui Wang, Lingjun Zhang, Shaobin Zhuang, Xin Ma, Jiashuo Yu, Yali Wang, Dahua Lin, Yu Qiao, and Ziwei Liu.
\newblock {SEINE}: Short-to-long video diffusion model for generative transition and prediction.
\newblock In \emph{The Twelfth International Conference on Learning Representations}, 2024.

\bibitem[Chen et~al.(2026{\natexlab{b}})Chen, Wang, Huang, Yang, Zhang, Xiao, Chu, Mao, Hu, Liu, et~al.]{chen2026longlive2}
Yukang Chen, Luozhou Wang, Wei Huang, Shuai Yang, Bohan Zhang, Yicheng Xiao, Ruihang Chu, Weian Mao, Qixin Hu, Shaoteng Liu, et~al.
\newblock Longlive-2.0: An nvfp4 parallel infrastructure for long video generation.
\newblock \emph{arXiv preprint arXiv:2605.18739}, 2026{\natexlab{b}}.

\bibitem[Cui et~al.(2026)Cui, Wu, Li, Yang, Li, Wang, Bai, Ban, and Hsieh]{cui2026selfforcingpp}
Justin Cui, Jie Wu, Ming Li, Tao Yang, Xiaojie Li, Rui Wang, Andrew Bai, Yuanhao Ban, and Cho-Jui Hsieh.
\newblock Self-forcing++: Towards minute-scale high-quality video generation.
\newblock In \emph{The Fourteenth International Conference on Learning Representations}, 2026.

\bibitem[{HappyOyster}(2026)]{happyoyster2026}
{HappyOyster}.
\newblock {HappyOyster: Real-Time Interactive Open-World Model}.
\newblock \url{https://www.happyoyster.com/home}, 2026.

\bibitem[He et~al.(2025)He, Peng, Liu, Wang, Zhang, Cui, Kang, Jiang, An, Ren, et~al.]{he2025matrix}
Xianglong He, Chunli Peng, Zexiang Liu, Boyang Wang, Yifan Zhang, Qi Cui, Fei Kang, Biao Jiang, Mengyin An, Yangyang Ren, et~al.
\newblock Matrix-game 2.0: An open-source real-time and streaming interactive world model.
\newblock \emph{arXiv preprint arXiv:2508.13009}, 2025.

\bibitem[Hong et~al.(2023)Hong, Seo, Shin, Hong, and Kim]{hong2023direct2v}
Susung Hong, Junyoung Seo, Heeseong Shin, Sunghwan Hong, and Seungryong Kim.
\newblock Direct2v: Large language models are frame-level directors for zero-shot text-to-video generation.
\newblock \emph{arXiv preprint arXiv:2305.14330}, 2023.

\bibitem[Hu et~al.(2026)Hu, Yang, Huang, Han, and Chen]{hu2026longlive}
Qixin Hu, Shuai Yang, Wei Huang, Song Han, and Yukang Chen.
\newblock Longlive-rag: A general retrieval-augmented framework for long video generation.
\newblock \emph{arXiv preprint arXiv:2606.02553}, 2026.

\bibitem[Huang et~al.(2025)Huang, Li, He, Zhou, and Shechtman]{huang2025selfforcing}
Xun Huang, Zhengqi Li, Guande He, Mingyuan Zhou, and Eli Shechtman.
\newblock Self forcing: Bridging the train-test gap in autoregressive video diffusion.
\newblock \emph{arXiv preprint arXiv:2506.08009}, 2025.

\bibitem[Huang et~al.(2024{\natexlab{a}})Huang, He, Yu, Zhang, Si, Jiang, Zhang, Wu, Jin, Chanpaisit, Wang, Chen, Wang, Lin, Qiao, and Liu]{huang2024vbench}
Ziqi Huang, Yinan He, Jiashuo Yu, Fan Zhang, Chenyang Si, Yuming Jiang, Yuanhan Zhang, Tianxing Wu, Qingyang Jin, Nattapol Chanpaisit, Yaohui Wang, Xinyuan Chen, Limin Wang, Dahua Lin, Yu Qiao, and Ziwei Liu.
\newblock {VBench}: Comprehensive benchmark suite for video generative models.
\newblock In \emph{Proceedings of the IEEE/CVF Conference on Computer Vision and Pattern Recognition}, 2024{\natexlab{a}}.

\bibitem[Huang et~al.(2024{\natexlab{b}})Huang, Zhang, Xu, He, Yu, Dong, Ma, Chanpaisit, Si, Jiang, Wang, Chen, Chen, Wang, Lin, Qiao, and Liu]{huang2024vbenchcomprehensiveversatilebenchmark}
Ziqi Huang, Fan Zhang, Xiaojie Xu, Yinan He, Jiashuo Yu, Ziyue Dong, Qianli Ma, Nattapol Chanpaisit, Chenyang Si, Yuming Jiang, Yaohui Wang, Xinyuan Chen, Ying-Cong Chen, Limin Wang, Dahua Lin, Yu Qiao, and Ziwei Liu.
\newblock Vbench++: Comprehensive and versatile benchmark suite for video generative models, 2024{\natexlab{b}}.

\bibitem[Ji et~al.(2024)Ji, Xiao, Tai, and Huo]{ji2024t2vbench}
Pengliang Ji, Chuyang Xiao, Huilin Tai, and Mingxiao Huo.
\newblock T2vbench: Benchmarking temporal dynamics for text-to-video generation.
\newblock In \emph{2024 IEEE/CVF Conference on Computer Vision and Pattern Recognition Workshops (CVPRW)}, 2024.

\bibitem[Ji et~al.(2025)Ji, Chen, Yang, Tao, Wan, and Zhao]{ji2025memflow}
Sihui Ji, Xi Chen, Shuai Yang, Xin Tao, Pengfei Wan, and Hengshuang Zhao.
\newblock Memflow: Flowing adaptive memory for consistent and efficient long video narratives.
\newblock \emph{arXiv preprint arXiv:2512.14699}, 2025.

\bibitem[Liang et~al.(2025)Liang, Kodaira, Xu, Tomizuka, Keutzer, and Marculescu]{liang2025looking}
Feng Liang, Akio Kodaira, Chenfeng Xu, Masayoshi Tomizuka, Kurt Keutzer, and Diana Marculescu.
\newblock Looking backward: Streaming video-to-video translation with feature banks.
\newblock In \emph{The Thirteenth International Conference on Learning Representations}, 2025.

\bibitem[Lin et~al.(2023)Lin, Zala, Cho, and Bansal]{lin2023videodirectorgpt}
Han Lin, Abhay Zala, Jaemin Cho, and Mohit Bansal.
\newblock Videodirectorgpt: Consistent multi-scene video generation via llm-guided planning.
\newblock \emph{arXiv preprint arXiv:2309.15091}, 2023.

\bibitem[Liu et~al.(2026{\natexlab{a}})Liu, Zhang, Jiang, Wang, Liang, Xue, Yi, and Liu]{liu2026iamflow}
Jinzhuo Liu, Jiangning Zhang, Wencan Jiang, Yabiao Wang, Dingkang Liang, Zhucun Xue, Ran Yi, and Yong Liu.
\newblock Advancing narrative long video generation via training-free identity-aware memory.
\newblock \emph{arXiv preprint arXiv:2605.18733}, 2026{\natexlab{a}}.

\bibitem[Liu et~al.(2026{\natexlab{b}})Liu, Hu, Xu, Shan, and Lu]{liu2026rolling}
Kunhao Liu, Wenbo Hu, Jiale Xu, Ying Shan, and Shijian Lu.
\newblock Rolling forcing: Autoregressive long video diffusion in real time.
\newblock In \emph{The Fourteenth International Conference on Learning Representations}, 2026{\natexlab{b}}.

\bibitem[Liu et~al.(2026{\natexlab{c}})Liu, Zeng, Liu, Fang, Cai, Mao, Liu, Sheng, Weng, and Shi]{liu2026streamav}
Kaiqi Liu, Haoxuan Zeng, Jingqi Liu, Jiacong Fang, Ziqi Cai, Yunyao Mao, Henglin Liu, Yu Sheng, Shuchen Weng, and Boxin Shi.
\newblock Streamav-bench: A comprehensive benchmark for streaming audio-video generation.
\newblock \emph{arXiv preprint arXiv:2608.26336}, 2026{\natexlab{c}}.

\bibitem[Liu et~al.(2024)Liu, Cun, Liu, Wang, Zhang, Chen, Liu, Zeng, Chan, and Shan]{liu2024evalcrafter}
Yaofang Liu, Xiaodong Cun, Xuebo Liu, Xintao Wang, Yong Zhang, Haoxin Chen, Yang Liu, Tieyong Zeng, Raymond Chan, and Ying Shan.
\newblock Evalcrafter: Benchmarking and evaluating large video generation models.
\newblock In \emph{2024 IEEE/CVF Conference on Computer Vision and Pattern Recognition (CVPR)}, 2024.

\bibitem[Meng et~al.(2026)Meng, Liu, Ouyang, Wang, Cheng, Yu, Wang, Li, Zhu, Zeng, et~al.]{meng2026causalcine}
Yihao Meng, Zichen Liu, Hao Ouyang, Qiuyu Wang, Ka~Leong Cheng, Yue Yu, Hanlin Wang, Haobo Li, Jiapeng Zhu, Yanhong Zeng, et~al.
\newblock Causalcine: Real-time autoregressive generation for multi-shot video narratives.
\newblock \emph{arXiv preprint arXiv:2605.12496}, 2026.

\bibitem[Miao et~al.(2026)Miao, Zhang, Xu, Yu, Tang, Chu, Shen, and Lin]{miao2026video}
Yingmao Miao, Pengfei Zhang, Chaoran Xu, Meng Yu, Jing Tang, Xiangxiang Chu, Chao Shen, and Chenhao Lin.
\newblock Do video generators track the world across segments? a benchmark and method for world-state reasoning in video continuation.
\newblock \emph{arXiv preprint arXiv:2609.03673}, 2026.

\bibitem[{PixVerse}(2026)]{pixverse2026r1}
{PixVerse}.
\newblock {PixVerse Launches R1: A Real-Time World Model That Redefines AI Video Generation}, 2026.

\bibitem[Qwen et~al.(2025)Qwen, Yang, Yang, Zhang, Hui, Zheng, Yu, Li, Liu, Huang, Wei, Lin, Yang, Tu, Zhang, Yang, Yang, Zhou, Lin, Dang, Lu, Bao, Yang, Yu, Li, Xue, Zhang, Zhu, Men, Lin, Li, Tang, Xia, Ren, Ren, Fan, Su, Zhang, Wan, Liu, Cui, Zhang, and Qiu]{qwen2025qwen25technicalreport}
Qwen, An Yang, Baosong Yang, Beichen Zhang, Binyuan Hui, Bo Zheng, Bowen Yu, Chengyuan Li, Dayiheng Liu, Fei Huang, Haoran Wei, Huan Lin, Jian Yang, Jianhong Tu, Jianwei Zhang, Jianxin Yang, Jiaxi Yang, Jingren Zhou, Junyang Lin, Kai Dang, Keming Lu, Keqin Bao, Kexin Yang, Le Yu, Mei Li, Mingfeng Xue, Pei Zhang, Qin Zhu, Rui Men, Runji Lin, Tianhao Li, Tianyi Tang, Tingyu Xia, Xingzhang Ren, Xuancheng Ren, Yang Fan, Yang Su, Yichang Zhang, Yu Wan, Yuqiong Liu, Zeyu Cui, Zhenru Zhang, and Zihan Qiu.
\newblock Qwen2.5 technical report, 2025.

\bibitem[Radford et~al.(2021)Radford, Kim, Hallacy, Ramesh, Goh, Agarwal, Sastry, Askell, Mishkin, Clark, Krueger, and Sutskever]{radford2021learningtransferablevisualmodels}
Alec Radford, Jong~Wook Kim, Chris Hallacy, Aditya Ramesh, Gabriel Goh, Sandhini Agarwal, Girish Sastry, Amanda Askell, Pamela Mishkin, Jack Clark, Gretchen Krueger, and Ilya Sutskever.
\newblock Learning transferable visual models from natural language supervision, 2021.

\bibitem[Rana et~al.(2023)Rana, Haviland, Garg, Abou-Chakra, Reid, and Suenderhauf]{rana2023sayplan}
Krishan Rana, Jesse Haviland, Sourav Garg, Jad Abou-Chakra, Ian Reid, and Niko Suenderhauf.
\newblock Sayplan: Grounding large language models using 3d scene graphs for scalable robot task planning.
\newblock \emph{arXiv preprint arXiv:2307.06135}, 2023.

\bibitem[Rao et~al.(2026)Rao, Zhang, Liu, Meng, Zhang, Wan, Tang, and Lee]{rao2026streamhoi}
Zejing Rao, Haoxian Zhang, Xiaoqiang Liu, Yiping Meng, Guoxin Zhang, Pengfei Wan, Fan Tang, and Tong-Yee Lee.
\newblock Streamhoi: Interaction-aware temporal memory adaptation for streaming hoi video generation.
\newblock \emph{arXiv preprint arXiv:2607.20174}, 2026.

\bibitem[Sun et~al.(2025)Sun, Huang, Liu, Wu, Xu, Li, and Liu]{sun2025t2v}
Kaiyue Sun, Kaiyi Huang, Xian Liu, Yue Wu, Zihan Xu, Zhenguo Li, and Xihui Liu.
\newblock T2v-compbench: A comprehensive benchmark for compositional text-to-video generation.
\newblock In \emph{2025 IEEE/CVF Conference on Computer Vision and Pattern Recognition (CVPR)}, 2025.

\bibitem[Tan et~al.(2026)Tan, Li, Zhang, Jia, Yang, Zhang, and Zhang]{tan2026swift}
Shanwen Tan, Hao Li, Jingtao Zhang, Xiaosong Jia, Xue Yang, Shaofeng Zhang, and Yanyong Zhang.
\newblock Swift: Prompt-adaptive memory for efficient interactive long video generation.
\newblock \emph{arXiv preprint arXiv:2605.09442}, 2026.

\bibitem[Team(2025)]{qwen3technicalreport}
Qwen Team.
\newblock Qwen3 technical report, 2025.

\bibitem[Team et~al.(2026)Team, Gao, Wang, Zeng, Zhu, Cheng, Li, Wang, Xu, Ma, Chen, Liu, Cheng, Yao, Zhu, Meng, Zheng, Bai, Chen, Shen, Yu, Zhu, Shen, and Ouyang]{lingbot-world}
Robbyant Team, Zelin Gao, Qiuyu Wang, Yanhong Zeng, Jiapeng Zhu, Ka~Leong Cheng, Yixuan Li, Hanlin Wang, Yinghao Xu, Shuailei Ma, Yihang Chen, Jie Liu, Yansong Cheng, Yao Yao, Jiayi Zhu, Yihao Meng, Kecheng Zheng, Qingyan Bai, Jingye Chen, Zehong Shen, Yue Yu, Xing Zhu, Yujun Shen, and Hao Ouyang.
\newblock Advancing open-source world models.
\newblock \emph{arXiv preprint arXiv:2601.20540}, 2026.

\bibitem[{Video Understanding Team of JoyAI-VL @ Joy Future Academy, JD}(2026)]{joyai2026vlinteraction}
{Video Understanding Team of JoyAI-VL @ Joy Future Academy, JD}.
\newblock Joyai-vl-interaction: Real-time vision-language interaction intelligence.
\newblock Technical report, Joy Future Academy, JD, 2026.

\bibitem[Wan et~al.(2025)Wan, Wang, Ai, Wen, Mao, Xie, Chen, Yu, Zhao, Yang, Zeng, Wang, Zhang, Zhou, Wang, Chen, Zhu, Zhao, Yan, Huang, Feng, Zhang, Li, Wu, Chu, Feng, Zhang, Sun, Fang, Wang, Gui, Weng, Shen, Lin, Wang, Wang, Zhou, Wang, Shen, Yu, Shi, Huang, Xu, Kou, Lv, Li, Liu, Wang, Zhang, Huang, Li, Wu, Liu, Pan, Zheng, Hong, Shi, Feng, Jiang, Han, Wu, and Liu]{wan2025}
Team Wan, Ang Wang, Baole Ai, Bin Wen, Chaojie Mao, Chen-Wei Xie, Di Chen, Feiwu Yu, Haiming Zhao, Jianxiao Yang, Jianyuan Zeng, Jiayu Wang, Jingfeng Zhang, Jingren Zhou, Jinkai Wang, Jixuan Chen, Kai Zhu, Kang Zhao, Keyu Yan, Lianghua Huang, Mengyang Feng, Ningyi Zhang, Pandeng Li, Pingyu Wu, Ruihang Chu, Ruili Feng, Shiwei Zhang, Siyang Sun, Tao Fang, Tianxing Wang, Tianyi Gui, Tingyu Weng, Tong Shen, Wei Lin, Wei Wang, Wei Wang, Wenmeng Zhou, Wente Wang, Wenting Shen, Wenyuan Yu, Xianzhong Shi, Xiaoming Huang, Xin Xu, Yan Kou, Yangyu Lv, Yifei Li, Yijing Liu, Yiming Wang, Yingya Zhang, Yitong Huang, Yong Li, You Wu, Yu Liu, Yulin Pan, Yun Zheng, Yuntao Hong, Yupeng Shi, Yutong Feng, Zeyinzi Jiang, Zhen Han, Zhi-Fan Wu, and Ziyu Liu.
\newblock Wan: Open and advanced large-scale video generative models.
\newblock \emph{arXiv preprint arXiv:2503.20314}, 2025.

\bibitem[Wang et~al.(2026{\natexlab{a}})Wang, Zhou, Liang, Liu, Pang, Xie, Pang, and Liang]{wang2026tempact}
Jing Wang, Xiangxin Zhou, Jiajun Liang, Kaiqi Liu, Wanyun Pang, Zhenyu Xie, Tianyu Pang, and Xiaodan Liang.
\newblock Tempact: Advancing temporal plausibility in autoregressive video generation via planner-executor rl.
\newblock \emph{arXiv preprint arXiv:2606.28016}, 2026{\natexlab{a}}.

\bibitem[Wang and Yang(2024)]{wang2024vidprom}
Wenhao Wang and Yi Yang.
\newblock Vidprom: A million-scale real prompt-gallery dataset for text-to-video diffusion models.
\newblock \emph{Thirty-eighth Conference on Neural Information Processing Systems}, 2024.

\bibitem[Wang et~al.(2025)Wang, Gao, Gu, Pu, Cui, Wei, Liu, Jing, Ye, Shao, Wang, Chen, Zhang, Yang, Wang, Wei, Yin, Li, Cui, Chen, Ding, Tian, Wu, Xie, Li, Yang, Duan, Wang, Hou, Hao, Zhang, Li, Zhao, Duan, Deng, Fu, He, Wang, He, Shi, He, Xiong, Lv, Wu, Shao, Zhang, Deng, Qi, Ge, Guo, Zhang, Zhang, Cao, Lin, Tang, Gao, Huang, Gu, Lyu, Tang, Wang, Lv, Ouyang, Wang, Dou, Zhu, Lu, Lin, Dai, Su, Zhou, Chen, Qiao, Wang, and Luo]{wang2025internvl35advancingopensourcemultimodal}
Weiyun Wang, Zhangwei Gao, Lixin Gu, Hengjun Pu, Long Cui, Xingguang Wei, Zhaoyang Liu, Linglin Jing, Shenglong Ye, Jie Shao, Zhaokai Wang, Zhe Chen, Hongjie Zhang, Ganlin Yang, Haomin Wang, Qi Wei, Jinhui Yin, Wenhao Li, Erfei Cui, Guanzhou Chen, Zichen Ding, Changyao Tian, Zhenyu Wu, Jingjing Xie, Zehao Li, Bowen Yang, Yuchen Duan, Xuehui Wang, Zhi Hou, Haoran Hao, Tianyi Zhang, Songze Li, Xiangyu Zhao, Haodong Duan, Nianchen Deng, Bin Fu, Yinan He, Yi Wang, Conghui He, Botian Shi, Junjun He, Yingtong Xiong, Han Lv, Lijun Wu, Wenqi Shao, Kaipeng Zhang, Huipeng Deng, Biqing Qi, Jiaye Ge, Qipeng Guo, Wenwei Zhang, Songyang Zhang, Maosong Cao, Junyao Lin, Kexian Tang, Jianfei Gao, Haian Huang, Yuzhe Gu, Chengqi Lyu, Huanze Tang, Rui Wang, Haijun Lv, Wanli Ouyang, Limin Wang, Min Dou, Xizhou Zhu, Tong Lu, Dahua Lin, Jifeng Dai, Weijie Su, Bowen Zhou, Kai Chen, Yu Qiao, Wenhai Wang, and Gen Luo.
\newblock Internvl3.5: Advancing open-source multimodal models in versatility, reasoning, and efficiency, 2025.

\bibitem[Wang et~al.(2026{\natexlab{b}})Wang, Zhao, Zhan, and Ma]{wang2026liveedit}
Xinyu Wang, Chongbo Zhao, Fangneng Zhan, and Yue Ma.
\newblock Liveedit: Towards real-time diffusion-based streaming video editing.
\newblock \emph{arXiv preprint arXiv:2606.26740}, 2026{\natexlab{b}}.

\bibitem[Wu et~al.(2025)Wu, Siarohin, Menapace, Skorokhodov, Fang, Chordia, Gilitschenski, and Tulyakov]{wu2025mind}
Ziyi Wu, Aliaksandr Siarohin, Willi Menapace, Ivan Skorokhodov, Yuwei Fang, Varnith Chordia, Igor Gilitschenski, and Sergey Tulyakov.
\newblock Mind the time: Temporally-controlled multi-event video generation.
\newblock In \emph{2025 IEEE/CVF Conference on Computer Vision and Pattern Recognition (CVPR)}, 2025.

\bibitem[Yang et~al.(2025{\natexlab{a}})Yang, Huang, Chu, Xiao, Zhao, Wang, Li, Xie, Chen, Lu, et~al.]{yang2025longlive}
Shuai Yang, Wei Huang, Ruihang Chu, Yicheng Xiao, Yuyang Zhao, Xianbang Wang, Muyang Li, Enze Xie, Yingcong Chen, Yao Lu, et~al.
\newblock Longlive: Real-time interactive long video generation.
\newblock \emph{arXiv preprint arXiv:2509.22622}, 2025{\natexlab{a}}.

\bibitem[Yang et~al.(2025{\natexlab{b}})Yang, Li, Zhang, Yin, Bai, Ma, Wang, Cai, Wong, Lu, et~al.]{yang2025vlipp}
Xindi Yang, Baolu Li, Yiming Zhang, Zhenfei Yin, Lei Bai, Liqian Ma, Zhiyong Wang, Jianfei Cai, Tien-Tsin Wong, Huchuan Lu, et~al.
\newblock Vlipp: Towards physically plausible video generation with vision and language informed physical prior.
\newblock In \emph{2025 IEEE/CVF International Conference on Computer Vision (ICCV)}, 2025{\natexlab{b}}.

\bibitem[Yang et~al.(2026)Yang, Zhang, Huang, Chen, Wu, He, Cai, Li, and Jiang]{yang2026anchor}
Yang Yang, Tianyi Zhang, Wei Huang, Jinwei Chen, Boxi Wu, Xiaofei He, Deng Cai, Bo Li, and Peng-Tao Jiang.
\newblock Anchor forcing: Anchor memory and tri-region rope for interactive streaming video diffusion.
\newblock \emph{arXiv preprint arXiv:2603.13405}, 2026.

\bibitem[Yi et~al.(2025)Yi, Jang, Cho, Nam, Yoon, and Kim]{yi2025deepforcing}
Jung Yi, Wooseok Jang, Paul~Hyunbin Cho, Jisu Nam, Heeji Yoon, and Seungryong Kim.
\newblock Deep forcing: Training-free long video generation with deep sink and participative compression.
\newblock \emph{arXiv preprint arXiv:2512.05081}, 2025.

\bibitem[Yin et~al.(2024{\natexlab{a}})Yin, Gharbi, Park, Zhang, Shechtman, Durand, and Freeman]{yin2024dmd2}
Tianwei Yin, Micha{\"e}l Gharbi, Taesung Park, Richard Zhang, Eli Shechtman, Fredo Durand, and William~T Freeman.
\newblock Improved distribution matching distillation for fast image synthesis.
\newblock In \emph{NeurIPS}, 2024{\natexlab{a}}.

\bibitem[Yin et~al.(2024{\natexlab{b}})Yin, Gharbi, Zhang, Shechtman, Durand, Freeman, and Park]{yin2024dmd}
Tianwei Yin, Micha{\"e}l Gharbi, Richard Zhang, Eli Shechtman, Fr{\'e}do Durand, William~T Freeman, and Taesung Park.
\newblock One-step diffusion with distribution matching distillation.
\newblock In \emph{CVPR}, 2024{\natexlab{b}}.

\bibitem[Yin et~al.(2025)Yin, Zhang, Zhang, Freeman, Durand, Shechtman, and Huang]{yin2025causvid}
Tianwei Yin, Qiang Zhang, Richard Zhang, William~T Freeman, Fredo Durand, Eli Shechtman, and Xun Huang.
\newblock From slow bidirectional to fast autoregressive video diffusion models.
\newblock In \emph{CVPR}, 2025.

\bibitem[Zhang et~al.(2024{\natexlab{a}})Zhang, Xie, Lu, Ma, Li, and Guo]{zhang2024mavin}
Bowen Zhang, Xiaofei Xie, Haotian Lu, Na Ma, Tianlin Li, and Qing Guo.
\newblock Mavin: Multi-action video generation with diffusion models via transition video infilling.
\newblock \emph{arXiv preprint arXiv:2405.18003}, 2024{\natexlab{a}}.

\bibitem[Zhang et~al.(2024{\natexlab{b}})Zhang, Yaosen, Liu, Wang, Wen, and Wang]{zhang2024tvg}
Rui Zhang, Chen Yaosen, Yuegen Liu, Wei Wang, Xuming Wen, and Hongxia Wang.
\newblock Tvg: A training-free transition video generation method with diffusion models.
\newblock In \emph{arxiv}, 2024{\natexlab{b}}.

\bibitem[Zhao et~al.(2026{\natexlab{a}})Zhao, Zhu, Zheng, Zhou, Yan, Li, Yang, Li, and Zhu]{zhao2026causal}
Min Zhao, Hongzhou Zhu, Kaiwen Zheng, Zihan Zhou, Bokai Yan, Xinyuan Li, Xiao Yang, Chongxuan Li, and Jun Zhu.
\newblock Causal forcing++: Scalable few-step autoregressive diffusion distillation for real-time interactive video generation.
\newblock \emph{arXiv preprint arXiv:2605.15141}, 2026{\natexlab{a}}.

\bibitem[Zhao et~al.(2026{\natexlab{b}})Zhao, Pan, He, Yu, Chen, Ye, Liu, Xie, and Han]{zhao2026sana}
Yuyang Zhao, Yicheng Pan, Qiyuan He, Jincheng Yu, Junsong Chen, Tian Ye, Haozhe Liu, Enze Xie, and Song Han.
\newblock Sana-streaming: Real-time streaming video editing with hybrid diffusion transformer.
\newblock \emph{arXiv preprint arXiv:2605.30409}, 2026{\natexlab{b}}.

\bibitem[Zheng et~al.(2025)Zheng, Huang, Liu, Zou, He, Zhang, Zhang, He, Zheng, Qiao, and Liu]{zheng2025vbench}
Dian Zheng, Ziqi Huang, Hongbo Liu, Kai Zou, Yinan He, Fan Zhang, Yuanhan Zhang, Jingwen He, Wei-Shi Zheng, Yu Qiao, and Ziwei Liu.
\newblock Vbench-2.0: Advancing video generation benchmark suite for intrinsic faithfulness.
\newblock \emph{arXiv preprint arXiv:2503.21755}, 2025.

\bibitem[Zhu et~al.(2026)Zhu, Zhao, He, Su, Li, and Zhu]{zhu2026causal}
Hongzhou Zhu, Min Zhao, Guande He, Hang Su, Chongxuan Li, and Jun Zhu.
\newblock Causal forcing: Autoregressive diffusion distillation done right for high-quality real-time interactive video generation.
\newblock \emph{arXiv preprint arXiv:2602.02214}, 2026.

\end{thebibliography}
